\documentclass[runningheads]{llncs}

\usepackage{eccv}

\usepackage{eccvabbrv}
\usepackage{amsmath}
\usepackage{epsfig}
\usepackage{graphicx}
\usepackage{booktabs}
\usepackage{subcaption}
\usepackage{comment}
\usepackage{multirow}
\usepackage{multicol}

\definecolor{cvprblue}{rgb}{0.21,0.49,0.74}
\usepackage[normalem]{ulem}
\useunder{\uline}{\ul}{}
\usepackage{pdflscape}

\usepackage[rightcaption]{sidecap}

\usepackage[pagebackref,breaklinks,colorlinks,citecolor=eccvblue]{hyperref}
\usepackage{hyperref}

\usepackage{orcidlink}

\begin{document}
	
\title{From Multi-Resolution Cells to Gigapixel Whole Slide Images Foundation Model for Computational Pathology}
 

\titlerunning{Multi-Resolution Pyramid Transformer}

\author{Basit Alawode\inst{1}\orcidlink{0000-0001-6680-7948} \and
	Moshira Ali Abdalla\inst{1}\orcidlink{0009-0008-2608-5352} \and
	Dwarikanath Mahapatra\inst{1}\orcidlink{0000-0001-9749-7858} \and
    Muzammal Naseer \inst{1,2}\orcidlink{0000-0001-7663-7161} \and
    Sajid Javed \inst{1}\orcidlink{0000-0002-0036-2875}} 

\authorrunning{B.~Alawode et al.}

\institute{Khalifa University of Science and Technology, UAE \and
	University of Western Australia, Australia \\
	\email{\{basit.alawode,sajid.javed\}@ku.ac.ae}}
\maketitle

\begin{abstract}
Vision Transformers (ViTs) and their hierarchical variants have achieved strong performance in Computational Pathology (CPath).
However, most are pre-trained on single-resolution Whole Slide Images (WSIs), limiting their generalization across arbitrary resolutions. 
Gigapixel WSIs inherently contain diagnostic patterns at multiple scales, including cellular morphologies, tissue architectures, and global context, mirroring how expert pathologists examine WSIs. 
We introduce Multi-Resolution Pyramid Transformer (MRPT), a model that hierarchically aggregates multi-resolution information from cellular to tissue and WSI levels.
MRPT employs a biologically meaningful Consecutive Cross-Resolution Attention (CCRA) mechanism to capture scale-independent interactions and enforces multi-resolution semantic consistency by aligning embeddings across resolutions, yielding robust and generalizable WSI representations.
Pre-trained in a multi-resolution self-supervised manner on 624M patches, 2.4M regions, and 36K WSIs, MRPT learns rich coarse-to-fine histopathology features.
Extensive experiments on 34 diverse datasets show that MRPT surpasses recent foundation models and Multimodal Large Language Models (MLLMs) in cancer subtype classification, tissue phenotyping, and Visual Question Answering (VQA) for WSI understanding. Our code and models available on \href{https://github.com/BasitAlawode}{link}.
\end{abstract}
\vspace{-8mm}
\section{Introduction}
\label{sec:intro}

\begin{figure}[t]
\centering

\begin{minipage}{0.48\columnwidth}
    \begin{subfigure}{\textwidth}
        \centering
        \includegraphics[width=\textwidth]{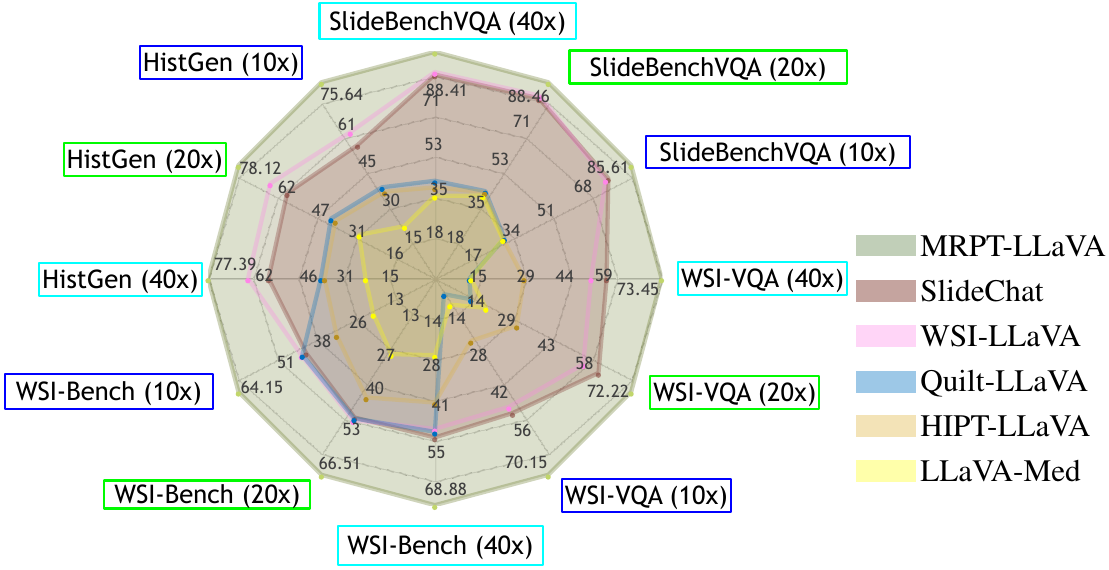}
        \caption{WSI VQA Task (Accuracy)}
        \label{fig1:vqa}
    \end{subfigure}

    \vspace{1mm} 

    \begin{subfigure}{\textwidth}
        \centering
        \includegraphics[width=0.9\textwidth]{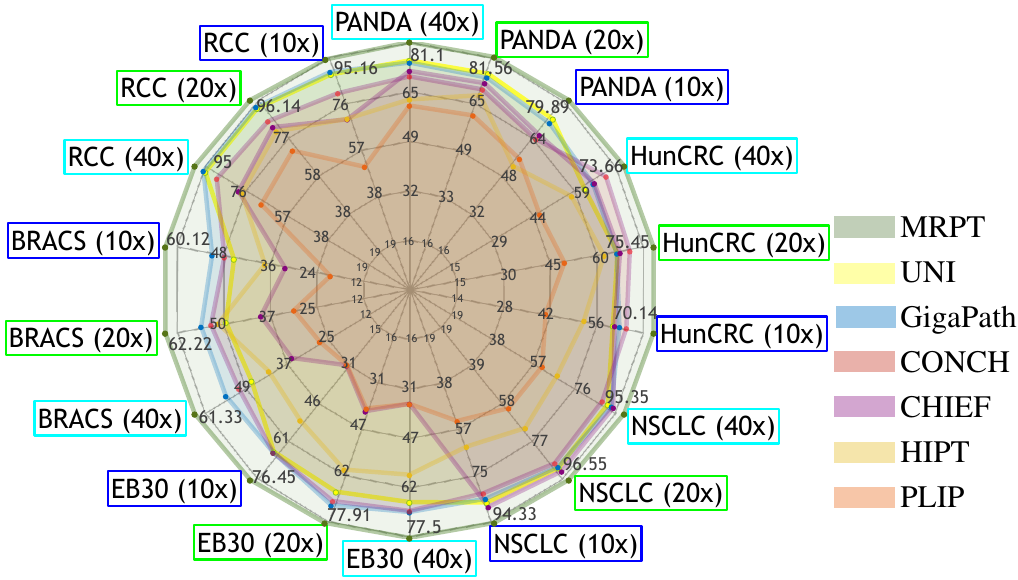}
        \caption{WSI Classification Task (Balanced Accuracy)}
        \label{fig1:classification}
    \end{subfigure}
\end{minipage}
\hfill
\begin{minipage}{0.48\columnwidth}
    \begin{subfigure}{\textwidth}
        \centering
        \includegraphics[width=0.8\textwidth, height=7cm]{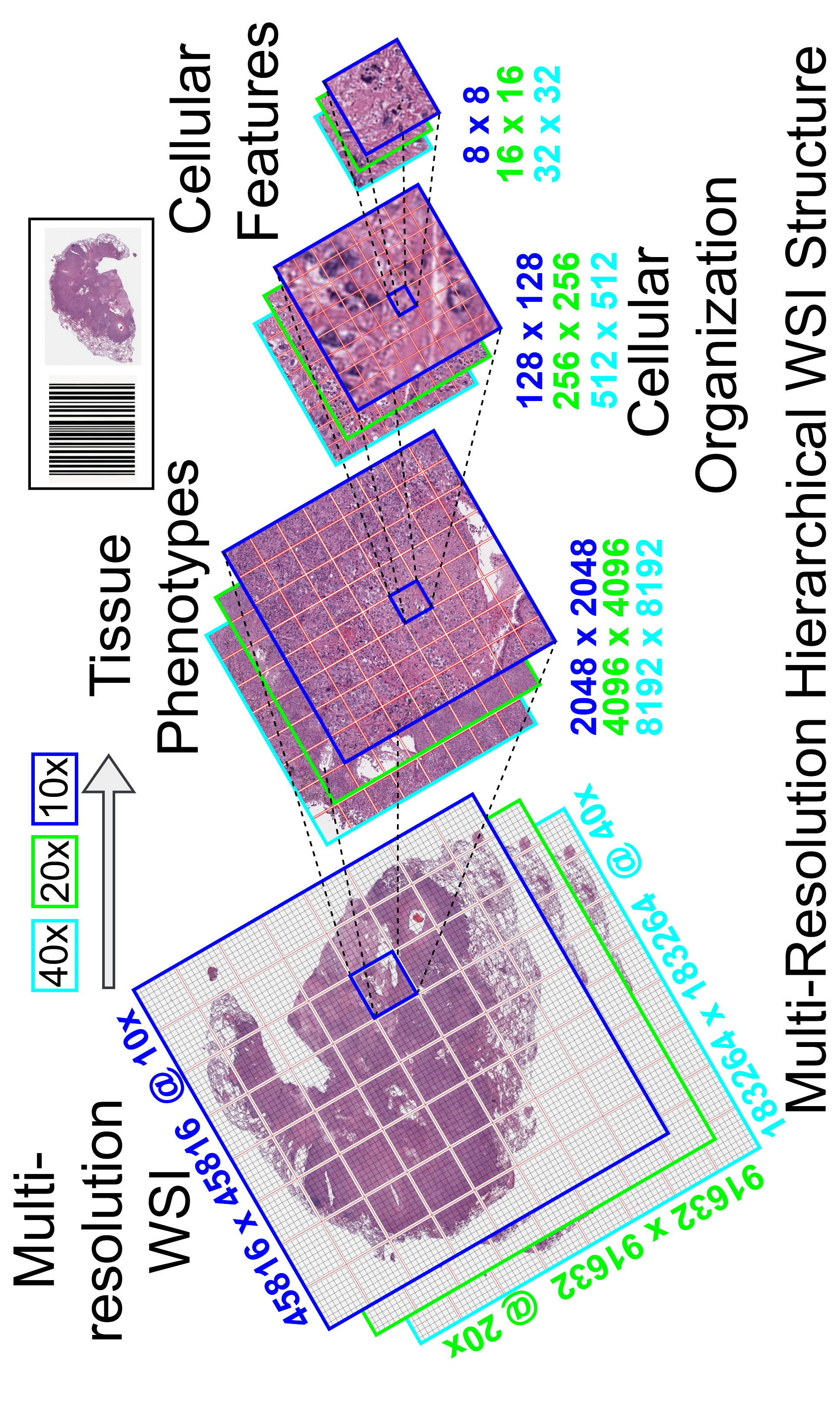}
        \caption{MRPT Hierarchical and Multi-resolution representation}
        \label{fig1:mrpt}
    \end{subfigure}
\end{minipage}
\vspace{-3mm}
\caption{\textbf{(a)} and \textbf{(b)}: Our MRPT model remains consistent across resolutions and outperforms existing SOTA models by a significant margin. \textbf{(c)}: MRPT exploits hierarchical and multi-resolution representations to capture rich contextual information from WSIs.}
\label{fig1}
\vspace{-9mm}
\end{figure}

Diagnosing cancer and assessing patient outcomes from gigapixel Whole Slide Images (WSIs) remains the gold standard in clinical pathology \cite{mcguire2016world, petersen2009oral, zech1995validation}.
The digitization of glass slides into WSIs has transformed modern pathology, enabling large-scale and systematic computational analysis of tissue morphology \cite{echle2021deep, qu2024rise, song2023artificial, van2021deep, cui2021artificial}.
This digital transformation has given rise to the field of Computational Pathology (CPath), which leverages advances in machine learning and computer vision to automatically extract diagnostic and prognostic insights from gigapixel WSIs \cite{srinidhi2021deep, ahmedt2022survey, hosseini2024computational, louis2016computational}.
\textit{A WSI inherently exhibits a hierarchical and multi-resolution structures (Fig. \ref{fig1:mrpt}) \cite{albastaki2025multi, chen2022scaling, vanRijthoven2021HookNet, abels2019computational}.}
The hierarchical structure reflects the nested organization of visual information: small regions comprising individual cellular entities (e.g., tumor cells, stroma, lymphocytes) form local clusters that capture cell-cell interactions \cite{chen2022scaling}.
These clusters aggregate into larger tissue microenvironments, which collectively compose the full WSI, encapsulating the overall tissue heterogeneity \cite{chen2022fast, srinidhi2021deep, lu2020capturing}.
Conversely, the multi-resolution structure refers to viewing the same spatial region at varying resolutions (e.g., 10$\times$, 20$\times$, or 40$\times$), where each resolution maintains spatial alignment while offering distinct levels of visual detail \cite{hanna2020whole, jain2024whole, zarella2019practical, ardon2025digital}.

Expert pathologists routinely perform holistic analyses of WSIs across multiple resolutions, synthesizing evidence from different spatial scales and regions to generate comprehensive diagnostic reports \cite{baidoshvili2023whole, nan2025deep, zarella2022high}.
For example, in colorectal cancer, low resolutions capture global tissue architecture such as mucosal orientation, intermediate resolutions reveal glandular organization and architectural distortion, and high resolution expose cellular features such as nuclear atypia, pleomorphism, and mitotic activity \cite{tepus2020non, harada2020molecular}.
\textit{Therefore, multi-resolution analysis of WSIs is indispensable for accurate cancer diagnosis, as it integrates complementary information across scales (Fig. \ref{fig1:mrpt}) \cite{schuffler2022efficient, ghezloo2022analysis}.}

Recently, numerous CPath foundation models have been developed to help pathologists in diagnostic and prognostic decision-making \cite{song2023artificial, lu2024visual, chen2024towards}.
Most SOTA models are patch-level at a single resolution such as CONCH \cite{lu2024visual}, UNI \cite{chen2024towards}, and QuiltNet \cite{ikezogwo2024quilt}.
MR-PLIP is the only multi-resolution patch-level model \cite{albastaki2025multi}; however, it does not fully exploit the hierarchical structure. 
In SOTA models, a WSI is divided into smaller patches during training \cite{albastaki2025multi, ikezogwo2024quilt, wang2022transformer}.
\textit{However, such patch-level representations often lack global spatial context of the entire WSI, limiting their ability to capture tissue-level organization and cross-regional dependencies \cite{ding2024multimodal, chen2025slidechat}.}
To overcome this limitation, several WSI-level models, such as GigaPath \cite{xu2024whole}, WSI-LLaVA \cite{liang2025wsillavamultimodallargelanguage}, and SlideChat \cite{chen2025slidechat}, have been introduced.
These methods aggregate information across all patches to learn holistic WSI-level representations, thereby incorporating global contextual cues \cite{ding2024multimodal}.
\textit{While these models successfully encode global information, they often overlook the intrinsic WSI hierarchical organization, resulting in suboptimal WSI-level representations and reduced performance (Figs. \ref{fig1} (a) \& (b)) \cite{ding2024multimodal}.}
To address this, hierarchical models have been proposed, which exploit the nested structure of WSIs for improved feature learning \cite{chen2022scaling, jin2024hmil, guo2023higt}.
For instance, the HIPT model captures the natural hierarchy within WSIs via a two-level Self-Supervised Learning (SSL) framework, producing enriched WSI representations \cite{chen2022scaling}.
While hierarchical modeling has improved WSI understanding, \textit{the explicit integration of multi-resolution information within these hierarchies remains largely underexplored (see Table 1 in supplementary for more comparisons of SOTA models).}

\noindent To this end, \textit{we propose a novel Multi-Resolution Pyramid Transformer (\textbf{MRPT}) that jointly models hierarchical and multi-resolution representations at the WSI level.}
MRPT captures complementary diagnostic cues across multi-resolution cellular, tissue, and WSI scales, mirroring the diagnostic workflow of expert pathologists who continuously zoom in and out to assess both local details and global context.
This design enables the model to preserve fine-grained morphological information while maintaining global spatial awareness, resulting in a more robust, generalizable, and clinically meaningful CPath model.
\textit{MRPT is hierarchically pre-trained on 624M patches, 2.4M regions, and 36K WSIs curated from TCGA \cite{hutter2018cancer} and CPTAC \cite{Edwards2015CPTAC}, leveraging multi-resolution SSL.}
To maintain biological interpretability and semantic consistency, MRPT performs cross-attention only between consecutive resolutions, emulating the progressive zooming behavior of pathologists.
At the cell-level pre-training stage, we introduce a \textit{Consecutive Cross-Resolution Attention (CCRA)} mechanism that fuses information across resolutions using a novel ViT architecture designed to align token representations at multiple scales.
The aggregated cell-level embeddings are used to represent corresponding patches, which are trained via multi-resolution SSL at the patch-level.
Subsequently, the aggregated multi-resolution patch-level features are employed to represent higher-level regions.
At the region-level, SSL is again applied to learn region-specific representations, which are ultimately aggregated into holistic, multi-resolution WSI-level embeddings.
Finally, SSL is performed at the WSI-level to learn a unified, hierarchical representation encompassing all three scales.
\textit{Through this pipeline, MRPT effectively learns a multi-resolution, hierarchically structured representation.}

To further extend MRPT, we introduce a multimodal conversational model for WSI understanding, termed \textbf{MRPT-LLaVA}.
This MLLM integrates MRPT representations with LLaVA \cite{liu2023visual} to facilitate cross-modal reasoning.
Most existing MLLMs such as Quilt-LLaVA \cite{seyfioglu2024quilt}, operate only at the patch-level, thereby lacking global WSI-level context.
Recent WSI-level models, including SlideChat \cite{chen2025slidechat} and WSI-LLaVA \cite{liang2025wsillavamultimodallargelanguage}, incorporate WSI information but rely on a single resolution, limiting their ability to capture hierarchical and multi-resolution representations.
\textit{In contrast, MRPT-LLaVA leverages multi-resolution hierarchical WSI-level representations to perform complex pathology VQA tasks.}
MRPT-LLaVA employs a three-stage training approach: \textit{cross-modal alignment}, \textit{feature space alignment}, and \textit{visual-instruction tuning} on WSI-Bench dataset \cite{liang2025wsillavamultimodallargelanguage}.
To evaluate the effectiveness of MRPT, we conduct extensive experiments on 34 publicly available datasets covering a range of CPath tasks, including classification, caption generation, and VQA.
Our model achieves superior performance compared to the SOTA CPath foundation models.
Our main contributions are:
\begin{enumerate}
\item We introduce a three-stage multi-resolution Self-Supervised Learning (SSL) paradigm that jointly models multi-resolution and hierarchical representations of WSIs, enabling comprehensive learning of both fine-grained cellular features and global tissue context.
\item We propose a novel multi-resolution cell-level ViT architecture based on the biologically inspored Consecutive Cross-Resolution Attention (CCRA) mechanism, which efficiently integrates cellular information across multi-resolution.
\item We develop MRPT-LLaVA, a CPath MLLM that leverages multi-resolution hierarchical WSI representations to perform complex VQA tasks.
\end{enumerate}

\section{{Literature Review}}
\label{sec:review}
\noindent \textbf{1. Vision-only Foundation Models (VFMs):} These models are pre-trained on large-scale histopathology datasets using SSL and serve as backbones for downstream tasks \cite{huang2023visual, kang2023benchmarking, horst2024cellvit}. 
Most existing VFMs including UNI \cite{chen2024towards}, REMEDIS \cite{azizi2022robust}, and CHIEF \cite{wang2024pathology}, operate at the patch-level \cite{lu2024visual, ikezogwo2024quilt}, and employ a DINO-based SSL framework using millions scale histology patches \cite{caron2020unsupervised}.
Many patch-level VFMs including Virchow \cite{vorontsov2024foundation} and GigaPath \cite{xu2024whole}, aggregate patch-level embeddings to form holistic WSI representations \cite{vorontsov2024foundation, xu2024whole, wang2024pathology}.
\textit{Patch-level learning inherently lacks the comprehensive global context embedded within the WSI}.
Hierarchical models use SSL to combine local and global features \cite{jin2024hmil, chen2022scaling, guo2023higt}.
A well-known example is HIPT \cite{chen2022scaling}, which is pre-trained at the cellular and tissue levels, and merges them into a WSI-level embedding \cite{chen2022scaling}.
\textit{Nonetheless, existing hierarchical approaches neglect the inherent multi-resolution WSI information, which can lead to suboptimal representation and performance degradation.}
Some patch-level VFMs, including RudolfV \cite{dippel2024rudolfv} and Virchow2 \cite{vorontsov2024foundation}, utilize multi-resolution patches primarily to enhance dataset diversity \cite{albastaki2025multi, vanRijthoven2021HookNet, kang2023benchmarking}.
However, these models do not explicitly exploit the cross-resolution dependencies within WSIs.
\textit{Our proposed MRPT introduces a WSI-level framework that hierarchically fuses information across multi-resolution, effectively combining the benefits of global context and multi-resolution information hierarchy.}
\textbf{2. Vision-Language Models (VLMs):} CPath VLMs including PLIP \cite{huang2023visual}, QuiltNet \cite{ikezogwo2024quilt}, and CONCH \cite{lu2024visual} etc., learn a shared embedding space between histology images and textual descriptions using contrastive learning \cite{radford2021learning}, enabling cross-modal alignment, retrieval, and zero-shot classification \cite{lu2023visual, javed2024cplip, huang2023visual}.
Most existing VLMs, however, are pre-trained at a single resolution, failing to capture the coarse-to-fine contextual hierarchy in WSIs.
MR-PLIP is the only multi-resolution patch-level  VLM \cite{albastaki2025multi}; however, it does not fully exploit the hierarchical structure. 
\textit{Aggregating local features without respecting the hierarchical structure of WSIs inherently limits the global contextual understanding necessary for comprehensive WSI-level interpretation.}
\textbf{3. MLLMs:} These models integrate histology images, textual information, and patient-level clinical data into a unified reasoning framework for complex VQA task \cite{chen2025slidechat, seyfioglu2024quilt}.
Most existing MLLMs operate at the \textit{patch-level}, including PathChat \cite{lu2024multimodal}, Quilt-LLaVA \cite{seyfioglu2024quilt}, and PathAsst \cite{sun2024pathasst} etc. 
WSI-level MLLMs, such as TITAN \cite{ding2024multimodal}, SlideChat \cite{chen2025slidechat}, and WSI-LLaVA \cite{liang2025wsillavamultimodallargelanguage}, aggregate patch-level local features for WSI reasoning.
While these MLLMs have significantly advanced complex WSI understanding, they still fail to leverage the intrinsic WSIs \textit{hierarchical} and \textit{multi-resolution} structures.
\textit{Our MRPT-LLaVA explicitly exploits these structures in a WSI.}
See Table 1 in supplementary for more SOTA models comparisons.
\vspace{-2mm}
\section{{Methodology}}
\label{sec:method}
\begin{figure*}[t!]
   \centering 
    \includegraphics[width=\linewidth]{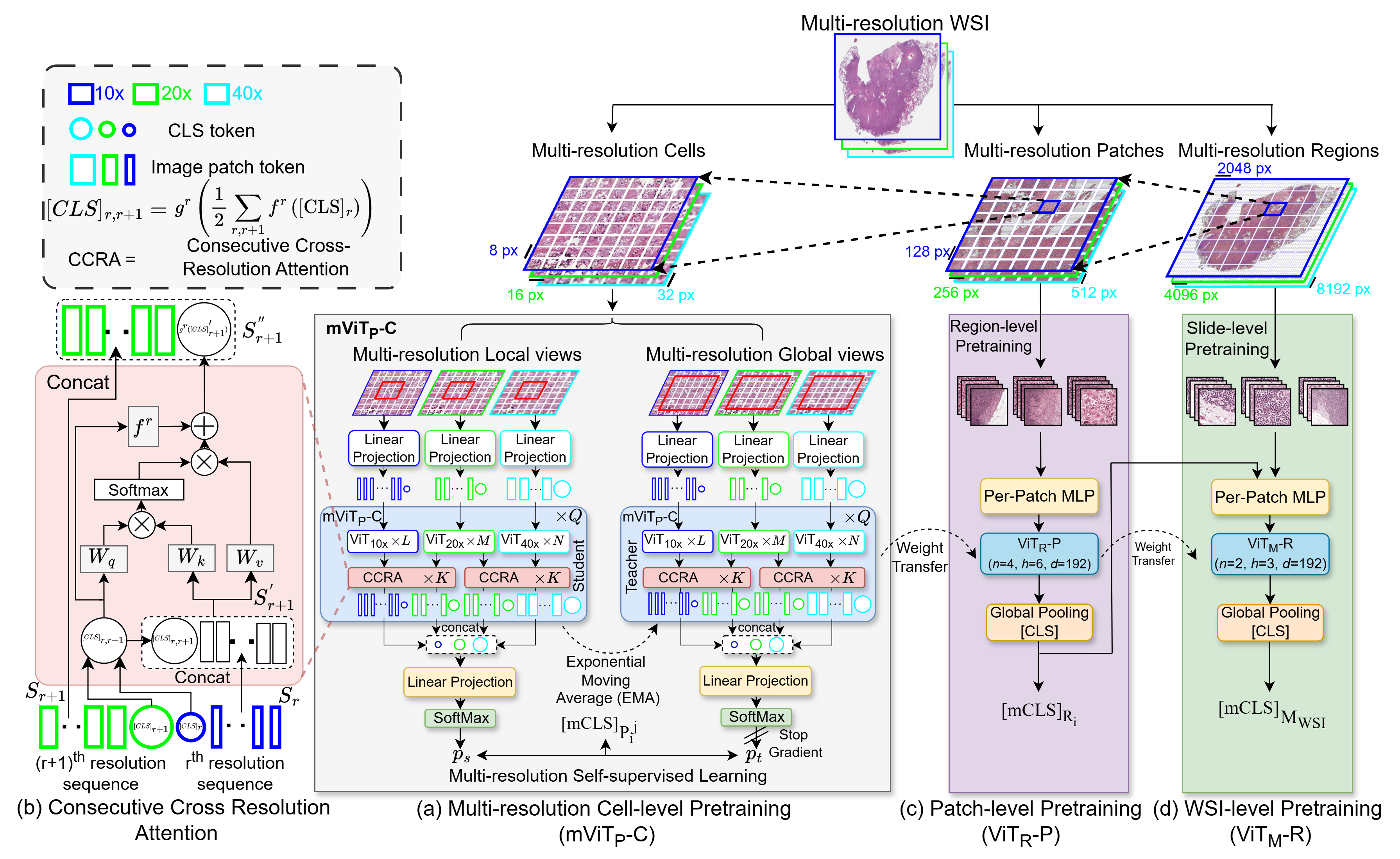}
    \vspace{-8mm}
   \caption{An overview of proposed MRPT architecture, which comprises three hierarchical pre-training stages: (a)-(b) cell-level ($\textrm{mViT}_{\textbf{P}}\textrm{-}\textrm{C}$), (c) patch-level ($\textrm{ViT}_{\textbf{R}}\textrm{-}\textrm{P}$), and (d) region-level ($\textrm{ViT}_{\textbf{M}}\textrm{-}\textrm{R}$). 
   The cell-level ViT, $\textrm{mViT}_{\textbf{P}}\textrm{-}\textrm{C}$, employs multi-resolution SSL paradigm. 
   The teacher and student networks learn multi-resolution representations using the proposed Consecutive Cross-Resolution Attention (CCRA) mechanism that fuses features across consecutive resolutions (10$\times$, 20$\times$, and 40$\times$). The learned multi-resolution [CLS] tokens are progressively aggregated from cell- to patch- to region-level through $\textrm{ViT}_{\textbf{R}}\textrm{-}\textrm{P}$ and $\textrm{ViT}_{\textbf{M}}\textrm{-}\textrm{R}$ to produce WSI-level embedding.}
   \label{fig3} 
\vspace{-2em}
%
\end{figure*}

\vspace{-2mm}
A schematic illustration of our proposed MRPT model is shown in Fig.~\ref{fig3}. 
The multi-resolution representations are learned in a hierarchical manner, progressing from the cell-level to the patch-level, then to the region-level, and finally to the WSI-level. 
At each level, multi-resolution SSL is employed to learn rich, fine-grained, and coarse-grained representations.
Our proposed MRPT-LLaVA is pre-trained in three different stages to address complex VQA tasks for WSI interpretation. 

\vspace{-3mm}
\subsection{Proposed MRPT Model}
Given an input multi-resolution WSI, $\textbf{M} = \{\textbf{X}_{\textrm{WSI}}~\textrm{at}~10\times,~\textbf{Y}_{\textrm{WSI}}~\textrm{at}~20\times,~\textbf{Z}_{\textrm{WSI}}~\textrm{at}~40\times\}$,
we extract $m$ non-overlapping regions each of size $4096 \times 4096$ at 20$\times$, denoted as $\{\textbf{Y}_{4096}^{i}\}_{i=1}^{m}$. 
For the corresponding 10$\times$ and 40$\times$ resolutions, we extract aligned regions of sizes $\{\textbf{X}_{2048}^{i}\}_{i=1}^{m}$ and $\{\textbf{Z}_{8192}^{i}\}_{i=1}^{m}$, respectively. 
These triplets of multi-resolution regions are represented as 
$\{\textbf{R}_{i}\}_{i=1}^{m} = \{\textbf{X}^{i}_{2048},~\textbf{Y}^{i}_{4096},~\textbf{Z}^{i}_{8192}\}_{i=1}^{m}$, 
where each $\textbf{R}_{i}$ corresponds to the same anatomical region across resolutions, preserving spatial alignment and contextual integrity. 
Each region $\textbf{R}_{i}$ is subdivided into smaller patches represented as 
$\{\textbf{P}_{i}^{j}\}_{j=1}^{256} = \{\textbf{X}^{j}_{128},~\textbf{Y}^{j}_{256},~\textbf{Z}^{j}_{512}\}_{j=1}^{256}$.
Each $\textbf{P}^{j}_{i}$ is further decomposed into multi-resolution cell-level patches denoted as 
$\{\textbf{C}^{j}_{i,k}\}_{k=1}^{256} = \{\textbf{X}^{k}_{8},~\textbf{Y}^{k}_{16},~\textbf{Z}^{k}_{32}\}_{k=1}^{256}$. 
This co-registration allows the model to jointly capture global tissue context from low-resolution views and localized morphological information from high-resolution ones. 
Our objective is to pre-train a large-scale, multi-resolution, hierarchical WSI-level foundation model enabling unified representation learning across all resolutions.

\vspace{-3mm}
\subsection{MRPT Architecture}
\label{sec:architecture}
Learning a robust multi-resolution WSI-level representation presents two main challenges: 
(1) a multi-resolution WSI contains a significantly larger sequence length of tokens compared to a single resolution, and 
(2) aggregating and aligning information across multi-resolution is non-trivial.  

To address the first challenge, we adopt a hierarchical WSI structure inspired by HIPT \cite{chen2022scaling}. 
We extend the ViT architecture for multi-resolution WSI representation learning by employing a nested aggregation of visual tokens that are recursively decomposed into smaller units. 
To address the second challenge, we propose a multi-resolution ViT that incorporates an efficient \textit{Consecutive Cross-Resolution Attention (CCRA)} mechanism to effectively integrate information across consecutive resolutions. Our proposed MRPT 
follows a multi-resolution hierarchical structure 
($\textbf{M} \rightarrow \{\textbf{R}_{i}\}_{i=1}^{m} \rightarrow \{\textbf{P}^{j}_{i}\}_{j=1}^{256} \rightarrow \{\textbf{C}^{j}_{i,k}\}_{k=1}^{256}$) (Fig. \ref{fig2}). 
At the lowest hierarchical level, a ViT module aggregates cell-level features across consecutive resolutions using the CCRA mechanism. 
This design enables MRPT to capture hierarchical and multi-resolution structures of WSIs and to model clinically meaningful relationships across resolutions.  

The pre-training of MRPT is performed in three distinct stages.  
In \textbf{Stage 1} (Fig. \ref{fig3} (a)-(b)), we pre-train cell-level features $\{\textbf{C}^{j}_{i,k}\}_{k=1}^{256}$ using multi-resolution SSL paradigm, where both teacher and student networks leverage a CCRA-enabled ViT architecture, referred to as $\textrm{mViT}_{\textbf{P}}\textrm{-}\textrm{C}$.
Here, $\textrm{C} \in \textbf{C}_{i,k}^{j}$ represents multi-resolution cell-level tokens extracted from patches $\{\textbf{P}_{i}^{j}\}_{j=1}^{256}$.
In \textbf{Stage 2} (Fig. \ref{fig3} (c)), the weights of $\textrm{mViT}_{\textbf{P}}\textrm{-}\textrm{C}$ are frozen and re-used as an embedding layer for patch-level pre-training using another ViT module, $\textrm{ViT}_{\textbf{R}}\textrm{-}\textrm{P}$, where $\textrm{P} \in \textbf{P}_{i}^{j}$ represents multi-resolution patch-level tokens.  
$\textrm{ViT}_{\textbf{R}}\textrm{-}\textrm{P}$ aggregates patch-level embeddings to learn region-level representations $\textbf{R}_{i}$.  
In \textbf{Stage 3} (Fig. \ref{fig3} (d)), the weights of $\textrm{ViT}_{\textbf{R}}\textrm{-}\textrm{P}$ are frozen and re-used as a backbone for WSI-level pre-training using $\textrm{ViT}_{\textbf{M}}\textrm{-}\textrm{R}$, where $\textrm{R} \in \{\textbf{R}_{i}\}_{i=1}^{m}$ denotes multi-resolution region-level tokens.  
$\textrm{ViT}_{\textbf{M}}\textrm{-}\textrm{R}$ aggregates the region-level embeddings to produce the final WSI-level representation.  
The overall mathematical formulation of MRPT is defined as:
\begin{equation}
\begin{split}
&{} \textrm{MRPT}(\textbf{M}) = \textrm{ViT}_{\textbf{M}}\textrm{-}\textrm{R}\left(\{\textrm{[mCLS]}_{\textbf{R}_{i}}\}_{i=1}^{m}\right), \textrm{where}\\
&{}\textrm{[mCLS]}_{\textbf{R}_{i}} = \textrm{ViT}_{\textbf{R}}\textrm{-}\textrm{P}\left(\{\textrm{[mCLS]}_{\textbf{P}^{j}_{i}}\}_{j=1}^{256}\right), \textrm{where} \\
&{}\textrm{[mCLS]}_{\textbf{P}^{j}_{i}} = \textrm{mViT}_{\textbf{P}}\textrm{-}\textrm{C}\left(\{\textbf{C}_{i,k}^{j}\}_{k=1}^{256}\right),
\end{split}
\label{eqn1}
\end{equation}
\noindent where $\textrm{[mCLS]}_{\textbf{P}^{j}_{i}}$ denotes the multi-resolution [CLS] token computed over patch $\textbf{P}^{j}_{i}$, and $\textrm{[mCLS]}_{\textbf{R}_{i}}$ denotes the multi-resolution [CLS] token at the region level $\textbf{R}_{i}$.  
Both $\textrm{mViT}_{\textbf{P}}\textrm{-}\textrm{C}$ and $\textrm{ViT}_{\textbf{R}}\textrm{-}\textrm{P}$ operate with a sequence length of 256 tokens, while $\textrm{ViT}_{\textbf{M}}\textrm{-}\textrm{R}$ processes approximately $m \approx 256$ region tokens per resolution.  
When small ViT backbones are employed at each stage, MRPT contains fewer than 12M parameters, making it computationally efficient and trainable on standard GPU workstations.  
Each pre-training stage of MRPT is detailed in the following sections.
\vspace{-3mm}
\subsubsection{Multi-Resolution Cell Aggregation
($\textrm{mViT}_{\textbf{P}}\textrm{-}\textrm{C}$)}
\label{sec:cellarchitecture}
The aggregation of multi-resolution cell-level tokens within $\textbf{P}_{i}^{j}$ is performed using the proposed $\textrm{mViT}_{\textbf{P}}\textrm{-}\textrm{C}$ (Figs.~\ref{fig3}(a)–(b)). 
This module employs the CCRA mechanism to integrate information across consecutive resolutions. 
To obtain the learnable embeddings $\textrm{[mCLS]}_{\textbf{P}_{i}^{j}}$, the cell-level representations across all resolutions are aggregated along the token sequence.
Multi-resolution SSL paradigm is employed in which local views are processed by the student network and global views are processed by the teacher.
Both teacher and student learn to align the multi-resolution representation.
Below, we discuss the general architecture of $\textrm{mViT}_{\textbf{P}}\textrm{-}\textrm{C}$, which is the same for both student and teacher.

Given $\textbf{P}_{i}^{j}$, $\textrm{mViT}_{\textbf{P}}\textrm{-}\textrm{C}$ decomposes it into $r$ sequences, each of length 256, corresponding to non-overlapping $c_{r} \times c_{r}$ pixel tokens at different resolutions $r \in \{10\times, 20\times, 40\times\}$, where $c_{r}$ denotes the cell-level patch size at $r$. 
For each resolution, a linear projection layer with positional embedding is applied to obtain a sequence embedding $\textbf{S}_{r}$, which includes an appended $[\textrm{CLS}]_{r}$ token. 
A transformer $\textrm{ViT}_{r}$ is then pre-trained independently to learn resolution-specific embeddings. 
The number of encoder layers in each $\textrm{ViT}_{r}$ is adjusted according to the resolution to balance computation. 
Finally, the outputs of $\textrm{ViT}_{r}$ from two consecutive resolutions are fused via the proposed CCRA mechanism to yield the multi-resolution representation.

\noindent \textbf{$\textrm{mViT}_{\textbf{P}}\textrm{-}\textrm{C}$ Architecture:} 
$\textrm{mViT}_{\textbf{P}}\textrm{-}\textrm{C}$ consists of a stack of $Q$ hierarchical fusion blocks. 
Each block contains three branches that process tokens from the three resolutions and fuse their outputs using the CCRA mechanism. 
As illustrated in Fig.~\ref{fig3} (a), the architecture is composed of $Q$ multi-resolution transformer encoders, where each encoder comprises:  
(1) a \textit{10$\times$ branch} handling coarse-grained features ($8 \times 8$ patch size) with $L$ transformer layers,  
(2) a \textit{20$\times$ branch} processing medium-grained features ($16 \times 16$ patch size) with $M$ layers, and  
(3) a \textit{40$\times$ branch} focusing on fine-grained features ($32 \times 32$ patch size) with $N$ layers.  
The three branches are fused $Q$ times through CCRA layers that enable the $[\textrm{CLS}]_{r}$ tokens of all branches to exchange contextual information across resolutions.

\noindent \textbf{Consecutive Cross-Resolution Attention (CCRA):} 
The proposed CCRA module (Fig.~\ref{fig3}(b)) fuses information between two consecutive resolutions by averaging their $[\textrm{CLS}]_{r}$ tokens and appending them to the patch tokens of the finer sequence. 
This design allows efficient and interpretable cross-resolution information exchange.  
Specifically, the $[\textrm{CLS}]_{r}$ tokens from both sequences serve as information carriers between resolutions and are back-projected to their respective branches. 
Because $[\textrm{CLS}]_{r}$ tokens encode abstract-level semantics of their sequences, interacting with patch tokens of other resolutions enriches both coarse- and fine-grained contextual representations. 
The process continues across CCRA layers, where $[\textrm{CLS}]_{r}$ tokens iteratively refine their own sequence representations, thereby enhancing semantic alignment across resolutions.  
\textit{This mechanism efficiently fuses coarse and fine features at the cell level while maintaining semantic consistency.}

While performing cross-attention between consecutive resolutions, we preserve the hierarchical diagnostic logic used by pathologists:  
10$\times$ magnification provides a global overview of tissue architecture,  
20$\times$ reveals local glandular and stromal organization, and  
40$\times$ exposes nuclear and cellular morphology.  
\textit{Pathologists examine WSIs sequentially across resolutions rather than skipping scales, as intermediate resolutions contextualize cellular detail within broader architecture.}  
Inspired by this workflow, MRPT restricts cross-attention to consecutive resolution pairs (10$\times \leftrightarrow 20\times$, 20$\times \leftrightarrow 40\times$), ensuring semantic continuity and minimizing misalignment across distant representations.  
This design produces biologically interpretable and diagnostically meaningful embeddings (\textit{see ablation}). 

\noindent \textbf{Class Token Fusion:} 
The $[\textrm{CLS}]_{r}$ tokens represent global contextual summaries of their corresponding sequences and are used as final embeddings for downstream tasks \cite{chen2024towards,xu2024whole}. 
We first average the $[\textrm{CLS}]_{r}$ tokens of two consecutive resolutions (Fig.~\ref{fig3}(d)) and append the averaged token to both sequences, enabling bidirectional feature exchange. 
Formally, the fused sequence $\mathbf{S}_{r+1}^{'}$ is given by:
\begin{equation}
\mathbf{S}_{r+1}^{'}=  \left[  g^r \Big(\frac{1}{2} \sum_{r, r+1} f^r([\textrm{CLS}]_{r})\Big)\| \textbf{S}_{r} \right], 
\label{eqn2}
\end{equation}
\noindent where $f^r(\cdot)$ and $g^r(\cdot)$ denote projection and back-projection functions for dimensional alignment.

\noindent \textbf{Multi-Resolution Fusion Using CCRA:} 
We then perform CCRA between the averaged $[\textrm{CLS}]_{r,r+1}$ tokens and $\mathbf{S}_{r+1}^{'}$. 
The query corresponds to $[\textrm{CLS}]_{r,r+1}$, while keys and values correspond to $\mathbf{S}_{r+1}^{'}$, expressed as:
\begin{equation*}
\begin{split}
&{} \mathbf{q} = [\textrm{CLS}]_{r,r+1} \mathbf{W}_{q}, \quad 
\mathbf{k} = \mathbf{S}_{r+1}^{'} \mathbf{W}_{k}, \quad 
\mathbf{v} = \mathbf{S}_{r+1}^{'} \mathbf{W}_{v}, \\ 
&{} \mathbf{A} = \mathrm{softmax}\!\left(\frac{\mathbf{q}\mathbf{k}^{T}}{\sqrt{d}}\right), 
\quad \textrm{CCRA}(\mathbf{S}_{r+1}^{'}) = \mathbf{A}\mathbf{v},~~~~~~~~(3)
\end{split}
\label{eqn3}
\end{equation*}
\noindent where $\mathbf{W}_{q}$, $\mathbf{W}_{k}$, and $\mathbf{W}_{v}$ are learnable projection matrices, and $d$ is the embedding dimension.  
\textit{Because the query consists of a single vector $[\textrm{CLS}]_{r,r+1}$, the computational complexity of CCRA is linear rather than quadratic, making it highly efficient.}  
We further employ multi-head CCRA (MHCCRA) for richer feature fusion.  
The updated sequence $\mathbf{S}_{r+1}^{''}$ for a given $\mathbf{S}_{r+1}$ is obtained with Layer Normalization (LN) and residual connections as:
\begin{equation*}
\begin{split}
[\textrm{CLS}]_{r+1}^{'} &= f^{r}\!\left([\textrm{CLS}]_{r,r+1}\right) 
+ \textrm{MHCCRA}\!\left(\textrm{LN}(\textbf{S}_{r+1}^{'})\right), \\
\mathbf{S}_{r+1}^{''} &= 
\left[ 
    g^{r}\!\left([\textrm{CLS}]_{r+1}^{'} \right) \;\Vert\; \mathbf{S}_{r+1} 
\right].~~~~~~~~~~~~~~~~~~~~~~~~~(4)
\end{split}
\label{eqn4}
\end{equation*}
\noindent Empirical results confirm that both class-token fusion and CCRA are critical for efficient and effective multi-resolution feature learning (\textit{Table \ref{table3})}.  
For intermediate resolution (20$\times$), two sequences are obtained: $\mathbf{S}_{10\times-20\times}^{''}$ and $\mathbf{S}_{20\times-40\times}^{''}$, whose corresponding $[\textrm{CLS}]$ tokens ($[\textrm{CLS}]_{10\times-20\times}^{'}$, $[\textrm{CLS}]_{20\times-40\times}^{'}$) are concatenated and projected back to their original dimension to form $[\textrm{CLS}]_{20\times}^{'}$.  

There are $K$ stacked CCRA layers in total. 
At each layer, fine-grained features from $40\times$ are propagated to $20\times$, then to $10\times$, and vice versa.  
The outputs of the final CCRA layer yield three sequences: $\mathbf{S}_{10\times}^{''}$, $\mathbf{S}_{20\times}^{''}$, and $\mathbf{S}_{40\times}^{''}$.  
The three $[\textrm{CLS}]$ tokens—$[\textrm{CLS}]_{10\times}^{'}$, $[\textrm{CLS}]_{20\times}^{'}$, and $[\textrm{CLS}]_{40\times}^{'}$—are concatenated and passed through a projection layer to form multi-resolution patch representation $\textrm{[mCLS]}_{\textbf{P}_{i}^{j}}$.
\vspace{-4mm}
\subsubsection{{Multi-Resolution Patch Aggregation ($\textrm{ViT}_{\textbf{R}}\textrm{-}\textrm{P}$)}} 
To obtain multi-resolution region-level embedding $\textrm{[mCLS]}_{\textbf{R}_{i}}$, we aggregate the sequence $\{\textrm{[mCLS]}_{\textbf{P}_{i}^{j}}\}_{j=1}^{256}$, derived via $\textrm{mViT}_{\mathbf{P}}\textrm{-C}$, by appending a $[\textrm{CLS}]$ token and feeding it into $\textrm{ViT}_{\mathbf{R}}\textrm{-P}$ architecture, to model broader tissue-level contextual relationships (Fig.~\ref{fig3} (c)). 
\vspace{-3mm}
\subsubsection{{Multi-Resolution Region Aggregation ($\textrm{ViT}_{\textbf{M}}\textrm{-}\textrm{R}$)}} 
To obtain the multi-resolution WSI-level representation $\textrm{[mCLS]}_{\textbf{M}_{\text{WSI}}}$, we aggregate sequence $\{\textrm{[mCLS]}_{\textbf{R}_{i}}\}_{i=1}^{m}$, derived via $\textrm{ViT}_{\mathbf{R}}\textrm{-P}$, by appending a $[\textrm{CLS}]$ token and feeding it into $\textrm{ViT}_{\textbf{M}}\textrm{-}\textrm{R}$ to model holistic representation (Fig.~\ref{fig3} (c)). 

\vspace{-4mm}
\subsubsection{{Hierarchical Pre-training of MRPT Model}} 
\label{sec:hierarchical}
\noindent \textbf{Stage 1: Multi-Resolution Patch-Level ($\textbf{P}_{i}^{j}$) Pre-training.} 
We pre-train $\textrm{mViT}_{\textbf{P}}\textrm{-}\textrm{C}$ using the multi-resolution SSL framework, wherein a student network learns to match the output distribution of a teacher network  via cross-entropy loss: $H = -\sum_{i,j \in \{10\times, 20\times, 40\times\}} p_{t_{i}} \log p_{s_{j}}$, where $p_{t_{i}}$ and $p_{s_{j}}$ represent the outputs of teacher and student for resolutions $i$ and $j$, respectively. 
For each multi-resolution patch $\textbf{P}_{i}^{j}$, DINO generates a set of $L_{r}=8$ local views (of sizes $48\times48$, $96\times96$, and $192\times192$ for 10$\times$, 20$\times$, and 40$\times$, respectively) input to $\Psi_{s}$, and $G_{r}=2$ global views (of sizes $112\times112$, $224\times224$, and $448\times448$) input to $\Psi_{t}$. 
This encourages local-to-global consistency across resolutions, minimizing:
\vspace{-3mm}
\begin{equation}
\nonumber
\min_{\Psi_{s}} 
\sum_{r}
\sum_{G_{r}}
\sum_{L_{r}}
H \big( p_{t_{r}}(G_{r}), p_{s_{r}}(L_{r}) \big).~~~~~~~~~~~(6)
\label{eqn6}
\vspace{-3mm}
\end{equation}

\noindent \textbf{Stage 2: Multi-Resolution Region-Level ($\textbf{R}_{i}$) Pre-training.} 
In this SSL stage,
$[\textrm{mCLS}]_{\textbf{P}_{i}^{j}}$ from $\textrm{mViT}_{\textbf{P}}\textrm{-}\textrm{C}$ serve as inputs to $\textrm{ViT}_{\textbf{R}}\textrm{-}\textrm{P}$.  
The resulting $\{[\textrm{mCLS}]_{\textbf{P}_{i}^{j}}\}_{j=1}^{256}$ are rearranged into a $16\times16\times384$ feature grid, on which local-global patch-level crops of sizes $[6\times6]$ and $[14\times14]$ are applied to form augmented views for contrastive pre-training.

\noindent \textbf{Stage 3: Multi-Resolution WSI-Level Pre-training.} 
In this stage, both $\textrm{mViT}_{\textbf{P}}\textrm{-}\textrm{C}$ and $\textrm{ViT}_{\textbf{R}}\textrm{-}\textrm{P}$ are frozen and $[\textrm{mCLS}]_{\textbf{R}_{i}}$ from $\textrm{ViT}_{\textbf{R}}\textrm{-}\textrm{P}$ are input to $\textrm{ViT}_{\textbf{M}}\textrm{-}\textrm{R}$.
The resulting $\{\textrm{[mCLS]}_{\textbf{R}_{i}}\}_{i=1}^{m}$ are rearranged into a $\sqrt{m}\times\sqrt{m}\times192$ feature grid (with $m \approx 64$ for most WSIs).  
Local and global crops of sizes $[6 \times 6]$ and $[7 \times 7]$ are used in SSL.
This stage enables the model to capture holistic WSI-level contextual dependencies while preserving fine-grained semantic consistency across resolutions.
\textit{We assess the contribution of each stage through ablation studies (\textit{see Table~\ref{table2}}).}
\vspace{-2mm}
\subsection{{MRPT-LLaVA Model Pre-Training}}
\label{sec:llava}
\textbf{MRPT-LLaVA} follows a three-stage training pipeline~\cite{chen2025slidechat, liang2025wsillavamultimodallargelanguage, chen2025wsi, seyfioglu2024quilt}: \textit{cross-modal alignment},  \textit{feature space alignment}, and \textit{instruction tuning}.
In Stage I, 9,642 WSI–report pairs from WSI-Bench~\cite{liang2025wsillavamultimodallargelanguage} are aligned via contrastive learning using the MRPT encoder and the Qwen2-1.5B text encoder~\cite{yang2024qwen2}.
In Stage II, the MRPT encoder and LLM are frozen while the projection layer is updated.
In Stage III, we fine-tune the projection layer and LLM on all WSI-Bench tasks.
For patch-level VQA, $\textrm{mViT}_{\textbf{P}}\textrm{-}\textrm{C}$ and $\textrm{ViT}_{\textbf{R}}\textrm{-}\textrm{P}$ backbones are trained using the aforementioned approach on QuiltNet-1M~\cite{ikezogwo2024quilt} and QuiltInstruct~\cite{seyfioglu2024quilt} datasets.
\vspace{-3mm}

\section{Experiments}
\label{sec:results}
\vspace{-2mm}

\begin{table}[t]
\caption{Configuration of the $\textrm{mViT}_{\textbf{P}}\textrm{-}\textrm{C}$ architecture across Tiny (T), Small (S), and Base (B) variants. 
Each model processes three resolutions (10$\times$, 20$\times$, 40$\times$) with corresponding embedding dimensions, number of attention heads $(h_L, h_M, h_N)$, and encoder depths $(L, M, N)$. 
For all variants, we fix CCRA layers to $K = 1$ and hierarchical fusion blocks to $Q = 4$.}
\vspace{-1em}
\centering
\makebox[\linewidth]{
\scalebox{0.80}{
\begin{tabular}{l|c c c|c c}
\hline
Variants &  \multicolumn{3}{c|}{Dimension} & \# heads & Encoders \\
 &  $10 \times$ & $20 \times$ & $40 \times$ & $h_L,h_M,h_N$ &$L, M, N$ \\
\hline
$\textrm{mViT}_{\textbf{\textrm{P}}}\textrm{-}\textrm{C}$-T   & 96 & 192 & 192 &  3,3,3 & 2,4, 4 \\
$\textrm{mViT}_{\textbf{\textrm{P}}}\textrm{-}\textrm{C}$-S   &192 & 384 & 384 &  6,6,6 & 2,4,4  \\
$\textrm{mViT}_{\textbf{\textrm{P}}}\textrm{-}\textrm{C}$-B   &384 & 768 & 768 &  12,12,12 & 2,4,4 \\
\hline
\end{tabular}
}
}
\label{table1}
\vspace{-1em}
\end{table}

\begin{table*}[!t]
   \centering 
\caption{Performance of different hierarchical levels of the proposed MRPT model.  Each variant combines the patch-level encoder $\textrm{mViT}_{\textbf{P}}\textrm{-}\textrm{C}$ with the region-level encoder $\textrm{ViT}_{\textbf{R}}\textrm{-}\textrm{P}$ and the WSI-level encoder $\textrm{ViT}_{\textbf{M}}\textrm{-}\textrm{R}$. ‘PF’ denotes \textit{Pre-trained and Frozen}, and ‘LP’ indicates \textit{Linear Probe}. “Params” represent the number of learnable parameters during downstream fine-tuning task. Results are reported across three WSI benchmark datasets, comparing MRPT hierarchical variants (A–D) against the HIPT baseline~\cite{chen2022scaling}.}
\centering
\scalebox{0.75}{
\begin{tabular}{llccccc}
\hline
Experiments&Models& Params& PANDA &BRAINS&UBC-OCEAN \\
\hline
\multirow{4}{*}{A}&HIPT ($\textrm{ViT}_{256}\textrm{-}16_{\textrm{PF}}$) \cite{chen2022scaling}&0.505M&0.584&0.502&0.677\\
&$\textrm{mViT}_{\textbf{\textrm{P}}}\textrm{-}\textrm{C}\textrm{-}\textrm{T}_{\textrm{PF}}$&0.505M&0.732&0.687&0.779\\
&$\textrm{mViT}_{\textbf{\textrm{P}}}\textrm{-}\textrm{C}\textrm{-}\textrm{S}_{\textrm{PF}}$&0.505M&\underline{0.754}&\underline{0.705}&\underline{0.781}\\
&$\textrm{mViT}_{\textbf{\textrm{P}}}\textrm{-}\textrm{C}\textrm{-}\textrm{B}_{\textrm{PF}}$&0.505M&\textbf{0.773}&\textbf{0.723}&\textbf{0.811}\\
\hline
\multirow{4}{*}{B}&HIPT ($\textrm{ViT}_{256}\textrm{-}16_{\textrm{PF}}$, $\textrm{ViT}_{4096}\textrm{-}256_{\textrm{PF}}$) \cite{chen2022scaling}&0.505M&0.623&0.567&0.723 \\
&$\textrm{mViT}_{\textbf{\textrm{P}}}\textrm{-}\textrm{C}\textrm{-}\textrm{T}_{\textrm{PF}}$, $\textrm{ViT}_{\textbf{\textrm{R}}}\textrm{-}\textrm{P}_{\textrm{PF}}$&0.505M&\underline{0.773}&0.704&\underline{0.855}\\
&$\textrm{mViT}_{\textbf{\textrm{P}}}\textrm{-}\textrm{C}\textrm{-}\textrm{S}_{\textrm{PF}}$, $\textrm{ViT}_{\textbf{\textrm{R}}}\textrm{-}\textrm{P}_{\textrm{PF}}$&0.505M&0.768&\underline{0.712}&0.842\\
&$\textrm{mViT}_{\textbf{\textrm{P}}}\textrm{-}\textrm{C}\textrm{-}\textrm{B}_{\textrm{PF}}$, $\textrm{ViT}_{\textbf{\textrm{R}}}\textrm{-}\textrm{P}_{\textrm{PF}}$&0.505M&\textbf{0.822}&\textbf{0.772}&\textbf{0.926}\\
\hline
\multirow{4}{*}{C}&HIPT ($\textrm{ViT}_{256}\textrm{-}16_{\textrm{PF}}$, $\textrm{ViT}_{4096}\textrm{-}256_{\textrm{PF}}$, $\textrm{ViT}_{\textrm{WSI}}\textrm{-}4096$) \cite{chen2022scaling}&1.47M&0.576&0.514&0.653\\
&$\textrm{mViT}_{\textbf{\textrm{P}}}\textrm{-}\textrm{C}\textrm{-}\textrm{T}_{\textrm{PF}}$, $\textrm{ViT}_{\textbf{\textrm{R}}}\textrm{-}\textrm{P}_{\textrm{PF}}$, $\textrm{ViT}_{\textbf{\textrm{M}}}\textrm{-}\textrm{R}$ &1.47M&0.754&0.687&0.821 \\
&$\textrm{mViT}_{\textbf{\textrm{P}}}\textrm{-}\textrm{C}\textrm{-}\textrm{S}_{\textrm{PF}}$, $\textrm{ViT}_{\textbf{\textrm{R}}}\textrm{-}\textrm{P}_{\textrm{PF}}$, $\textrm{ViT}_{\textbf{\textrm{M}}}\textrm{-}\textrm{R}$ &1.47M&\underline{0.767}&\underline{0.706}&\underline{0.827}\\
&$\textrm{mViT}_{\textbf{\textrm{P}}}\textrm{-}\textrm{C}\textrm{-}\textrm{B}_{\textrm{PF}}$, $\textrm{ViT}_{\textbf{\textrm{R}}}\textrm{-}\textrm{P}_{\textrm{PF}}$, $\textrm{ViT}_{\textbf{\textrm{M}}}\textrm{-}\textrm{R}$ &1.47M&\textbf{0.807}&\textbf{0.740}&\textbf{0.892} \\
\hline
\multirow{3}{*}{D}&$\textrm{mViT}_{\textbf{\textrm{P}}}\textrm{-}\textrm{C}\textrm{-}\textrm{T}_{\textrm{PF}}$, $\textrm{ViT}_{\textbf{\textrm{R}}}\textrm{-}\textrm{P}_{\textrm{PF}}$, $\textrm{ViT}_{\textbf{\textrm{M}}}\textrm{-}\textrm{R}_{\textrm{PF}}+\textrm{LP}$&1.5K&0.786&\underline{0.728}&0.854 \\
&$\textrm{mViT}_{\textbf{\textrm{P}}}\textrm{-}\textrm{C}\textrm{-}\textrm{S}_{\textrm{PF}}$, $\textrm{ViT}_{\textbf{\textrm{R}}}\textrm{-}\textrm{P}_{\textrm{PF}}$, $\textrm{ViT}_{\textbf{\textrm{M}}}\textrm{-}\textrm{R}_{\textrm{PF}}+\textrm{LP}$ &1.5K&\underline{0.803}&0.720&\underline{0.862} \\
&$\textrm{mViT}_{\textbf{\textrm{P}}}\textrm{-}\textrm{C}\textrm{-}\textrm{B}_{\textrm{PF}}$, $\textrm{ViT}_{\textbf{\textrm{R}}}\textrm{-}\textrm{P}_{\textrm{PF}}$, $\textrm{ViT}_{\textbf{\textrm{M}}}\textrm{-}\textrm{R}_{\textrm{PF}}+\textrm{LP}$ &1.5K&\textbf{0.866}&\textbf{0.813}&\textbf{0.910} \\
\hline
\end{tabular}%
}
  \hfill
\label{table2}
 \vspace{-1em}
\end{table*}

 \begin{figure*}[!t]
\centering 
 \begin{subfigure}[t]{0.5\textwidth} 
 \centering 
 \includegraphics[width=\textwidth, height=3cm]{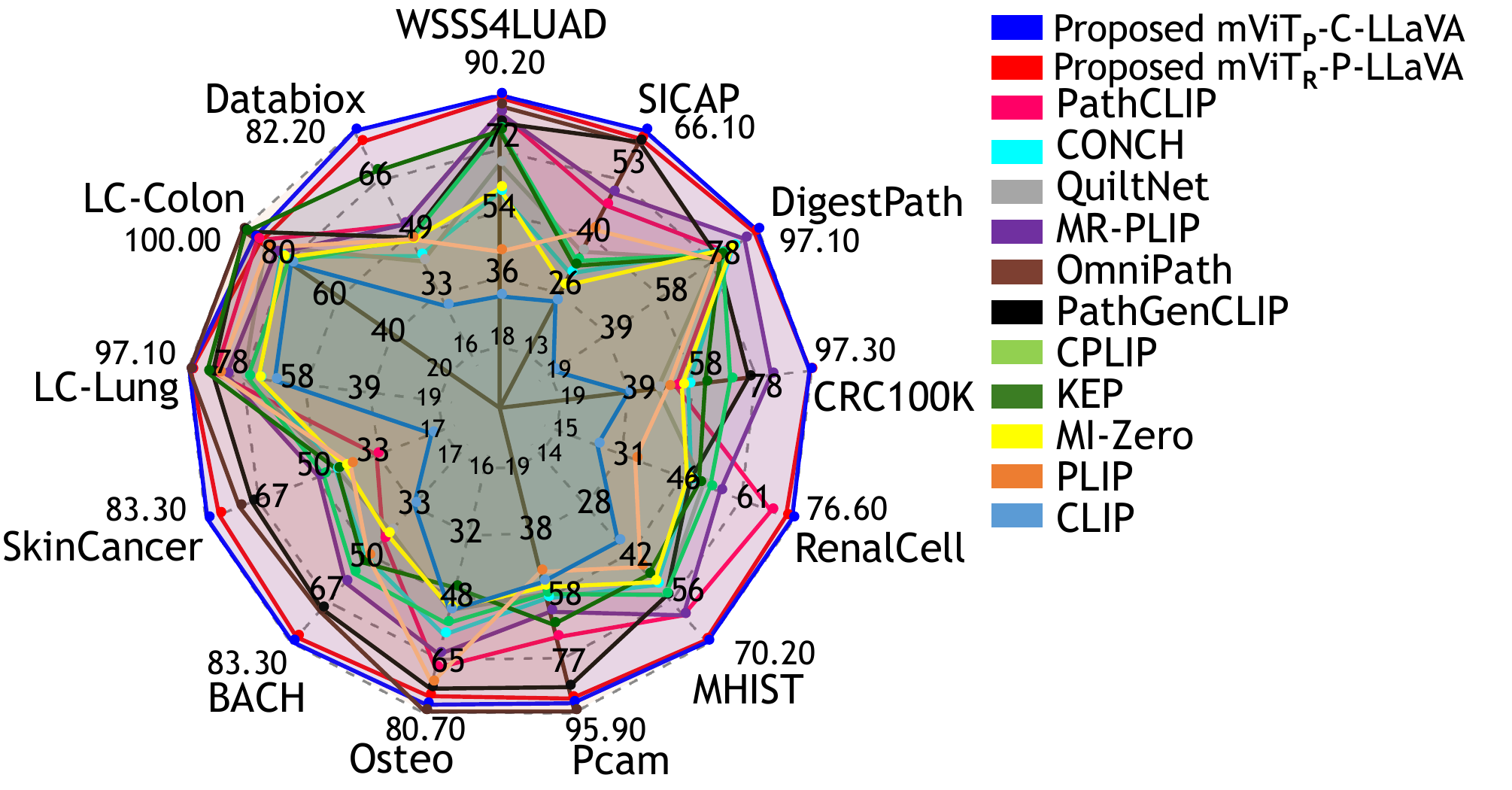} 
  \caption{Patch-level Zero-Shot Performance (A)} 
  \label{fig:patch_a} 
 \end{subfigure}%
 \hfill 
 \begin{subfigure}[t]{0.5\textwidth} 
 \centering 
 \includegraphics[width=\textwidth, height=3cm]{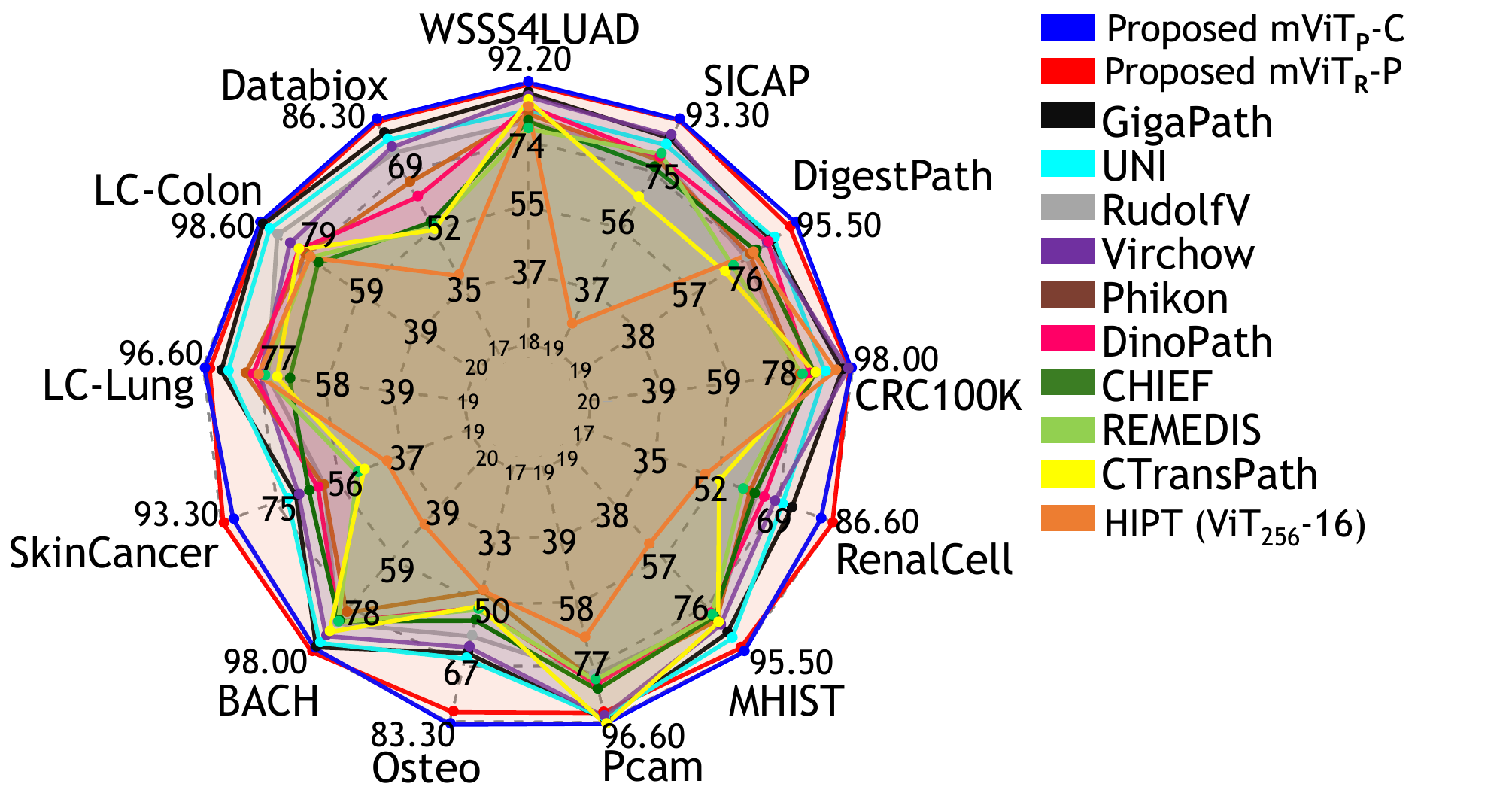} 
 \caption{Patch-level Linear Probe Evaluation (A)} 
 \label{fig:patch_b} 
 \end{subfigure} 
  \hfill 
 \begin{subfigure}[t]{0.5\textwidth} 
 \centering 
 \includegraphics[width=\textwidth, height=3cm]{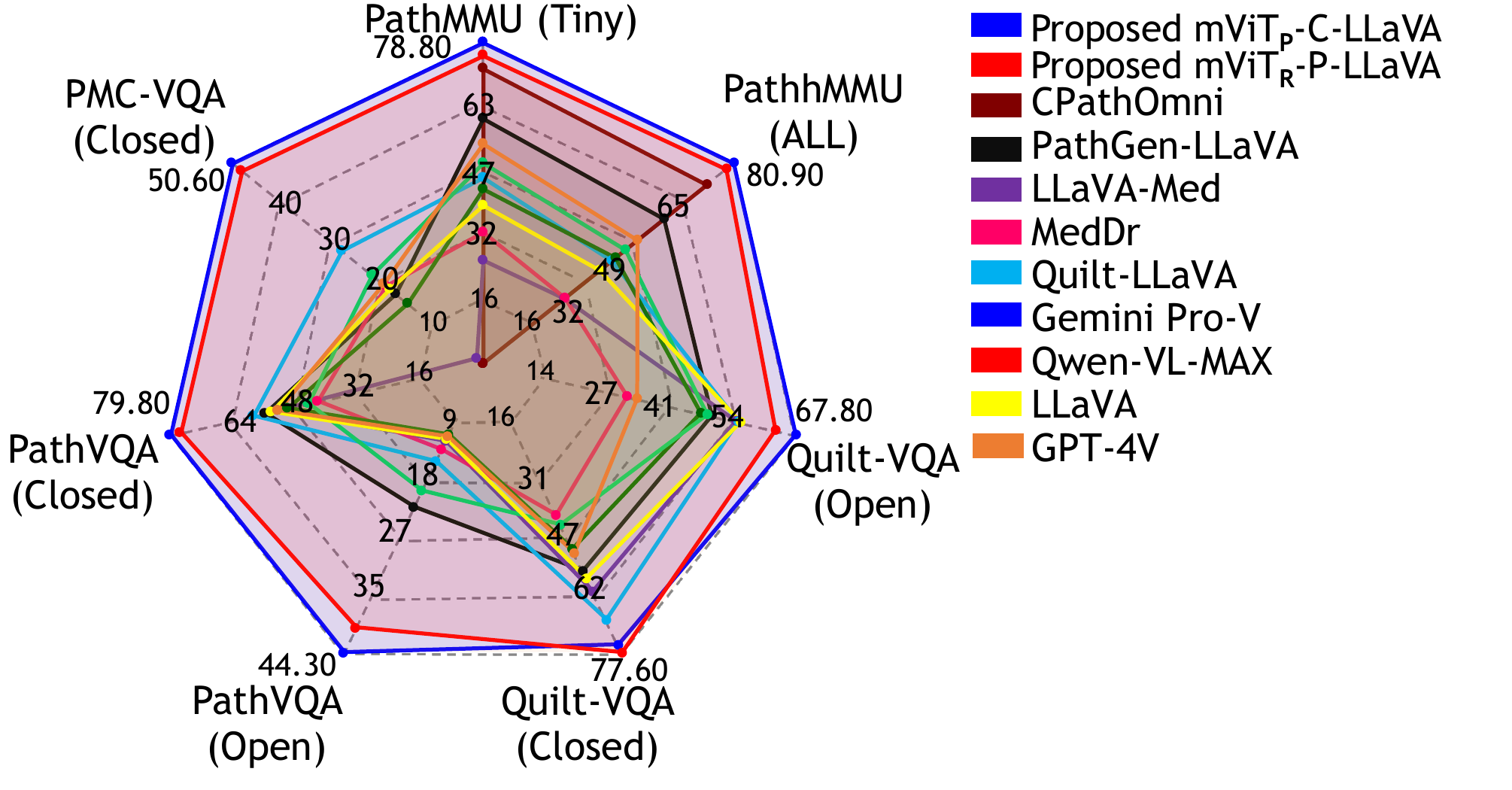} 
 \caption{Patch-level VQA (Accuracy)} 
 \label{fig:patch_c} 
 \end{subfigure} 
   \hfill 
 \begin{subfigure}[t]{0.45\textwidth} 
 \centering 
 \includegraphics[width=\textwidth, height=3cm]
 {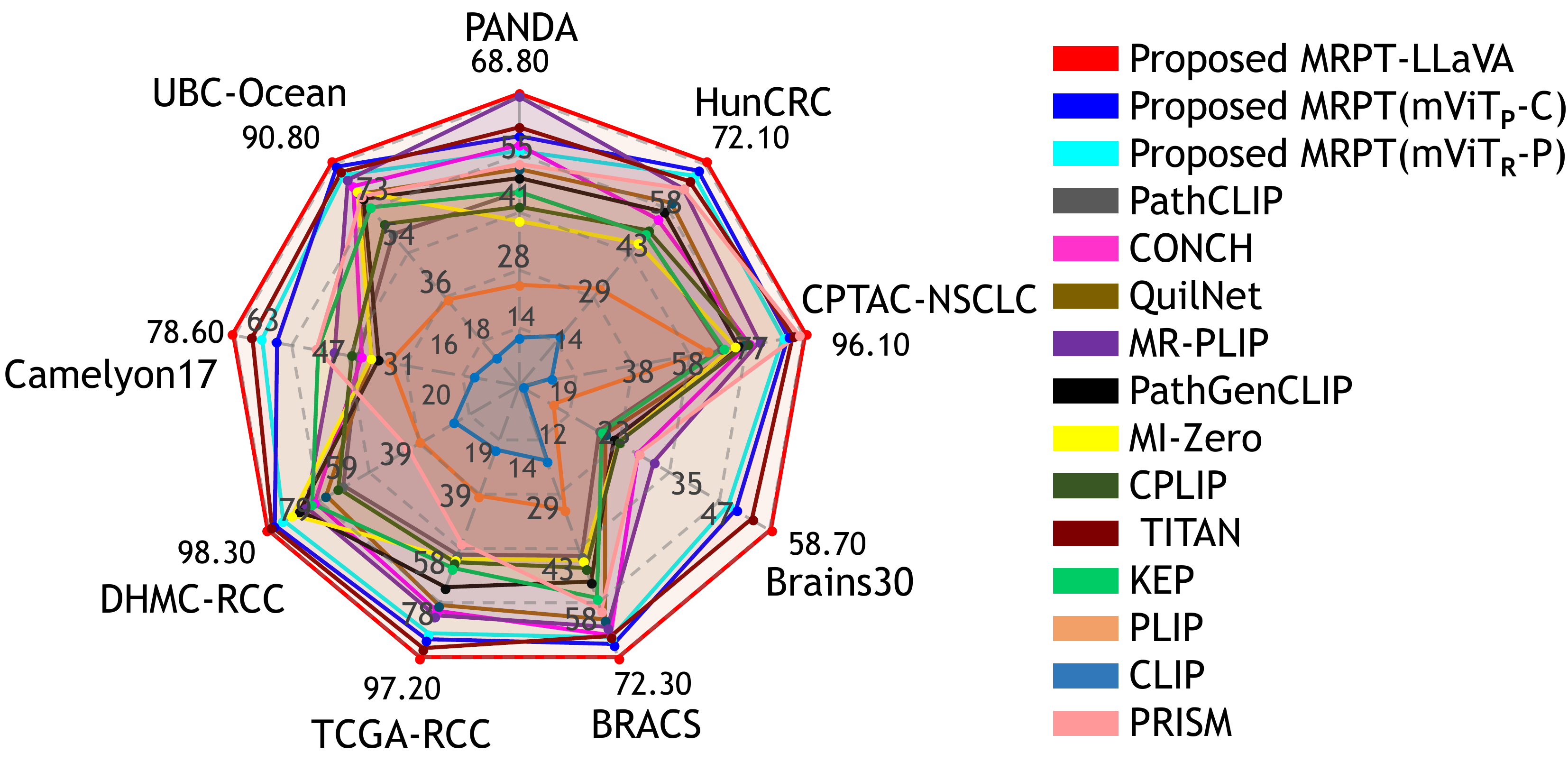} 
 \caption{WSI-level Zero-Shot Performance (BA)} 
 \label{fig:patch_d} 
 \end{subfigure} 
   \hfill 
 \begin{subfigure}[t]{0.45\textwidth} 
 \centering 
  \includegraphics[width=\textwidth, height=3cm]
 {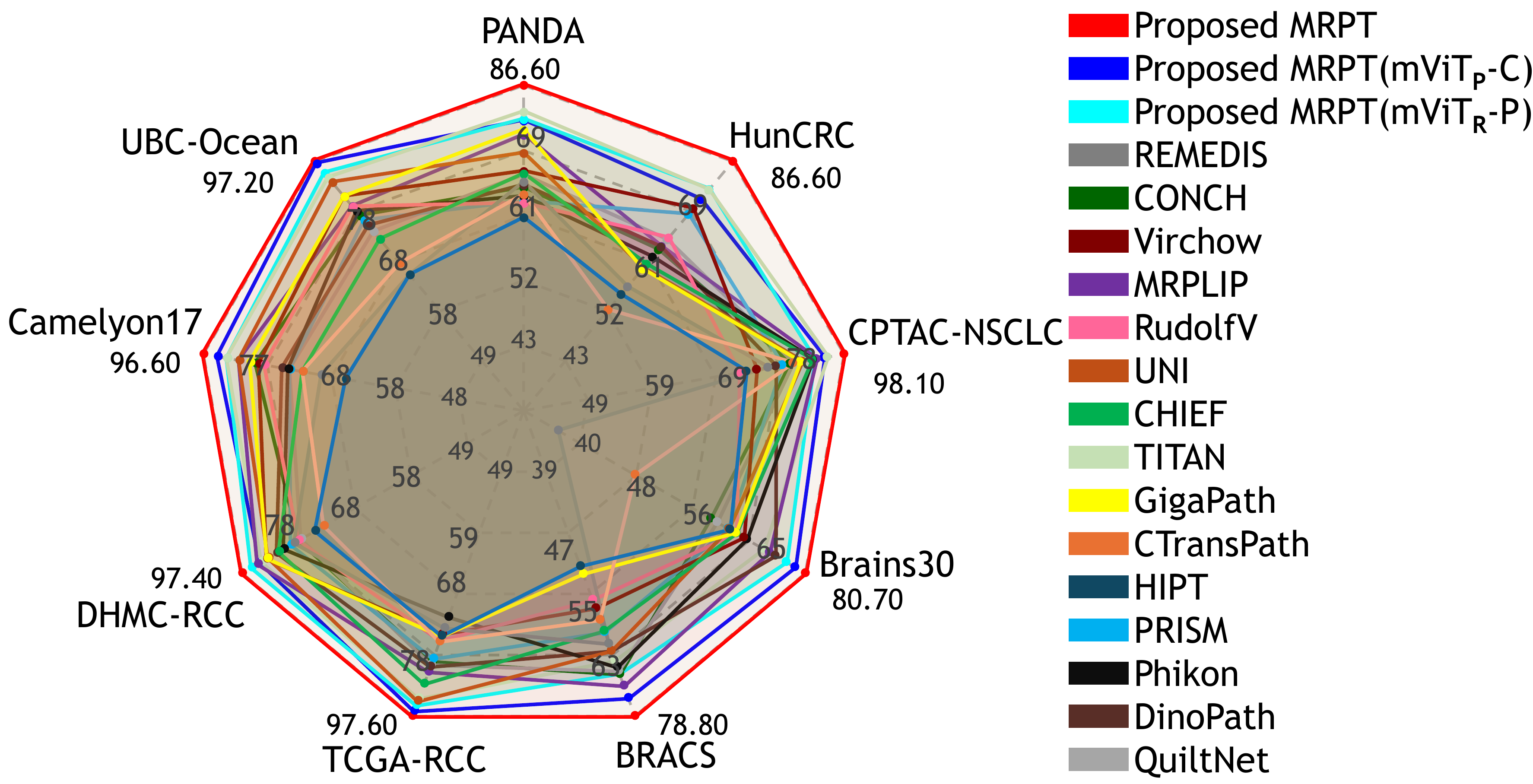}
 \caption{WSI-level Linear Probe and WSMIL (BA)} 
 \label{fig:patch_e} 
 \end{subfigure} 
   \hfill 
 \begin{subfigure}[t]{0.45\textwidth} 
 \centering 
 \includegraphics[width=\textwidth, height=3cm]{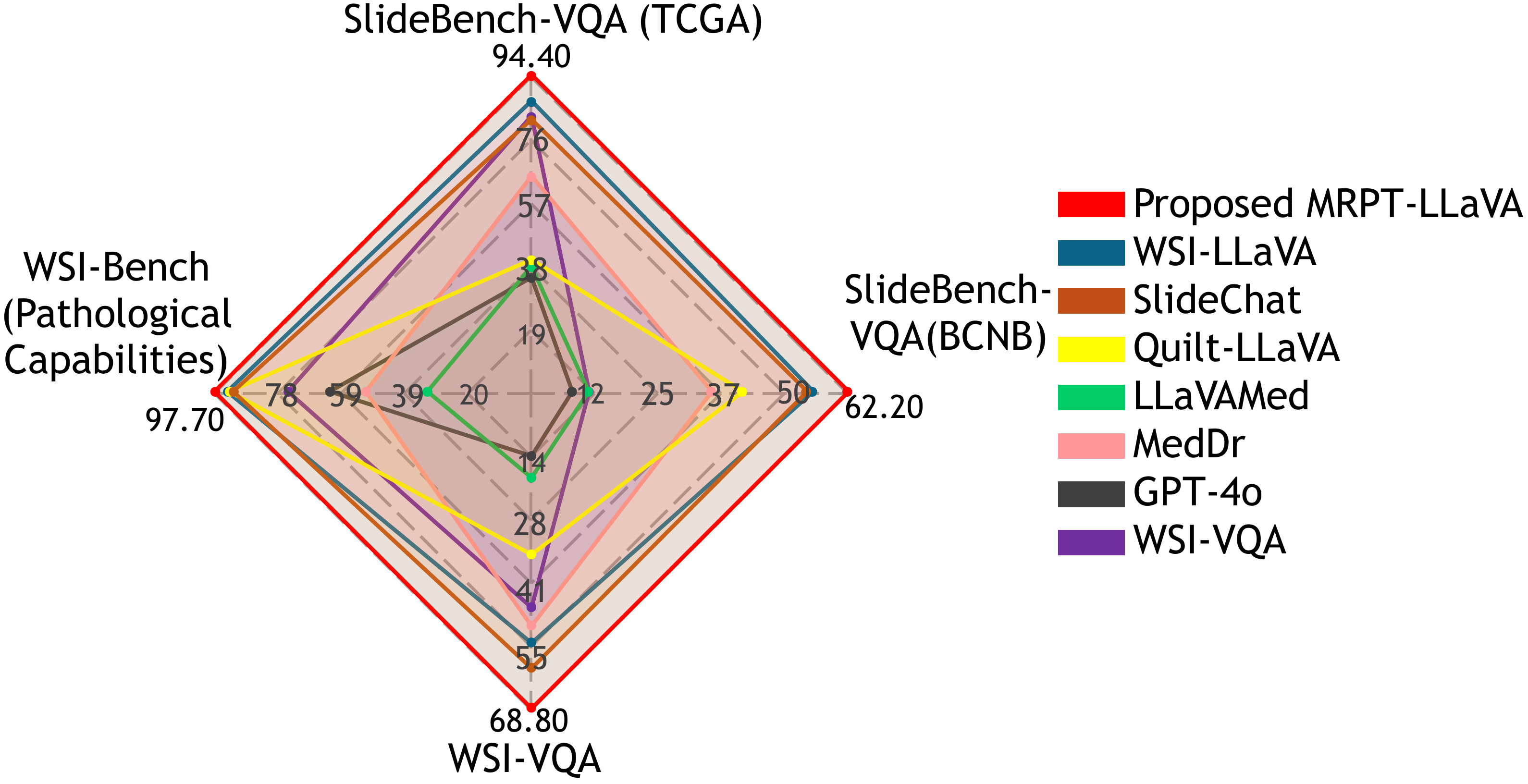} 
 \caption{WSI-level VQA (Accuracy)} 
 \label{fig:patch_f} 
 \end{subfigure} 
\caption{Proposed $\textrm{mViT}_{\textbf{P}}\textrm{-C}$-LLaVA, $\textrm{mViT}_{\textbf{R}}\textrm{-P}$-LLaVA, $\textrm{mViT}_{\textbf{P}}\textrm{-C}$, $\textrm{mViT}_{\textbf{R}}\textrm{-P}$, MRPT, and MRPT-LLaVA  outperform SOTA models. 
} 
 \label{fig:radar} 
 \vspace{-1em}
 \end{figure*}

\textbf{Training Details:} We pre-train \textbf{MRPT} on 36{,}000 WSIs in total, comprising 30{,}000 WSIs from diverse TCGA cancer types \cite{hutter2018cancer} and 6{,}000 WSIs from CPTAC \cite{Edwards2015CPTAC}. 
\textbf{MRPT} is pre-trained on 624M patches using $\textrm{mViT}_{\textbf{P}}\textrm{-}\textrm{C}$, 2.4M regions using $\textrm{ViT}_{\textbf{R}}\textrm{-}\textrm{P}$, and 36K WSIs using $\textrm{ViT}_{\textbf{M}}\textrm{-}\textrm{R}$. 
For $\textrm{mViT}_{\textbf{P}}\textrm{-}\textrm{C}$, we employ three resolution-specific ViTs with $L=2$, $M=4$, and $N=4$ encoder layers (see Table~\ref{table1}), using an embedding dimension of 384 for $\textrm{ViT}_{10\times}$ and 768 for both $\textrm{ViT}_{20\times}$ and $\textrm{ViT}_{40\times}$. 
We use $K=1$ CCRA layer and a total of $Q=4$ blocks in $\textrm{mViT}_{\textbf{P}}\textrm{-}\textrm{C}$. 
For $\textrm{ViT}_{\textbf{R}}\textrm{-}\textrm{P}$, we set $n=4$ encoders, $h=6$ heads, and $d=192$ embedding dimension; for $\textrm{ViT}_{\textbf{M}}\textrm{-}\textrm{R}$, we use $n=2$, $h=3$, and $d=192$, following HIPT \cite{chen2022scaling}.
We train $\textrm{mViT}_{\textbf{P}}\textrm{-}\textrm{C}$ for 250K iterations using AdamW \cite{adam2014method} with batch size 256 and base learning rate 0.0007; the first 20 epochs are warm-up to the base rate, followed by cosine decay \cite{loshchilov2016sgdr}. 
$\textrm{ViT}_{\textbf{R}}\textrm{-}\textrm{P}$ and $\textrm{ViT}_{\textbf{M}}\textrm{-}\textrm{R}$ are each trained for 200K iterations. 
In DINO \cite{caron2021emerging}, the teacher is updated via EMA of the student. 
For \textbf{MRPT-LLaVA}, stage 1 uses learning rate 0.001 and batch size 64; only a two-layer projection head is trained to align WSI-text features. 
We apply LoRA with rank 128 and $\alpha=256$ for efficient tuning.
All models are trained on four NVIDIA A100 GPUs.
\textbf{Patch-level Tasks:} We evaluate patch-level classification and VQA across 17 datasets. 
Patch-level classification uses 13 independent datasets: WSSS4LUAD \cite{han2022wsss4luad}, SICAP \cite{silva2021self}, DigestPath \cite{da2022digestpath}, CRC100K \cite{Kather2018_colorectal}, RenalCell \cite{brummer2022integrative}, MHIST \cite{wei2021petri}, PCam \cite{veeling2018rotation}, Osteo \cite{arunachalam2019viable}, BACH \cite{iciar2018grand}, SkinCancer \cite{kriegsmann2022deep}, LC-Lung \cite{Borkowski2019_LC25000}, LC-Colon \cite{Borkowski2019_LC25000}, and DataBiox \cite{bolhasani2020histopathological}, with both zero-shot and linear-probe settings. 
Zero-shot patch-level VQA is conducted on five datasets including QuiltVQA \cite{seyfioglu2024quilt}, PathVQA \cite{He2020_PathVQA}, PMC-VQA \cite{zhang2023pmcvqa}, and PathMMU \cite{Sun2024_PathMMU}. 
\textbf{WSI-level Tasks:} We evaluate WSI-level classification, VQA, captioning, and report generation across 17 datasets. 
WSI-level classification uses 10 external datasets: PANDA \cite{Bulten2022_PANDA}, HunCRC \cite{pataki2022huncrc}, CPTAC-NSCLC \cite{Edwards2015CPTAC}, Brains30 \cite{roetzer2022digital}, BRACS \cite{brancati2022bracs}, TCGA-RCC \cite{hutter2018cancer}, DHMC-RCC \cite{zhu2021development}, Camelyon17 \cite{bandi2018detection}, and UBC-OCEAN \cite{UBC-OCEAN}, 
with zero-shot, WSMIL, and FSWL evaluations. 
WSI-level VQA and captioning are evaluated on SlideBench-Caption \cite{Chen_2025_CVPR}, SlideBench-VQA (TCGA) \cite{Chen_2025_CVPR}, SlideBench-VQA (BCNB) \cite{Chen_2025_CVPR}, WSI-VQA \cite{Chen2025_WSI-VQA}, and WSI-Bench \cite{Liang2024_WSI-LLaVA}. 
WSI-level report generation uses HistGen \cite{guo2024histgen} and WSI-Bench (Report) \cite{Liang2024_WSI-LLaVA}. 
\textbf{\textit{See suppl. mat. for more details, ablations, and results.}}
\vspace{-3mm}
\subsection{Evaluation Setup}
\vspace{-2mm}
For classification task, we used Balanced Accuracy (BA), weighted $F_{1}$, and Accuracy (A) \cite{chen2024towards, lu2024visual}. 
For captioning and report generation: BLEU-2/4, ROUGE-L, and METEOR. 
For VQA: accuracy (closed-ended) and recall, where applicable.
For classification, we compare against CLIP \cite{radford2021learning}, PLIP \cite{huang2023visual}, PathCLIP \cite{sun2024pathasst}, HIPT \cite{chen2022scaling}, KEP \cite{zhou2024knowledge}, MI-Zero \cite{lu2023visual}, UNI \cite{chen2024towards}, CONCH \cite{lu2024visual}, QuiltNet \cite{ikezogwo2024quilt}, CPLIP \cite{javed2024cplip}, MR-PLIP \cite{albastaki2025multi}, OmniPath \cite{sun2025cpath}, CTransPath \cite{wang2022transformer}, REMEDIS \cite{azizi2022robust}, CHIEF \cite{wang2024pathology}, DinoPath \cite{kang2023benchmarking}, Phikon \cite{filiot2024phikon}, PathGenCLIP \cite{sun2024pathgen}, Virchow \cite{vorontsov2024foundation}, GigaPath \cite{xu2024whole}, TITAN \cite{ding2024multimodal}, PRISM \cite{shaikovski2024prism}, and RudolfV \cite{dippel2024rudolfv}. 
Zero-shot prompts follow dataset-specific templates from CONCH \cite{lu2024visual} and QuiltNet \cite{ikezogwo2024quilt}. For VQA, captioning, and report generation, we compare with general LMMs—GPT-4V \cite{hurst2024gpt}, GPT-4o \cite{hurst2024gpt}, LLaVA \cite{liu2024improved}, Qwen-VL-Max \cite{bai2023qwenvlversatilevisionlanguagemodel}, and Gemini Pro-V \cite{team2023gemini}—and CPath-specific MLLMs—Quilt-LLaVA \cite{ikezogwo2024quilt}, HistGen \cite{guo2024histgen}, MIGen \cite{chen2024wsicaption}, SlideChat \cite{chen2025slidechat}, WSI-LLaVA \cite{liang2025wsillavamultimodallargelanguage}, MedDr \cite{he2024meddr}, LLaVA-Med \cite{li2023llava}, and PathGen-LLaVA \cite{sun2024pathgen}. 
We use official implementations, matched test splits, and consistent inference prompts \cite{chen2024towards}. 
\vspace{-7mm}
\subsection{Ablation Studies}
\vspace{-3mm}
\textbf{1. Hierarchical Variants of MRPT (Table \ref{table2}).} 
We analyze variants using $\textrm{mViT}_{\textbf{P}}\textrm{-}\textrm{C}$ alone and in combination with $\textrm{ViT}_{\textbf{R}}\textrm{-}\textrm{P}$ and $\textrm{ViT}_{\textbf{M}}\textrm{-}\textrm{R}$. 
(A) Freeze $\textrm{mViT}_{\textbf{P}}\textrm{-}\textrm{C}$ and apply ABMIL \cite{ilse2018attention} for WSI classification (no hierarchy). 
(B) Freeze $\textrm{mViT}_{\textbf{P}}\textrm{-}\textrm{C}$ and $\textrm{ViT}_{\textbf{R}}\textrm{-}\textrm{P}$; use ABMIL for WSI classification. 
(C) Freeze $\textrm{mViT}_{\textbf{P}}\textrm{-}\textrm{C}$ and $\textrm{ViT}_{\textbf{R}}\textrm{-}\textrm{P}$; fine-tune only $\textrm{ViT}_{\textbf{M}}\textrm{-}\textrm{R}$. 
(D) Train a linear classifier on pre-extracted MRPT WSI-level features. 
\textit{\textbf{Results:}} All MRPT variants outperform HIPT \cite{chen2022scaling}, validating multi-resolution hierarchical representations. 
$\textrm{mViT}_{\textbf{P}}\textrm{-}\textrm{C}\text{-B}$ yields the strongest results among tiny/small/backbone options; thus we report MRPT with $\textrm{mViT}_{\textbf{P}}\textrm{-}\textrm{C}\text{-B}$ thereafter. 
Experiment~D averages $86.30\%$, exceeding B ($84.0\%$) and C ($81.30\%$), indicating richer WSI-level features from MRPT than from patch/region-only ViTs.

\begin{table}[t!]
\caption{Performance comparison across different datasets with same training and testing splits.}
\centering
\makebox[\linewidth]{
\scalebox{0.70}{
\begin{tabular}{l|ccc|ccc|cccc|cccc|cccc}
\hline
\textbf{Method} 
& \multicolumn{3}{c|}{\textbf{Camelyon16}} 
& \multicolumn{3}{c|}{\textbf{Camelyon17}} 
& \multicolumn{4}{c|}{\textbf{TCGA-NSCLC}} 
& \multicolumn{4}{c|}{\textbf{TCGA-RCC}} 
& \multicolumn{4}{c}{\textbf{TCGA-BRCA}} \\
\cline{2-19}
 & Train & Test & $F_{1}$
 & Train & Test &  $F_{1}$
 & WSIs & Train & Test & $F_{1}$
 & WSIs & Train & Test & $F_{1}$
 & WSIs & Train & Test & $F_{1}$\\
\hline
Baseline HIPT & 270 & 130 & 0.933&500 & 500 &0.782& 1041 & 80$\%$ & 20$\%$ &0.851& 1203 & 80$\%$ & 20$\%$ & 0.861& 1048 & 80$\%$ & 20$\%$& 0.866\\
ABMIL         & 270 & 130 &0.886& 500 & 500 &0.766& 1041 & 80$\%$ & 20$\%$ &0.809& 1203 & 80$\%$ & 20$\%$ & 0.831&1048 & 80$\%$ & 20$\%$& 0.831\\
DSMIL         & 270 & 130 &0.906& 500 & 500 &0.773&1041 & 80$\%$ & 20$\%$ &0.867& 1203 & 80$\%$ & 20$\%$ & 0.903&1048 & 80$\%$ & 20$\%$&0.871 \\
TransMIL         & 270 & 130 &0.924& 500 & 500 &0.796&1041 & 80$\%$ & 20$\%$ &0.812& 1203 & 80$\%$ & 20$\%$ &0.911&1048 & 80$\%$ & 20$\%$&0.881 \\
CLAM-MB         & 270 & 130 &0.922& 500 & 500 &0.805&1041 & 80$\%$ & 20$\%$ &0.844& 1203 & 80$\%$ & 20$\%$ &0.867&1048 & 80$\%$ & 20$\%$&0.873 \\
Focus         & 270 & 130 &0.902& 500 & 500 &0.802&1041 & 80$\%$ & 20$\%$ &0.882& 1203 & 80$\%$ & 20$\%$ & 0.882&1048 & 80$\%$ & 20$\%$&0.852 \\
UNI         & $\times$ & 130 &0.934& $\times$ & 500 &0.857&1041 &$\times$& 20$\%$ &0.908& 1203 & $\times$ & 20$\%$ &0.920 &1048 & $\times$& 20$\%$&0.911 \\
GigaPath        & $\times$ & 130 &0.911& $\times$ & 500 &0.802&1041 & $\times$& 20$\%$ &0.814& 1203 & $\times$& 20$\%$ & 0.905&1048 & $\times$ & 20$\%$&0.871 \\
Vila-MIL      & 270 & 130 & 0.891&500 & 500 &0.812&1041 & 80$\%$ & 20$\%$ &0.852& 1203 & 80$\%$ & 20$\%$ & 0.863&1048 & 80$\%$ & 20$\%$& 0.876\\
SI-MIL        & 270 & 130 & 0.921&500 & 500 &0.821& 1041 & 80$\%$ & 20$\%$ &0.820& 1203 & 80$\%$ & 20$\%$ & 0.847&1048 & 80$\%$ & 20$\%$& 0.837\\
MRPT ($\textrm{mViT}_{\mathbf{P}}\textrm{-C}$) 
              & 270 & 130 & 0.955&500 & 500 &\underline{0.855}& 1041 & 80$\%$ & 20$\%$ &0.908& 1203 & 80$\%$ & 20$\%$ & 0.923&1048 & 80$\%$ & 20$\%$ &\textbf{0.966}\\
MRPT ($\textrm{mViT}_{\mathbf{R}}\textrm{-P}$)
              & 270 & 130 &\textbf{0.971}& 500 & 500 &0.834& 1041 & 80$\%$ & 20$\%$ &\underline{0.918}& 1203 & 80$\%$ & 20$\%$ & \underline{0.936}&1048 & 80$\%$ & 20$\%$& 0.951\\
MRPT          & 270 & 130& \underline{0.962}& 500 & 500 &\textbf{0.883}& 1041 & 80$\%$ & 20$\%$ &\textbf{0.946}& 1203 & 80$\%$ & 20$\%$ & \textbf{0.956}&1048 & 80$\%$ & 20$\%$ &\underline{0.964}\\
\hline
\end{tabular}
}
}
\label{tab:dataset_split}
\end{table}

\begin{table}[t!]
\caption{Component-wise ablation of MRPT model.}
\centering
\makebox[\linewidth]{
\scalebox{0.65}{
\small
\begin{tabular}{lccccccc}
\hline
Variant &$\Delta$ Vs. Previous Row&Training Data&PANDA & BRAINS&UBC-OCEAN\\
\hline
(a) HIPT (20$\times$ only, 2-Stage Hier. SSL)&Baseline&36K WSIs&0.704&0.638&0.781 \\
(b) + Multi-Resolution Inputs (Concat)&+ Multi-resolution data, MRPT, no CCRA&36K WSIs&0.762&	0.701	&0.834 \\
(c) + CCRA (Consecutive Only)&+ CCRA Module&36K WSIs&0.823&	0.776&	0.882 \\
(d) + ACT+[CLS] Concat&+ Token Fusion&36K WSIs&0.848&0.795&0.902 \\
(e) + 3-Stage Hier. SSL (full)&+ Hierarchical SSL&36K WSIs&\textbf{0.866}&\textbf{0.813}&\textbf{0.910} \\
\hline
\end{tabular}
}
}
\label{tab:component_ablation}
\end{table}

\noindent\textbf{2. Multi-Resolution Cross-Attention (Table \ref{table3}).} 
We compare six fusion strategies: \textit{[CLS] Concat} (concatenate all [CLS] tokens), \textit{ACT} (average [CLS] tokens), \textit{All-Token Attention} (self-attention over concatenated tokens across resolutions; strong but quadratic), \textit{Pairwise Sum} (spatially aligned token-wise fusion with separate [CLS] fusion), \textit{CCRA} (Consecutive Cross-Resolution Attention), and our combined \textit{ACT+CCRA+[CLS] Concat}. 
\textit{ACT+CCRA+[CLS] Concat} achieves the best accuracy, highlighting the complementary roles of averaging (Eq.~(\ref{eqn2})) and [CLS] concatenation within CCRA. 
\textbf{See more ablations in supplementary material.} 

\noindent\textbf{3. Disentangling Data Scale and MRPT Contributions (Table \ref{tab:dataset_split}).}
To isolate data scale effects, we conduct a matched-data experiment shown in Table~\ref{tab:dataset_split} where all WSI classification methods are trained and tested under identical splits.
MRPT consistently outperforms baselines by a significant margin, showing gains stem from hierarchical multi-resolution SSL and cross-resolution alignment, not pre-training scale.

\noindent\textbf{4. Component-wise Contribution Analysis (\ref{tab:component_ablation}).}
MRPT component-wise ablation is presented in Table~\ref{tab:component_ablation}, showing consistent gains from multi-resolution inputs, CCRA, token fusion, and 3-stage SSL, confirming improvements arise from structural design, not a coupled pipeline.

\begin{table}[t]
\caption{Ablation of MRPT (Model D in Table \ref{table2}) across cross-resolution token fusion methods.
The combined ACT + CCRA + [CLS] Concat achieves the best accuracy.}
\vspace{-1em}
\centering
\makebox[\linewidth]{
\scalebox{0.75}{
\begin{tabular}{lccc}
\hline
Token Fusion Method &PANDA& BRAINS & UBC-OCEAN \\
\hline
[CLS] Concat&0.823&0.745&0.846\\
Average [CLS] Token (ACT)&0.848&0.789&0.882\\
All-Token Attention&0.850&0.795&0.902\\
Pairwise Sum&0.843&0.765&0.861\\
Consecutive Cross-Resolution Attention (CCRA)&\underline{0.862}&\underline{0.810}&\underline{0.907}\\
ACT$+$CCRA$+$[CLS] Concat&\textbf{0.866}&\textbf{0.813}&\textbf{0.910}\\
\hline
\end{tabular}
}
}
\label{table3}
\vspace{-1em}
\end{table}

\begin{table}[t]
\centering
\caption{Report generation performance comparison across SOTA methods on the WSI-Bench and HistGen datasets. }
\vspace{-1em}
\resizebox{\columnwidth}{!}{\centering
\begin{tabular}{l|cccc}
\toprule
Models&BLEU-2&BLEU-4&ROUGE-L&METEOR\\
\hline
WSICaption&0.111 $|$ 0.137 &0.136 $|$ 0.540&0.402 $|$ 0.251&0.366 $|$ 0.322\\
HistGen& 0.307 $|$ 0.297& 0.208  $|$ 0.184& 0.448 $|$ 0.344 &0.416 $|$ 0.182\\
GPT-4O&0.069 $|$ 0.420 & 0.016 $|$ 0.100& 0.132 $|$ 0.128 & 0.167 $|$ 0.144\\
MIGen& 0.306  $|$0.466 &0.209 $|$ 0.234& 0.446 $|$ 0.322& 0.407 $|$ 0.412 \\
Quilt-LLaVA &0.351 $|$ 0.331 &0.236 $|$ 0.488& 0.475 $|$ 0.491 & 0.460$|$ 0.443\\
SlideChat&0.310 $|$ \underline{0.533} &0.191 $|$ \underline{0.655}&0.463 $|$ 0.492  &0.422 $|$ 0.613  \\
WSI-LLaVA&\underline{0.358} $|$ 0.527 &\underline{0.240} $|$ 0.644 &\underline{0.490} $|$ \underline{0.544} &\underline{0.465} $|$ \underline{0.655}\\
MRPT-LLaVA&\textbf{0.401} $|$ \textbf{0.634}&\textbf{0.287} $|$ \textbf{0.699}&\textbf{0.522} $|$ \textbf{0.581}&     \textbf{0.511} $|$ \textbf{0.701}  \\
\hline
\end{tabular}
}
\label{table_reports}
\vspace{-5mm}
\end{table}
\subsection{Main Results and Comparisons}
\textbf{1. Patch-level Zero-Shot Classification (Fig.~\ref{fig:patch_a}).} 
We evaluate $\textrm{mViT}_{\textbf{P}}\textrm{-}\textrm{C}$-LLaVA and $\textrm{mViT}_{\textbf{R}}\textrm{-}\textrm{P}$-LLaVA against 11 SOTA VLMs on 13 datasets.
\textit{Across most datasets, our models achieve notably higher accuracy.} 
On average, $\textrm{mViT}_{\textbf{P}}\textrm{-}\textrm{C}$-LLaVA attains 83.66\% and $\textrm{mViT}_{\textbf{R}}\textrm{-}\textrm{P}$-LLaVA 82.11\%; the third-best, PathGenCLIP, reaches 73.60\%. 
\textit{Thus, our patch-level model improves by $\sim$10.0\% over the existing SOTA.} \noindent\textbf{2. Patch-level Linear Probe Evaluations (Fig.~\ref{fig:patch_b}).} 
With linear classifiers atop features from $\textrm{mViT}_{\textbf{P}}\textrm{-}\textrm{C}$ and $\textrm{mViT}_{\textbf{R}}\textrm{-}\textrm{P}$, we compare against 10 SOTA VFMs. 
\textit{Averages: 92.75\% for $\textrm{mViT}_{\textbf{P}}\textrm{-}\textrm{C}$ and 92.22\% for $\textrm{mViT}_{\textbf{R}}\textrm{-}\textrm{P}$; the next best, GigaPath, achieves 86.12\%.} 
This yields a 6.63\% average gain for our patch encoder. \noindent\textbf{3. Patch-level Zero-shot VQA (Fig.~\ref{fig:patch_c}).} 
We compare our $\textrm{mViT}_{\textbf{P}}\textrm{-}\textrm{C}$-LLaVA and $\textrm{mViT}_{\textbf{R}}\textrm{-}\textrm{P}$-LLaVA with nine MLLMs. 
\textit{On five closed-ended datasets: 73.12\% (patch) and 71.52\% (region) vs.\ 49.48\% (PathGen-LLaVA) and 48.60\% (Quilt-LLaVA). 
On two open-ended datasets: 56.05\% (patch) and 51.95\% (region) vs.\ 35.80\% (PathGen-LLaVA).} 
Overall improvements average 23.64\% (closed) and 20.25\% (open), emphasizing our multi-resolution design. \noindent \textbf{4. Zero-Shot WSI Classification (Fig.~\ref{fig:patch_d}).} 
We compare MRPT-LLaVA with 12 CPath SOTA VLMs on 10 datasets. 
\textit{Average balanced accuracy: 0.806 for MRPT-LLaVA vs.\ 0.751 for the next-best TITAN}, showcasing the benefits of multi-resolution WSI representations, whereas PRISM/TITAN rely on single-resolution pre-training. \noindent\textbf{5. WSI Classification Using WSMIL (Fig.~\ref{fig:patch_e}).} 
Following \cite{chen2024towards}, we apply ABMIL \cite{ilse2018attention} atop features from $\textrm{mViT}_{\textbf{P}}\textrm{-}\textrm{C}$ and $\textrm{mViT}_{\textbf{R}}\textrm{-}\textrm{P}$. 
\textit{Compared to 10 SOTA patch-level methods, our encoders achieve mean balanced accuracies of 0.863 and 0.851 across 10 datasets; MR-PLIP and UNI reach 0.812 and 0.798, respectively.} \noindent\textbf{6. WSI Linear Probe (Fig.~\ref{fig:patch_e}).} 
Using MRPT’s WSI-level features with a linear classifier, we compare against PRISM, TITAN, GigaPath, and CHIEF. 
\textit{MRPT averages 0.898 balanced accuracy vs.\ 0.845 for TITAN, underscoring the advantage of multi-resolution cues.} \noindent\textbf{7. WSI VQA (Fig.~\ref{fig:patch_f}).} 
Across four VQA datasets, we compare with seven CPath MLLMs. 
\textit{MRPT-LLaVA averages 80.77\% accuracy vs.\ 72.57\% for WSI-LLaVA, indicating substantial gains from aligning MRPT with the LLM.} \noindent\textbf{8. WSI Report Generation (Table \ref{table_reports}).} 
For report generation (WSI-Bench, HistGen), MRPT-LLaVA outperforms seven recent methods, including SlideChat and WSI-LLaVA (Tables \ref{table_reports}), \textit{evidencing strong generalization of multi-resolution features with LLMs}.
\vspace{-9mm}

\section{{Conclusion}}
\label{sec:conclusion}
\vspace{-2mm}
We introduced a multi-resolution hierarchical WSI foundation model (MRPT) that integrates information across resolutions to enhance diverse CPath tasks, including WSI classification, VQA, captioning, and report generation. 
At its core, MRPT employs a \emph{Consecutive Cross-Resolution Attention (CCRA)} mechanism to fuse information across adjacent resolutions. 
A multi-resolution ViT captures cell-level features at varying granularities using CCRA, aggregates them at the patch level, and progressively encodes region and WSI-level representations. 
Each hierarchy is pre-trained using multi-resolution SSL.
MRPT jointly models global tissue architecture and fine-grained cellular morphology, yielding substantial improvements over SOTA. 
Its biologically inspired, multi-resolution design reflects the diagnostic process of pathologists—integrating contextual and microscopic cues.
Building on MRPT, we further proposed \emph{MRPT-LLaVA}, a MLLM for patch and WSI-level reasoning. 
Extensive evaluations confirm that our models surpass single-resolution and existing MLLM baselines, achieving superior generalization.
Future extensions incorporating multi-modal data, such as genomics and radiology, promise to advance holistic patient-level understanding.

\section{Acknowledgment}
This research was funded by Khalifa University of Science and Technology through the Faculty Start-Ups under Project ID: KU-INT-FSU-2005-8474000775.

{
    \small
    \bibliographystyle{splncs04}
    \bibliography{main}
}

\newpage
\large
\textbf{{\centering
Supplementary Material: \\ From Multi-Resolution Cells to Gigapixel Whole Slide Images Foundation Model for Computational Pathology}}

\normalsize

\section{Architectural Comparison with Existing SOTA Pathology Foundation Models}
Table \ref{comaprison} compares our proposed models (MRPT and MRPT-LLaVA) with representative SOTA pathology foundation models in terms of input resolution, representation structure, and fusion strategy. 
Most existing methods operate under a single-resolution setting and rely on either patch-level or WSI-level representations, limiting their ability to jointly model fine-grained cellular details and global tissue architecture.
While some approaches introduce hierarchical modeling or partial multi-resolution designs, they typically remain constrained to a single magnification or lack unified cross-scale integration.

\textit{In contrast, MRPT and MRPT-LLaVA adopt a true multi-resolution paradigm (10×, 20×, 40×) and construct a unified hierarchical representation spanning cell, patch, region, and WSI levels. Combined with a dedicated CCRA-based fusion mechanism, our models enable structured cross-resolution interaction and more comprehensive multi-scale modeling, distinguishing them from prior approaches.}

\begin{table*}[h]
\caption{Architectural comparison between our proposed models (MRPT and MRPT-LLaVA) and representative SOTA pathology foundation models. 
The table contrasts input resolution (single vs. multi-resolution), representation structure (patch-level, WSI-level, or hierarchical across cell, patch, region, and WSI), and fusion strategy. Unlike prior single-resolution and non-hierarchical approaches, our models leverage multi-resolution inputs (10$\times$, 20$\times$, 40$\times$), hierarchical representations spanning cell to WSI levels, and a CCRA-based fusion mechanism.}
\resizebox{\linewidth}{!}{\centering
\begin{tabular}{l|c|c|c|}
\hline
Model&Resolution &  Representation Structure & Fusion\\
\hline
PLIP \cite{huang2023visual}&Single& Patch&$\times$\\
CONCH \cite{lu2024visual}&Single&Patch&$\times$\\
UNI \cite{chen2024towards}&Single& Patch&$\times$\\
QuiltNet \cite{ikezogwo2024quilt}&Single& Patch&$\times$\\
CPLIP \cite{javed2024cplip}&Single& Patch&$\times$\\
GigaPath \cite{xu2024whole}&Single& WSI&$\times$\\
Virchow \cite{vorontsov2024foundation}&Single&WSI&$\times$\\
RudolfV \cite{dippel2024rudolfv}&Single& Patch&$\times$\\
REMEDIS \cite{azizi2022robust}&Single& Patch&$\times$\\
CHIEF \cite{wang2024pathology}&Single& Patch&$\times$\\
SlideChat \cite{chen2025slidechat}&Single&WSI&$\times$\\
TITAN \cite{ding2024multimodal}&Single&WSI&$\times$\\
WSI-LLaVA \cite{liang2025wsillavamultimodallargelanguage}&Single&WSI&$\times$\\
Quilt-LLaVA \cite{seyfioglu2024quilt}&Single& Patch&$\times$\\
MR-PLIP \cite{albastaki2025multi}&Multi-resolution& Patch&Multi-modal Encoder\\
HIPT \cite{chen2022scaling}&20$\times$&Hierarchical (Cell, Patch, Region)&Aggregation\\
\hline
Proposed MRPT&Multi-Resolution (10$\times$, 20$\times$, 40$\times$)&Hierarchical (Cell, Patch, Region, WSI)&CCRA\\
Proposed MRPT-LLaVA&Multi-resolution (10$\times$, 20$\times$, 40$\times$)&Hierarchical (Cell, Patch, Region, WSI)&CCRA\\
\hline
\end{tabular}}
\label{comaprison}
\end{table*}

\begin{table*}[t!]
\caption{Performance evaluations by varying different WSI resolution levels and CCRA and distant cross-resolution attention combinations.
MRPT with $\textrm{mViT}_{\textbf{\textrm{P}}}\textrm{-}\textrm{C}\textrm{-}\textrm{B}$ is used in this experiment (setting D in Table \textcolor{NavyBlue}{2} in the main manuscript).}
\resizebox{\linewidth}{!}{\centering
\begin{tabular}{lcccc}
\hline
Model&Resolution &  PANDA& BRAINS & UBC-OCEAN \\
\hline
\multirow{4}{*}{MRPT}&10$\times$, 20$\times$&\underline{0.852}&0.784&\underline{0.885}\\
&10$\times$, 40$\times$&0.826&0.750&0.832\\
&20$\times$, 40$\times$&0.840&\underline{0.786}&0.883\\
&10$\times$, 20$\times$, 40$\times$&\textbf{0.866}&\textbf{0.813}&\textbf{0.910}\\
HIPT \cite{chen2022scaling}&20$\times$&0.623&0.567&0.723\\
\hline
Model&CCRA&PANDA&BRAINS&UBC-OCEAN \\
\hline
\multirow{2}{*}{MRPT}&10$\times \leftrightarrow 20 \times$, 20$\times \leftrightarrow 40 \times$&\textbf{0.866}&\textbf{0.813}&\textbf{0.910}\\
&10$\times \leftrightarrow 40 \times$, 20$\times \leftrightarrow 40 \times$&\underline{0.824}&\underline{0.776}&\underline{0.884}\\
&10$\times \leftrightarrow 40 \times$, 20$\times \leftrightarrow 40 \times$, 10$\times \leftrightarrow 20 \times$&0.820&0.751&0.876\\
\hline
\end{tabular}
}
\label{resolution}
\end{table*}

\begin{table*}[t!]
\caption{Performance evaluation by varying the number of CCRA layers ($K$) and $\textrm{mViT}_{\textbf{\textrm{P}}}\textrm{-}\textrm{C}$ encoder blocks ($Q$), and embedding dimension in $\textrm{ViT}_{10\times}$.
MRPT with $\textrm{mViT}_{\textbf{\textrm{P}}}\textrm{-}\textrm{C}\textrm{-}\textrm{B}$ is used in this experiment (setting D in Table \textcolor{NavyBlue}{2} in the main manuscript).
}
\resizebox{\linewidth}{!}{\centering
\begin{tabular}{l|c c c|c ccc|cc}
\hline
Variants &  \multicolumn{3}{c|}{Dimension} & \# heads & Encoders & $K$ & $Q$&PANDA&BRAINS\\
 &  $10 \times$ & $20 \times$ & $40 \times$ & $h_L,h_M,h_N$ &$L, M, N$ &&&&\\
\hline
MRPT&384 & 768 & 768 &  12,12,12 & 2,4,4&1&4&\textbf{0.866}&\textbf{0.813}\\
\hline
1 &\textcolor{blue}{768}& 768 & 768 &  12,12,12 & 2,4,4&1&4&\underline{0.846}&0.796\\
2  &384& 768 & 768 &  12,12,12 & \textcolor{blue}{4},4,4&1&4&0.841&0.795\\
3  &384& 768 & 768 &  12,12,12 &2,4,4&\textcolor{blue}{2}&4&0.843&0.794\\
4  &384& 768 & 768 &  12,12,12 &2,\textcolor{blue}{2},\textcolor{blue}{2}&2&\textcolor{blue}{8}&0.838&\underline{0.802}\\
5  &384& 768 & 768 &  12,12,12 &2,\textcolor{blue}{2},\textcolor{blue}{4}&2&\textcolor{blue}{2}&0.842&0.801\\
\hline
\end{tabular}
}
\label{architecture}
\end{table*}

\begin{table}[t!]
\caption{Comparison of different LLMs integrated with the MRPT encoder across three WSI-VQA benchmarks: SlideBench (BCNB), WSI-VQA, and WSI-Bench (Close-ended). 
Results highlight that Qwen2-1.5B achieves the best overall performance across all datasets.}
\resizebox{\columnwidth}{!}{\centering
\begin{tabular}{lccccc}
\hline
\multirow{2}{*}{\textbf{LLMs}} & SlideBench &WSI-&WSI-Bench \\
& (BCNB) &VQA&Close-ended \\
\hline
Vicuna-7B-v1.5 \cite{chiang2023vicuna}&0.600&\underline{0.661}&\underline{0.955}\\
Phi-3-Mini-4k-Instruct \cite{abdin2024phi}&0.566&0.633&0.905\\
Llama3-8B-Instruct \cite{grattafiori2024llama}&\underline{0.602}&0.655&0.911 \\
Internlm2-Chat-7B \cite{cai2024internlm2}&0.588&0.634&0.887 \\
\hline
Qwen2-1.5B \cite{yang2024qwen2}&\textbf{0.622}&\textbf{0.688}&\textbf{0.977}\\
\hline
\end{tabular}
}
\label{LLM}
\end{table}

\section{Additional Ablation Studies}

\subsection{Number of WSI Resolutions (Table \ref{resolution})} 
All multi-resolution variants outperform single-resolution HIPT at 20$\times$ \cite{chen2022scaling} as shown in Table \ref{resolution}. 
\textit{Consecutive pairs (10$\times$+20$\times$, 20$\times$+40$\times$) consistently surpass the distant 10$\times$+40$\times$ pairing, confirming the necessity of intermediate scales for robust cross-attention and semantic alignment.} 
The best results use all three resolutions (10$\times$, 20$\times$, 40$\times$), jointly capturing architectural context, mesoscopic organization, and nuclear detail.

\subsection{Consecutive vs.\ Distant Cross-Resolution Attention (Table \ref{resolution})} 
CCRA (10$\times\!\leftrightarrow\!20\times$, 20$\times\!\leftrightarrow\!40\times$) outperforms distant attention (10$\times\!\leftrightarrow\!40\times$, 20$\times\!\leftrightarrow\!40\times$) across datasets as shown in Table \ref{resolution}. 
Adding 10$\times\!\leftrightarrow\!40\times$ to MRPT increases parameters without improving accuracy. 
\textit{Anchoring information flow through intermediate scales preserves semantic consistency; directly coupling distant magnifications misaligns global (10$\times$) and fine (40$\times$) cues, degrading performance—supporting our pathologist-inspired CCRA design.}

\subsection{$\textrm{ViT}_{10\times}$ Width/Depth (Table \ref{architecture})} 
Increasing $\textrm{ViT}_{10\times}$ embedding size and encoder count raises FLOPs/parameters without gains (rows 1 \& 2 in Table \ref{architecture}). 
\textit{$\textrm{ViT}_{10\times}$ serves as a lightweight global-context provider; discriminative features primarily arise from $\textrm{ViT}_{20\times}$ and $\textrm{ViT}_{40\times}$ (Table \ref{architecture}).}
This yields accuracy with computational scalability.

\subsection{ CCRA Layers ($K$) and Encoders ($Q$) in $\textrm{mViT}_{\textbf{P}}\textrm{-}\textrm{C}$ (Table \ref{architecture})} 
Increasing $K$ or $Q$ inflates compute and parameters without accuracy gains (rows 3, 4, \& 5 in Table \ref{architecture}). 
\textit{A single CCRA layer with lightweight encoders suffices, indicating an efficient operating point.}

\subsection{ LLM Integration (Table \ref{LLM})} 
We pair the MRPT encoder with multiple LLMs across three WSI-VQA benchmarks. 
\textit{Qwen2-1.5B achieves the highest scores (62.20\%, 68.80\%, 97.70\%), surpassing larger LLMs (Llama3-8B, InternLM2-Chat-7B) as shown in Table \ref{LLM}.} 
This demonstrates strong generalization and alignment between MRPT’s multi-resolution features and Qwen2’s instruction following.

\subsection{Inference Using Single WSI Resolution (Table \ref{rigid})}
To address data flexibility, we present two ablations showing that MRPT operates with a single available resolution (Table \ref{rigid} (A \& B)).
In the first, MRPT is pre-trained on native multi-resolution WSIs and evaluated on PANDA, BRAINS, and UBC-OCEAN using only 20$\times$ WSIs via a synthetic pyramid (downsampling and upsampling), where it still outperforms a single-resolution hierarchical baseline (Table \ref{rigid} (A)). 
In the second, MRPT is both trained and tested using only 20$\times$ WSIs augmented with a synthetic pyramid (TCGA for pre-training; PANDA, UBC-OCEAN, and BRAINS for testing), again yielding consistent gains. 
\textit{These results show that MRPT leverages partial data without architectural changes and degrades gracefully to single-resolution settings}. 
MRPT uses cross-resolution alignment as an inductive bias, trading minimal pre-training constraints for improved representation quality

\begin{table*}[t!]
\captionof{table}{Results using real and synthetic multi-resolution WSIs.}
\centering
\makebox[\textwidth]{
\scalebox{0.80}{
\begin{tabular}{lcc|ccc}
\hline
Model &Trained&Tested&PANDA& BRAINS & UBC-OCEAN \\
\hline
Baseline HIPT&Single Resolution (20$\times$)&Single Resolution (20$\times$)&0.623&0.567&0.723\\
Proposed MRPT&Multi-Resoluiton&Multi-resolution&\textbf{0.866}&\textbf{0.813}&\textbf{0.910}\\
\textbf{A. MRPT}&Real Multi-Resolution&Synthetic Multi-resolution&\underline{0.813}&\underline{0.755}&\underline{0.861}\\
\textbf{B. MRPT$^{*}$}&Synthetic Multi-resolution&Synthetic Multi-resolution&0.723&0.687&0.743\\
\hline
\end{tabular}
}
}
\label{rigid}
\end{table*}

\subsection{Comparison with SOTA Hierarchical Vision Models (Table \ref{hierarchical})}
While hierarchical aggregation has been studied in generic vision models (e.g., SwinTrans \cite{liu2021swin}, HIIF \cite{jiang2025hiif}, HAFA \cite{chen2023building}, and ViT-Adapter \cite{chen2022vision}), these methods operate on single-resolution natural images and do not address multi-resolution gigapixel WSIs. 
MRPT targets a distinct setting by jointly modeling hierarchical structure and cross-resolution semantic consistency across WSI resolutions.
MRPT integrates hierarchical and multi-resolution SSL at all levels via biologically inspired constraints (CCRA restriction, class-token–only interaction, and progressive refinement). 
\textit{We compare MRPT with SwinTrans and ViT-Adapter models in Table~\ref{hierarchical}}. 
These models show lower performance despite feature aggregation and underperform due to a lack of multi-resolution WSIs information. 
\textit{The gains of MRPT stem from its hierarchical multi-resolution SSL framework and CCRA constraints rather than generic hierarchical aggregation.}

\begin{table*}[t!]
\captionof{table}{Comparison of hierarchical SOTA models with proposed MRPT.}
\centering
\makebox[\linewidth]{
\scalebox{0.60}{
\begin{tabular}{lcc|ccc}
\hline
Model &Trained&Tested&PANDA& BRAINS & UBC-OCEAN \\
\hline
Proposed MRPT&Multi-Resoluiton&Multi-resolution&\textbf{0.866}&\textbf{0.813}&\textbf{0.910}\\
SOTA SwinTrans&Single Resolution (20$\times$)&Single Resolution (20$\times$)&0.641&0.568&0.654\\
SOTA ViT-Adapter&Single Resolution (20$\times$)&Single Resolution (20$\times$)&0.651&0.587&0.661\\
\hline
\end{tabular}
}
}
\label{hierarchical}
\end{table*}

\section{WSI-level Caption Generation Results (Table \ref{caption})}
We also evaluated our proposed MRPT-LLaVA model on the WSI-level caption generation task on SlideBench-Caption dataset \cite{chen2025slidechat} as shown in Table \ref{caption}.
On SlideBench-Caption, MRPT-LLaVA surpasses six SOTA MLLMs across all NLP metrics (Table \ref{caption}) evidencing strong generalization of multi-resolution hierarchical features with LLMs.

\begin{table}[t!]
\caption{Computational cost comparison on PANDA dataset.}
\centering
\makebox[\linewidth]{
\scalebox{0.70}{
\small
\begin{tabular}{lccccc}
\toprule
\textbf{Model} & \textbf{Resolutions} & \textbf{FLOPs (GFLOPs/WSI)} & \textbf{Peak GPU mem (GB)} & \textbf{Latency (s/WSI)} \\
\midrule
HIPT        & 20$\times$ only & 412  & 8.7  & 150 \\
UNI         & 20$\times$ only & 1840 & 22.4 & 204 \\
GigaPath    & 20$\times$ only & 6210 & 38.6 & 258 \\
TITAN       & 20$\times$ only & 6480 & 39.2 & 264 \\
PRISM       & 20$\times$ only & 3120 & 28.8 & 228 \\
\midrule
MRPT (ours) & 10$\times$, 20$\times$, 40$\times$ & 1130 & 12.1 &  250\\
MRPT-mViT$_{\mathbf{P}}\textrm{-C}$ only & 10$\times$, 20$\times$, 40$\times$ & 488 & 9.4 & 170 \\
\bottomrule
\end{tabular}
}
}
\label{tab:compute}
\end{table}

\section{Computational Time}
All experiments use four NVIDIA A100 GPUs. 
During inference with linear probing, the MRPT encoder is frozen and only a linear classifier is trained. 
On PANDA WSI classification, MRPT with linear probing averages 5.2 minutes per WSI, compared to $\{4.3, 4.4, 3.8\}$ minutes for GigaPath, TITAN, and PRISM, respectively. 
Experiment~B (in Table 2 in manuscript) averages 3.2 minutes per WSI, while HIPT averages 2.5 minutes. 
\textit{Despite using multi-resolution, MRPT’s runtime remains comparable to SOTA methods, reflecting effective architectural choices that reduce complexity and parameter count (in Table 2 in manuscript).}
Table \ref{tab:compute} compares the complexity of the SOTA models and our proposed MRPT.

\begin{table}[t!]
\centering
\caption{Comparison of WSI-level captioning generation performance across SOTA methods on the SlideBench-Caption dataset. 
MRPT-LLaVA achieves the best results across all metrics (BLEU-1–4, ROUGE-L, and METEOR), demonstrating superior text generation and VL alignment.}
\resizebox{\columnwidth}{!}{\centering
\begin{tabular}{l|cccccc}
\toprule
Models& BLEU-1& BLEU-2&BLEU-3& BLEU-4&ROUGE-L&METEOR \\
\hline
GPT-4O& 0.100 & 0.030 & 0.010 & 0.010 & 0.110 & 0.131 \\
Quilt-LLaVA& 0.230 & 0.090 & 0.040 & 0.010 & 0.160 & 0.420 \\
MIGen & 0.370 & 0.240 & 0.150 & 0.100 & 0.250 & 0.381 \\
HistGen&0.300&0.181&0.110&0.090&0.171&0.288 \\
SlideChat & 0.370 & 0.210 & 0.120 & 0.080 & 0.240 & 0.488\\
WSI-LLaVA&\underline{0.411}&\underline{0.266}&\underline{0.180}&\underline{0.150}&\underline{0.320}&\underline{0.551}\\
MRPT-LLaVA&\textbf{0.491}&\textbf{0.361}&\textbf{0.230}&\textbf{0.266}&\textbf{0.393}&\textbf{0.596} \\
\hline
\end{tabular}
}
\label{caption}
\end{table}

\section{Interpretability analysis}
\begin{figure*}
\centering
\includegraphics[width=10cm, height=7cm]{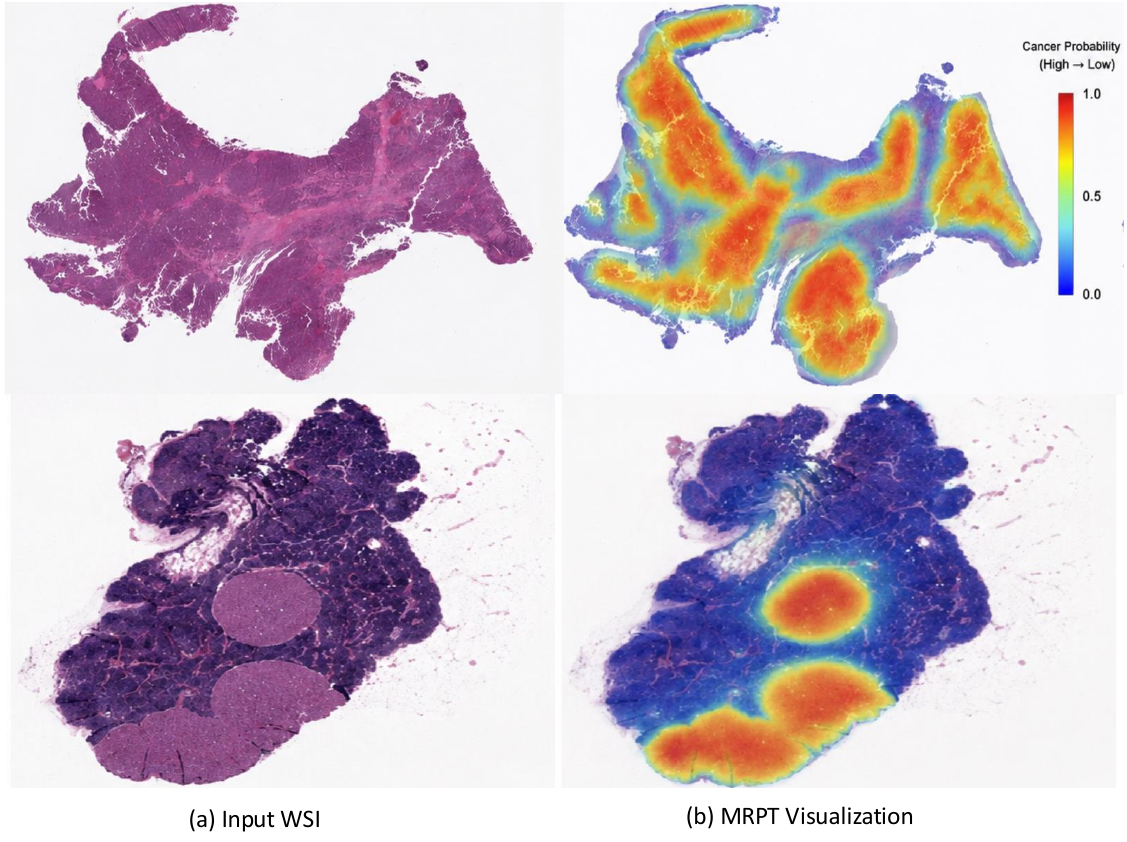}
\caption{Interpretability analysis (cancer heatmaps).}
\label{fig1_heatmap}
\end{figure*}

To evaluate the interpretability of MRPT, we estimated WSI-level cancer probability heatmaps, as shown in Fig. \ref{fig1_heatmap}.
Regions with higher predicted cancer probability are highlighted in red/yellow, while low-risk regions appear in blue.
The heatmaps demonstrate that MRPT consistently focuses on diagnostically relevant tissue regions and suppresses background areas. 
High-response regions correspond well with suspicious morphological structures, indicating that the model learns meaningful pathological features for cancer detection. 
These results enhance the transparency of the proposed framework and provide visual evidence supporting its predictions.

\section{Theortical Insights}
\subsection{Significance of Multi-Resolution Hierarchical Gigapixel WSI Representation}
\begin{figure} [t!]

   \centering
\begin{subfigure}{0.9\columnwidth}
   \includegraphics[width=\textwidth]{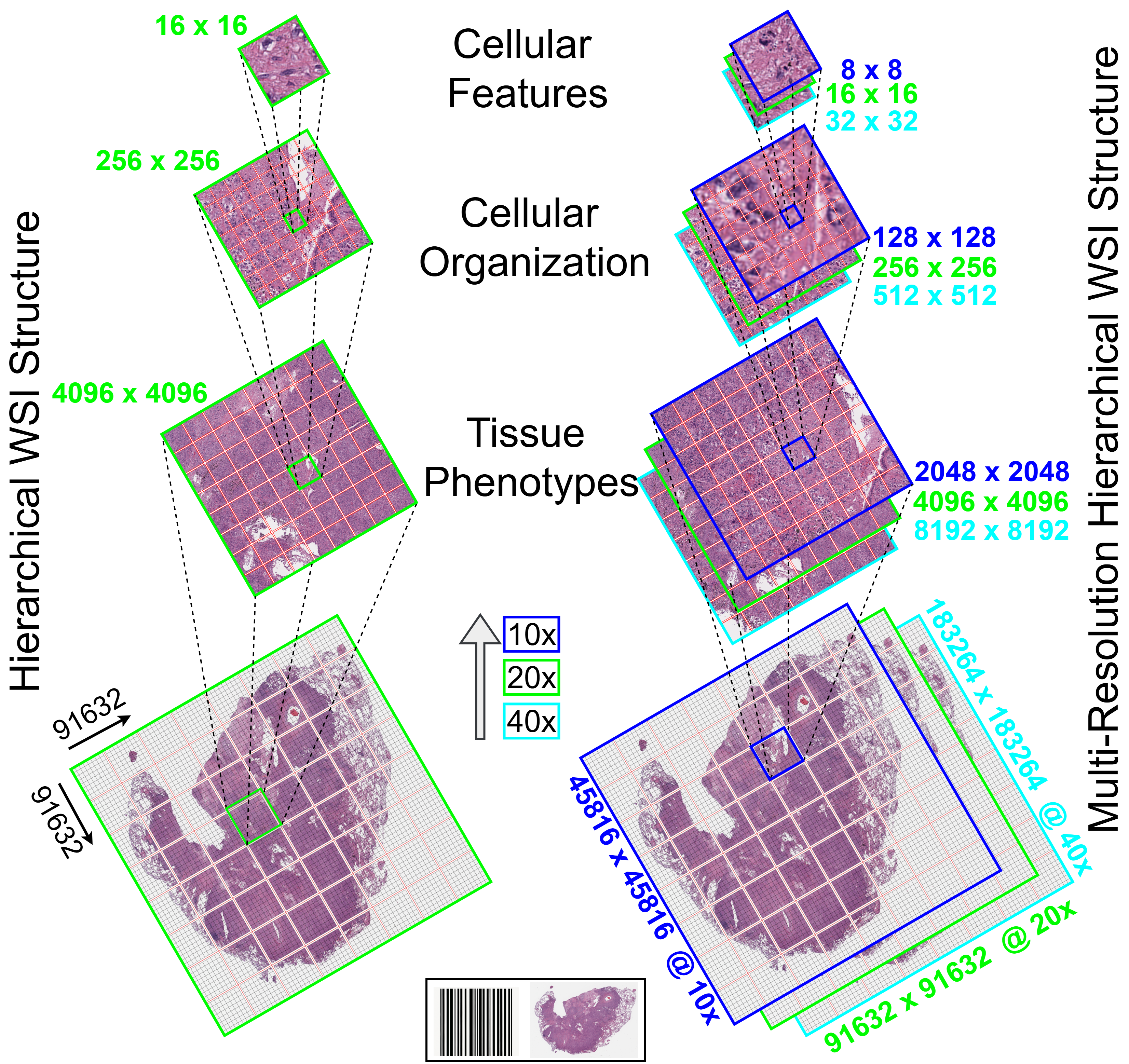}
   \caption*{\textbf{Left:} SOTA HIPT model \cite{chen2022scaling}. \textbf{Right:} Proposed MRPT Model}
   \label{fig1_supp:top}
\end{subfigure}
\caption{Compares SOTA HIPT model \cite{chen2022scaling} (\textbf{left}) with the proposed MRPT model (\textbf{right}).}
\label{fig1_supp}
\end{figure}

Histopathological interpretation is inherently a hierarchical and resolution-dependent process, where diagnostic reasoning emerges from the integration of morphological information observed across multiple resolutions (Fig. \ref{fig1_supp}) \cite{albastaki2025multi, chen2022scaling, vanRijthoven2021HookNet, abels2019computational, zhang2019pathologist, chen2022fast}.
In clinical practice, a pathologist never relies on a single resolution view to reach a diagnosis; instead, they continuously transition between low, medium, and high magnifications to interpret complementary visual cues. 
At low magnifications (2$\times$–10$\times$), the global organization of the tissue, including architectural patterns, stromal distribution, and tumor boundaries, provides the essential contextual scaffold for diagnosis. 
At intermediate magnifications (10$\times$–20$\times$), glandular structures, nest formation, and stromal invasion become apparent, enabling assessment of architectural distortion and local tissue heterogeneity \cite{baidoshvili2023whole, nan2025deep, zarella2022high}.
Finally, high magnifications (40$\times$) reveal the cellular and nuclear features that define malignancy—such as pleomorphism, mitotic figures, and nuclear atypia (Fig. \ref{fig1_supp}) \cite{trahearn2025computational, nascimento2024characteristics, ladabaum2020strategies, kanth2021screening}. 
Diagnostic confidence emerges only when information from multiple scales is coherently integrated across magnification levels \cite{schuffler2022efficient, ghezloo2022analysis}.

This multi-resolution reasoning implies that the information embedded within WSIs lies on a hierarchical feature manifold, where features extracted at different resolutions correspond to nested, complementary levels of abstraction (Fig. \ref{fig1_supp}) \cite{schuffler2022efficient, ghezloo2022analysis}.  
Low-resolution representations capture global spatial context but lack discriminative fine detail, whereas high-resolution representations provide localized discriminatory power but are contextually ambiguous \cite{baidoshvili2023whole}. 
Theoretical frameworks in hierarchical representation learning suggest that combining features from multiple resolutions enables models to approximate the complete feature manifold more effectively, achieving representations that are both semantically complete and topologically smooth \cite{ladabaum2020strategies, kanth2021screening}.

\begin{figure*}[t!]
\centering
\begin{subfigure}{0.45\columnwidth}
   \centering
    \includegraphics[width=\textwidth]{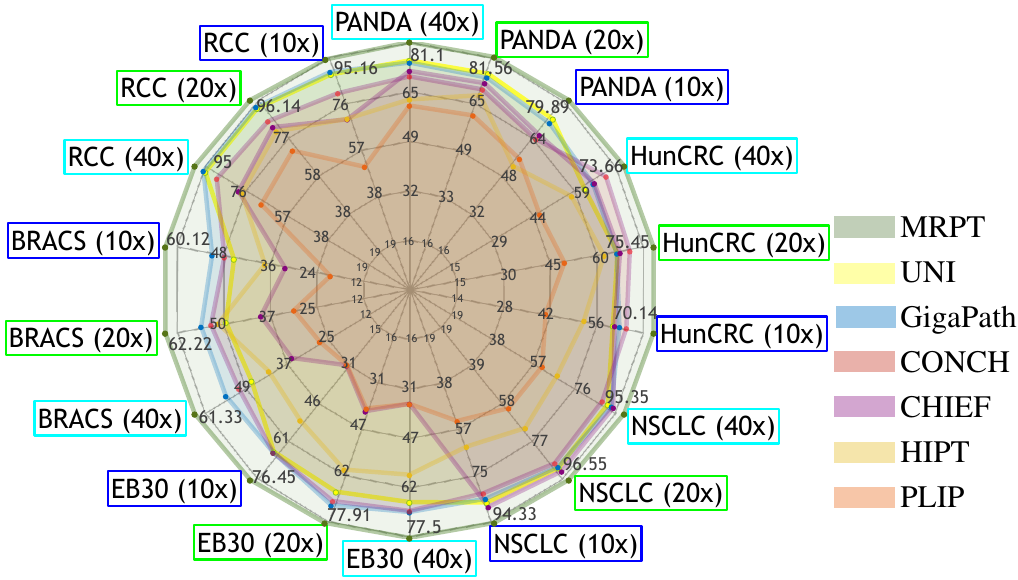}
    \caption{WSI Classification Task (Balanced Accuracy)}
    \label{fig:bottom-left}
\end{subfigure}
\begin{subfigure}{0.45\columnwidth}
    \centering
    \includegraphics[width=1.15\textwidth]{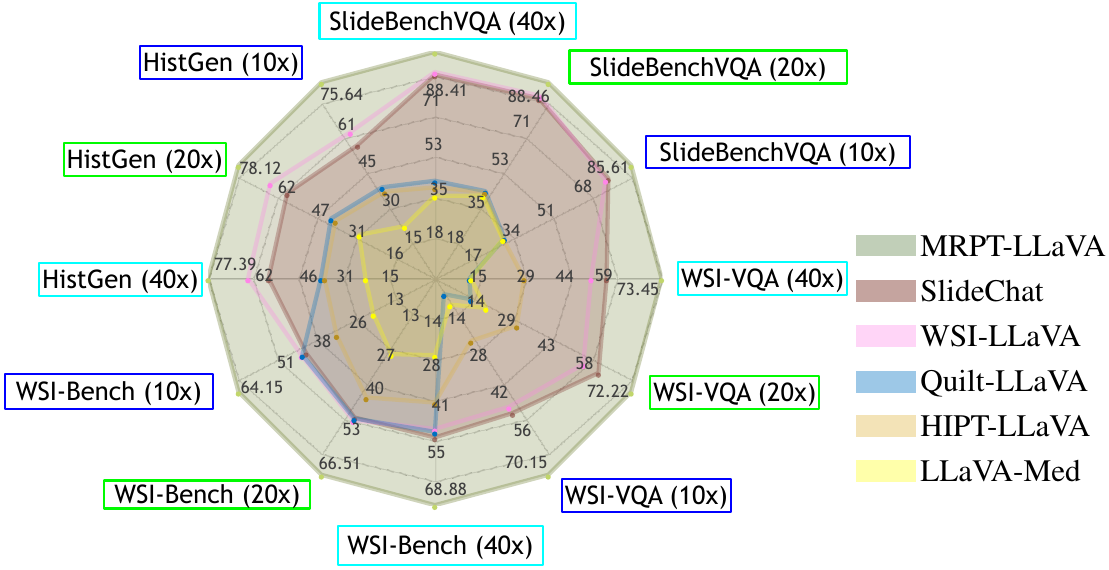}
    \caption{WSI VQA Task (Accuracy)}
    \label{fig:bottom-right}
\end{subfigure}
\caption{Our MRPT model remains consistent across resolutions and outperforms existing SOTA models by a significant margin.}
\label{fig2}
\end{figure*}

\begin{figure} [t!]
   \centering
\begin{subfigure}{0.45\columnwidth}
   \includegraphics[width=\textwidth]{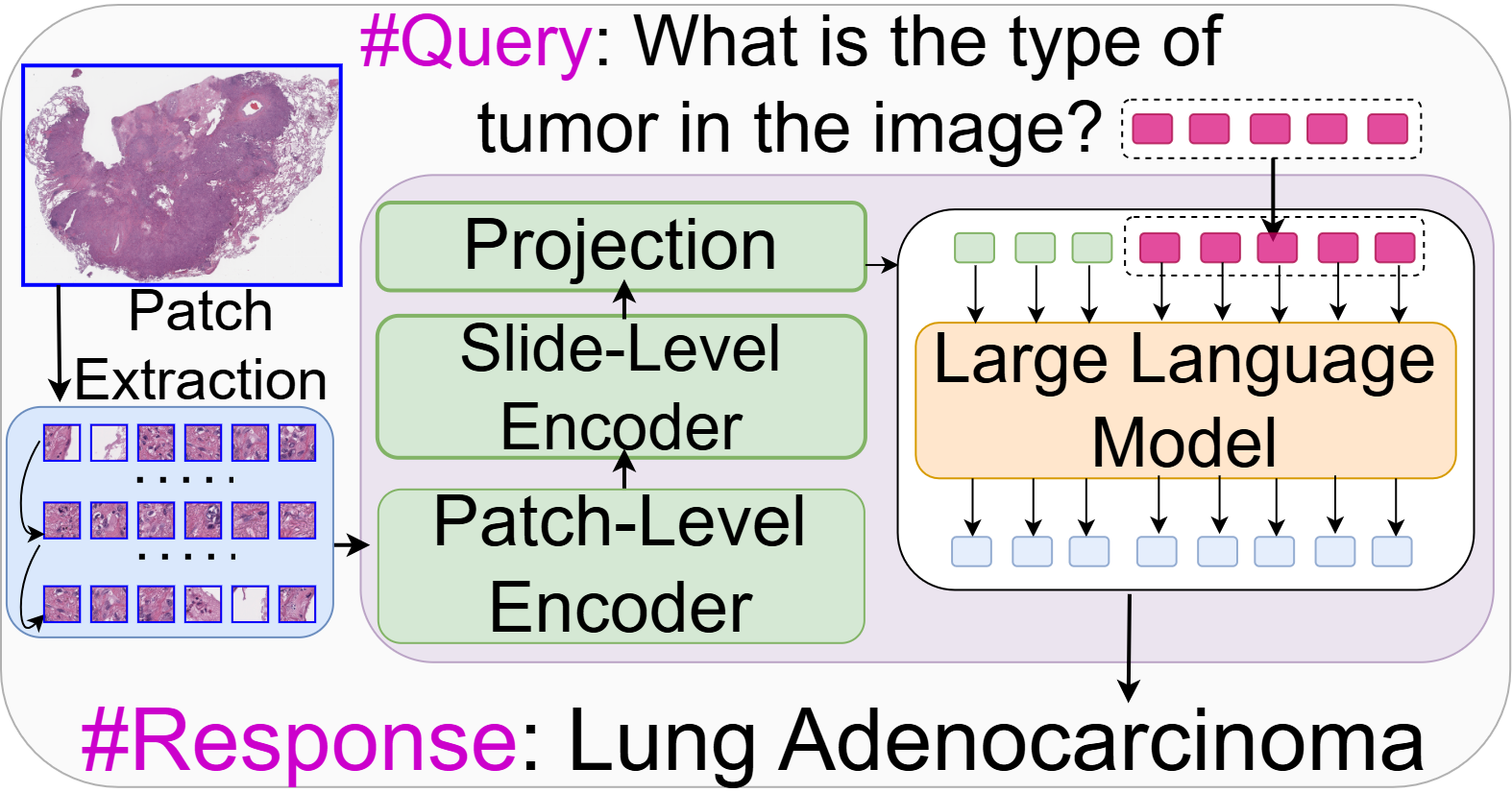}
   \caption{SOTA CPath MLLMs \cite{chen2025slidechat, liang2025wsillavamultimodallargelanguage}}
   \label{fig1_supp:left-bottom}
\end{subfigure}
\begin{subfigure}{0.45\columnwidth}
   \includegraphics[width=\textwidth]{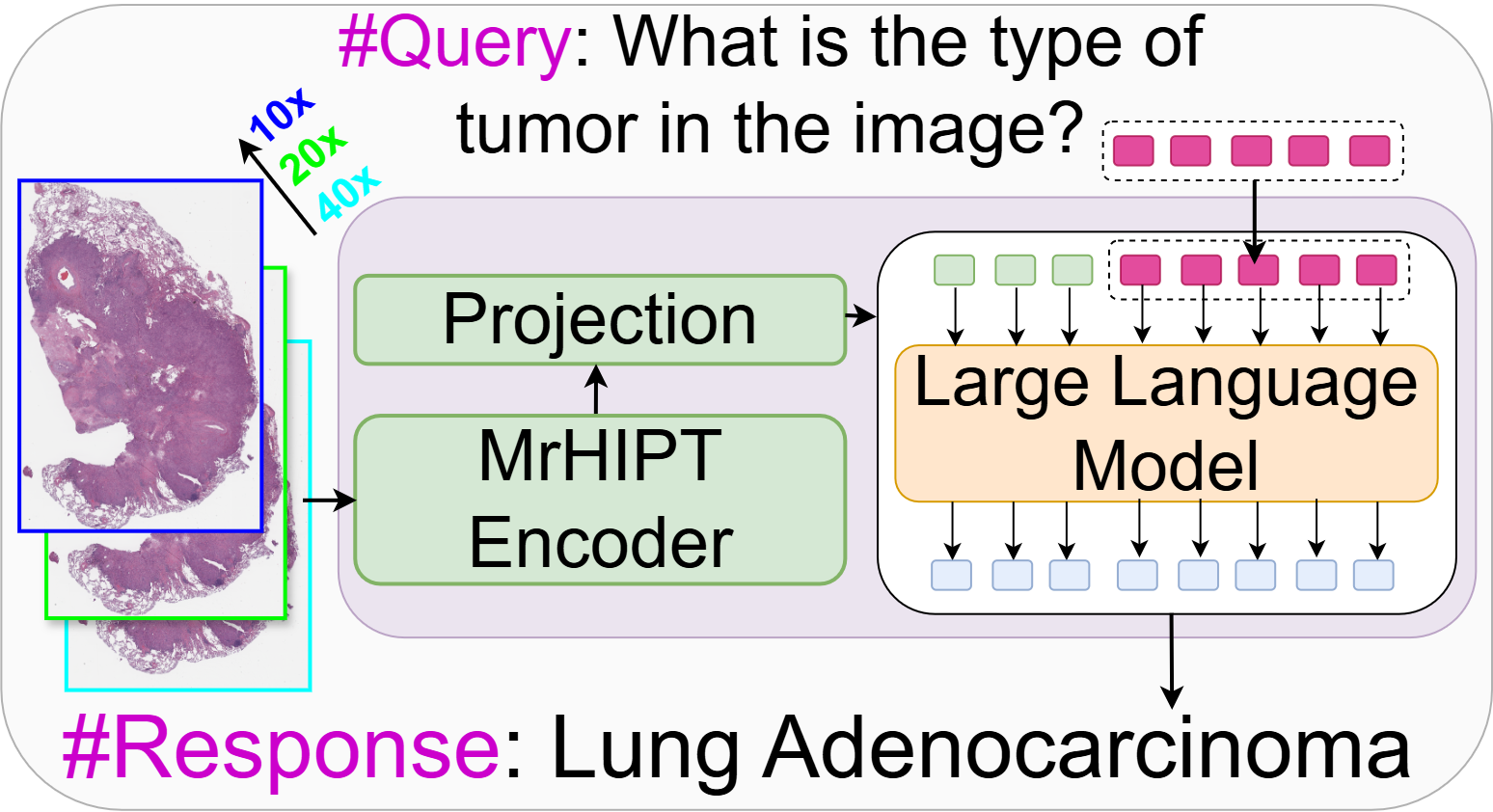}
   \caption{Proposed MRPT-LLaVA Model}
   \label{fig1_supp:right-bottom}
\end{subfigure}
\caption{Comparison between SOTA CPath WSI-level MLLMs \cite{chen2025slidechat, liang2025wsillavamultimodallargelanguage} with the proposed MRPT-LLaVA.
Oru model integrates multi-resolution hierarchical WSI-level representations with LLM to improve complex pathology VQA and report generation tasks.}
\label{fig3_supp}
\end{figure}

Existing CPath models, such as HIPT \cite{chen2022scaling} and TITAN \cite{ding2024multimodal}, based on single-resolution SSL pretraining—learn representations limited to a narrow slice of this manifold (Figs. \ref{fig1_supp} \& \ref{fig3_supp}). 
They fail to encode cross-resolution dependencies, and they often exhibit degraded generalization when applied to tissues with varying scanning resolutions (Fig. \ref{fig2}).
The single-scale encoders operate with restricted mutual information between hierarchical scales, whereas our multi-resolution WSI expands the joint information content by modeling conditional dependencies across resolutions.

Building on this principle, the proposed MRPT framework introduces a multi-resolution hierarchical SSL paradigm that aligns representations across consecutive resolutions (Fig. \ref{fig1_supp} (\textbf{right})).
This design is theoretically grounded in manifold continuity and hierarchical compositionality.
Adjacent resolutions (e.g., 10$\times$ $\leftrightarrow$ 20$\times$, 20$\times$ $\leftrightarrow$ 40$\times$) lie close on the feature manifold and share strong semantic correlations. 
Enforcing consistency between these resolutions through Consecutive Cross-Resolution Attention (CCRA) ensures that local features learned at high magnifications remain anchored to their global tissue context, forming a continuous embedding from coarse to fine representations.
In contrast, directly aligning distant magnifications (e.g., 10$\times$$\leftrightarrow$ 40$\times$) violates this manifold locality assumption, leading to unstable feature fusion and semantic misalignment.

From a clinical perspective, this multiresolution hierarchical WSI representation is not merely a computational convenience but a direct analog of pathologists' diagnostic reasoning.
A pathologist integrates global context (e.g., tumor extent) and cellular morphology (e.g., atypia, mitoses) through intermediate architectural patterns, forming a multi-resolution cognitive hierarchy. 
MRPT operationalizes this process algorithmically—ensuring that model reasoning is both interpretable and aligned with pathologists' diagnostic workflows.

In summary, MRPT multi-resolution hierarchical WSI modeling is essential because it (1) reconstructs the full morphological feature manifold through joint hierarchical learning, (2) enforces semantic continuity across consecutive resolutions via cross-resolution attention, and (3) mimics the multi-resolution hierarchical reasoning employed by expert pathologists.
These properties collectively enable MRPT to produce context-aware, scale-invariant, and clinically interpretable representations, setting the foundation for a robust and generalizable CPath model.

\section{Multi-resolution Pyramid Transformer (MRPT)}
Our Multi-Resolution Pyramid Transformer (MRPT) is a hierarchical foundation model designed to leverage the intrinsic multi-resolution structure of gigapixel WSIs.
MRPT learns representations jointly from 10$\times$, 20$\times$, and 40$\times$ resolutions, capturing both global tissue architecture and fine-grained cellular detail.
Our framework introduces Consecutive Cross-Resolution Attention (CCRA) to progressively fuse information between adjacent resolutions, ensuring semantic continuity and biologically coherent feature alignment.

\begin{figure*}[t!]
    \centering
    \includegraphics[width=\textwidth]{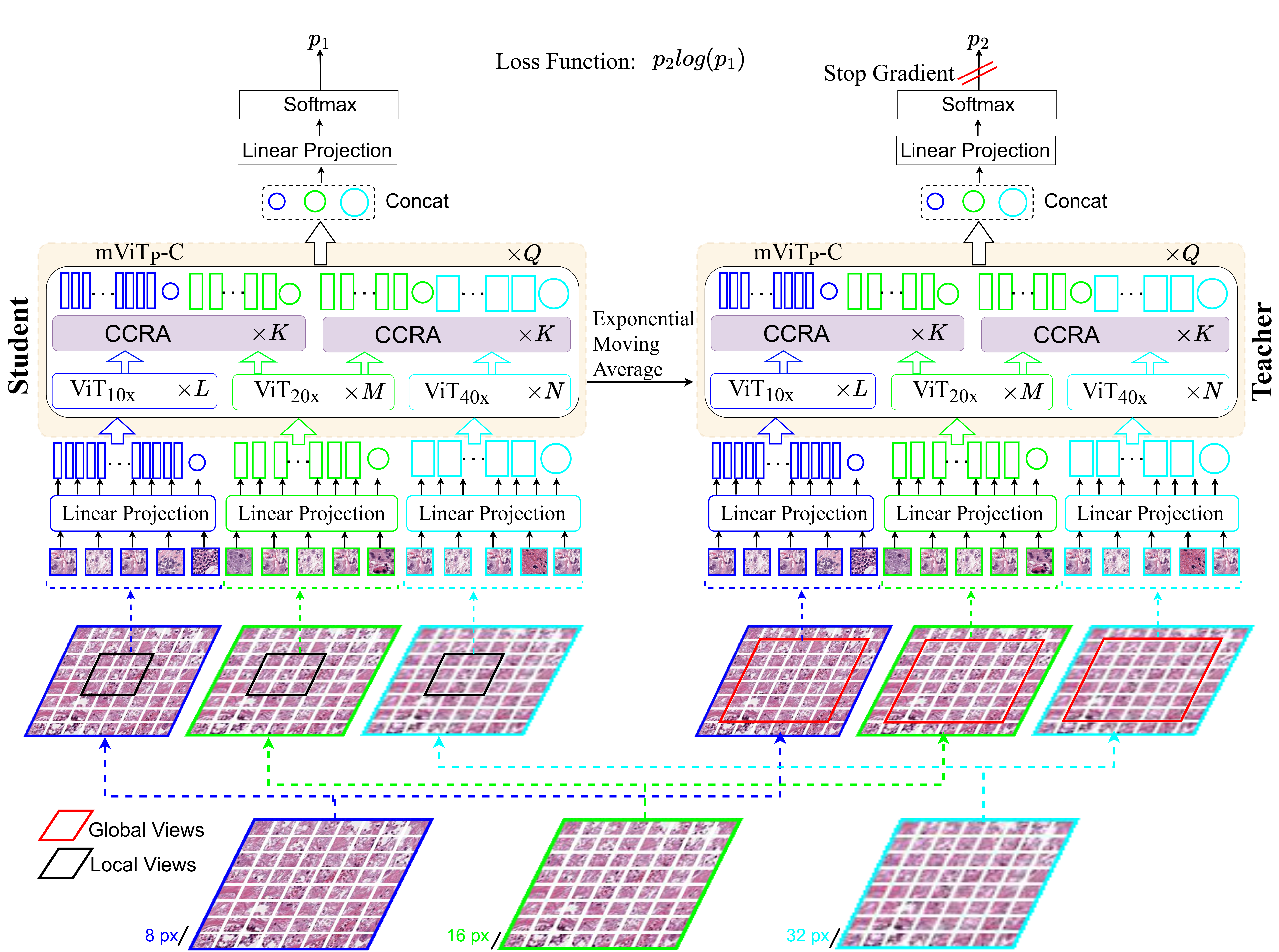}
    \caption{Schematic illustration of our proposed multi-resolution cellular transformer ($\textrm{mViT}_{\textbf{\textrm{P}}}\textrm{-}\textrm{C}$) employing multi-resolution Self-Supervised Learning (SSL) framework, where teacher and student networks utilize multi-resolution global and local views, employ the CCRA module, and then aggregate the final multi-resolution representation.
    MRPT employs a multi-resolution Self-Supervised Learning (SSL) framework at each hierarchical level, where the teacher and student networks utilize multi-resolution global and local views to estimate final multi-resolution embeddings.}
    \label{fig4}
\end{figure*}

MRPT is trained using a multi-resolution SSL paradigm (Fig. \ref{fig4}), where cell-, patch-, and region-level transformers are pre-trained hierarchically and then integrated into a unified WSI encoder. 
This multi-resolution hierarchical design mirrors the diagnostic reasoning process of pathologists—linking coarse contextual cues to high-resolution morphological patterns—while maintaining computational efficiency. 
As a result, MRPT produces context-aware, resolution-invariant, and clinically interpretable embeddings, setting a strong foundation for downstream CPath tasks such as classification, captioning, and VL alignment.

\section{Multi-resolution Self-Supervised Learning}
We follow a three-stage hierarchical pre-training multi-resolution SSL paradigm \cite{caron2021emerging}.
In stage 1, we pre-train $\textrm{mViT}_{\textbf{\textrm{P}}}\textrm{-}\textrm{C}$ in multi-resolution SSL manner (Fig. \ref{fig4}).
Then, in stage 2, we pre-train $\textrm{ViT}_{\textrm{\textbf{R}}}\textrm{-}\textrm{P}$ while keeping the weights of $\textrm{mViT}_{\textbf{\textrm{P}}}\textrm{-}\textrm{C}$ fixed.
Similarly, in stage 3 for WSI-level representation, we fixed the weights of both $\textrm{mViT}_{\textbf{\textrm{P}}}\textrm{-}\textrm{C}$ and $\textrm{ViT}_{\textrm{\textbf{R}}}\textrm{-}\textrm{P}$ and pre-trained $\textrm{ViT}_{\textrm{\textbf{M}}}\textrm{-}\textrm{R}$.
In our ablation study, we have evaluated the contribution of each pre-training stage in the overall MRPT model (\textit{see ablation in Table \textcolor{blue}{2} in the main manuscript)}.

\subsection{Stage 1: Multi-resolution Patch-Level ($\textbf{P}_{i}^{j}$) Pre-training } 
We pre-train $\textrm{mViT}_{\textbf{\textrm{P}}}\textrm{-}\textrm{C}$ using multi-resolution DINO-based SSL framework in which a student network $\Psi_{s}$ is trained to match the probability distribution of a teacher network $\Psi_{t}$ using a cross-entropy loss $H=-\sum_{i,j \in \{10 \times, 20 \times, 40 \times\}}  p_{t_{i}} \log p_{s_{j}}$ with momentum encoding, where $p_{t_{i}}$, and $p_{s_{j}}$ denoting the outputs of $\Psi_{s}$, and $\Psi_{t}$, respectively for resolutions $i$ and $j$ as shown in Fig. \ref{fig4}.
As data augmentation for each multi-resolution patch ($\textbf{P}_{i}^{j}$), DINO constructs a set of $L_{r} = 8$ local views at each resolution $r$ (crops of sizes $48 \times 48$ at $10\times$, $96 \times 96 $ at $20\times$, and $192 \times 192$ at $40\times$ input to $\Psi_{s}$ ) and $G_{r}= 2$ global views (crops of sizes $112 \times 112$, $224 \times 224$, and $448 \times 448$ input to $\Psi_{t}$) to encourage local-to-global correspondences between the student and teacher, minimizing function as:

\begin{equation}
\min_{\Psi_{s}} 
\sum_{r}
\sum_{G_{r}}
\sum_{L_{r}}
H \Big( p_{t_{r}} (G_{r}), p_{s_{r}}(L_{r}) \Big).
\end{equation}

\noindent Following \cite{caron2021emerging}, standard data augmentations (multi-resolution horizontal flips and color jittering) are applied to all views.

\subsection{Stage 2: Multi-Resolution Region-Level ($\textbf{R}_{i}$) Pretraining} 
By keeping the $\textrm{mViT}_{\textbf{\textrm{P}}}\textrm{-}\textrm{C}$ model frozen, we pre-train $\textrm{ViT}_{\textrm{\textbf{R}}}\textrm{-}\textrm{P}$ using multi-resolution SSL in which a student and teacher networks are pre-trained similar to stage 1.
The extracted multi-resolution patch-level tokens $[\textrm{mCLS}]_{\textbf{P}_{i}^{j}}$ from $\textrm{mViT}_{\textbf{\textrm{P}}}\textrm{-}\textrm{C}$ are input to the $\textrm{ViT}_{\textrm{\textbf{R}}}\textrm{-}\textrm{P}$.
We rearrange $\{[\textrm{mCLS}]_{\textbf{P}_{i}^{j}} \}_{j=1}^{256}$ as a $16 \times 16 \times 384$ feature grid for data augmentations, performing $[6 \times 6]$, $[14 \times 14]$ local-global patch-level crops.

\subsection{Stage 3: Multi-Resolution WSI-Level Pretraining}
Similar to stage 1 and 2, we pre-train $\textrm{ViT}_{\textrm{\textbf{M}}}\textrm{-}\textrm{R}$ using multiresolution SSL by keeping the $\textrm{mViT}_{\textbf{\textrm{P}}}\textrm{-}\textrm{C}$ and $\textrm{ViT}_{\textrm{\textbf{R}}}\textrm{-}\textrm{P}$ models frozen. 
The student and teacher networks are pre-trained similarly to stage 2.
The extracted multi-resolution region-level tokens $[\textrm{mCLS}]_{\textbf{R}_{i}}$ from $\textrm{ViT}_{\textrm{\textbf{R}}}\textrm{-}\textrm{P}$ are input to the $\textrm{ViT}_{\textrm{\textbf{M}}}\textrm{-}\textrm{R}$.
We re-arrange  $\{[\textrm{mCLS}]_{\textbf{R}_{i}} \}_{i=1}^{m}$ as a $\sqrt{m} \times \sqrt{m} \times 192$ feature grid for data augmentations assuming $m=256$ in most of the WSIs.
We perform $[6 \times 6]$, $[14 \times 14]$ local-global region-level crops for pre-training $\textrm{ViT}_{\textrm{\textbf{M}}}\textrm{-}\textrm{R}$.

\section{Consecutive Resolution Cross-Attention Mechanism}
The Consecutive Cross-Resolution Attention (CCRA) mechanism is a core component of MRPT model that enables effective information exchange across adjacent resolution ($10 \times  \leftrightarrow 20\times$ and $20 \times  \leftrightarrow 40\times$) in WSIs. Its design is grounded in both biological reasoning and representation learning, ensuring that multi-resolution fusion remains semantically meaningful, computationally stable, and hierarchically consistent.
CCRA is theoretically motivated by (1) the hierarchical organization of WSIs, (2) the manifold continuity and locality of cross-resolution representations, and (3) alignment with human diagnostic reasoning.
Together, these principles ensure that MRPT’s multi-resolution fusion remains semantically coherent, biologically interpretable, and computationally efficient, forming a principled foundation for hierarchical WSI representation learning.

The representations at different resolutions can be viewed as points on a multi-scale feature manifold. The distance between two representations at hierarchical scale $s$ and resolution $r$ correlates with the semantic gap between their visual abstractions.
Consecutive scales (e.g., $10 \times  \leftrightarrow 20\times$) lie closer on the manifold and share a high degree of mutual information, whereas distant scales (e.g., $10 \times  \leftrightarrow 40\times$ ) lie further apart, exhibiting larger representational heterogeneity.
By performing cross-attention only between adjacent magnifications, CCRA adheres to the locality principle of manifold learning, ensuring that information fusion occurs along smooth and semantically coherent transitions.

Clinically, CCRA mechanism parallels the diagnostic reasoning process of pathologists, who progressively zoom in from low to high magnifications rather than jumping directly across scales. 
The intermediate resolution acts as a semantic bridge, linking tissue-level patterns to cellular morphology. 
This consecutive integration mirrors human perceptual reasoning and ensures that the model’s learned features remain biologically interpretable and clinically grounded \cite{schuffler2022efficient, ghezloo2022analysis}.

\section{MRPT-LLaVA Model and Pre-Training}
We introduce MRPT-LLaVA, a large Vision–Language Model (VLM) that extends the proposed Multi-Resolution Pyramid Transformer (MRPT) into a multimodal framework for holistic gigapixel WSI understanding. 
MRPT-LLaVA bridges the gap between hierarchical visual representation learning and language-based reasoning, enabling natural language interaction, question answering, and report generation directly from gigapixel pathology slides.

A schematic illustration of MRPT-LLaVA is shown in Fig. \ref{fig3_supp} (b).
Most existing MLLMs in CPath operate at the patch-level, such as Quilt-LLaVA \cite{seyfioglu2024quilt}, PathGen-LLaVA \cite{sun2024pathgen}, and PathChat \cite{lu2024multimodal}, \textit{thereby lacking contextual information at the WSI-level. }
Recently proposed models like SlideChat \cite{chen2025slidechat} and WSI-LLaVA \cite{liang2025wsillavamultimodallargelanguage} address the WSI-level but rely on a single magnification, limiting their ability to capture hierarchical and multi-resolution representations (Figs. \ref{fig3_supp} (a)).
\textit{In our MRPT-LLaVA model, we exploit multi-resolution hierarchical WSI-level representation to solve complex pathology Visual Question Answering (VQA) tasks.}

At the core of MRPT-LLaVA lies the MRPT encoder, a multi-resolution transformer trained on WSIs across 10$\times$, 20$\times$, and 40$\times$ resolutions using Consecutive Cross-Resolution Attention (CCRA).
This design progressively fuses global, regional, and cellular information, capturing hierarchical tissue context that mirrors the diagnostic workflow of pathologists. 
Through hierarchical self-supervised pretraining, MRPT generates semantically consistent and scale-invariant embeddings that form a rich visual foundation for language alignment.

MRPT-LLaVA employs a three-stage training pipeline inspired by recent multimodal instruction-following frameworks~\cite{chen2025slidechat, liang2025wsillavamultimodallargelanguage, chen2025wsi, seyfioglu2024quilt}.
MRPT-LLaVA integrates an MRPT as a vision module, an LLM (Qwen2 1.5B model) as an LLM \cite{yang2024qwen2}, and a projection matrix, enabling the LLM to process visual information. 

\subsection{Stage 1: Cross-Modal Alignment} 
In the first stage, we align 9,642 WSI-caption pairs from TCGA dataset \cite{hutter2018cancer, liang2025wsillavamultimodallargelanguage} using contrastive learning \cite{radford2021learning}.
We extract multi-resolution WSI-level features using the MRPT encoder and textual features from the text encoder of the Qwen2-1.5B model \cite{yang2024qwen2} in this stage.
\subsection{Stage 2: Features Space Alignment}
In the second stage, the primary objective is to align the LLM’s word embeddings with the visual features extracted from the WSI-level MRPT encoder.
This alignment enables the LLM to interpret visual representations from the WSIs, facilitating the effective utilization of the intricate multi-resolution features within the WSIs.
During this process, MRPT-LLaVA is pre-trained to generate descriptive captions using 9,642 WSI-caption pairs from TCGA datasets \cite{hutter2018cancer, liang2025wsillavamultimodallargelanguage}.
Specifically, only the projection matrix is updated, while the LLM and MRPT weights are kept frozen.
\subsection{Stage 3: Visual Instruction Tuning}
In this stage, we focus on VQA tasks to train our model to accurately respond to domain-specific questions related to the WSIs.
We fine-tune our model on the WSI-LLaVA VQA dataset containing 175,450 VQA pairs at the WSI-level \cite{liang2025wsillavamultimodallargelanguage}.
In this stage, we freeze the MRPT encoder while training the projection matrix and LLM modules.
We employed Qwen2 1.5B model as an LLM \cite{yang2024qwen2}.

\section{Patch-level MRPT-LLaVA Model and Pre-Training}
The above-mentioned pre-training process of MRPT-LLaVA is for WSI-level VQA, caption, and report generation tasks. 
For the patch-level VQA tasks, we employed the same three-stage pre-training steps.
For this purpose, we utilized our MRPT, patch-level encoder $\textrm{mViT}_{\textbf{\textrm{P}}}\textrm{-}\textrm{C}$ and region-level encoder $\textrm{mViT}_{\textrm{\textbf{R}}}\textrm{-}\textrm{P}$ as backbone models, for three step pre-training.
We dub our patch-level LLaVA variants as $\textrm{mViT}_{\textbf{\textrm{P}}}\textrm{-}\textrm{C}$-LLaVA and $\textrm{mViT}_{\textrm{\textbf{R}}}\textrm{-}\textrm{P}$-LLaVA (please see results in the main manuscript in Fig. 4 (c) for VQA task).

Specifically, during Stage 1 and 2, we utilized the QuiltNet 1M image-text paired dataset \cite{ikezogwo2024quilt} for cross-modal retrieval and feature space alignment.
In stage 3, we employed the QuiltInstruct dataset containing 107,131 histology-specific instruction-answer pairs.

By combining the multi-resolution visual understanding of MRPT with the instruction-following capability of an LLM, MRPT-LLaVA provides an interpretable, scalable, and clinically aligned framework for CPath.
MRPT-LLaVA describes WSIs, answers diagnostic questions, and generates pathology-style narratives that reflect hierarchical tissue organization. 
This integration of pyramid-scale VL reasoning establishes MRPT-LLaVA as a step toward general-purpose pathology foundation models, capable of supporting automated diagnostics, clinical reporting, and interactive AI assistance in digital pathology.

\begin{table*}[t!]
\centering
\caption{WSI-Level zero-shot classification results and comparisons with SOTA CPath models.}
\resizebox{\textwidth}{!}{%
\begin{tabular}{c|cc|cc|cc|cc|cc|cc|cc|cc|cc|c}
\hline
 & \multicolumn{2}{|c|}{\textbf{PANDA}} & \multicolumn{2}{c|}{\textbf{HunCRC}} & \multicolumn{2}{c|}{\textbf{CPTAC-NSCLC}} & \multicolumn{2}{c|}{\textbf{Brains30}} & \multicolumn{2}{c|}{\textbf{BRACS}} & \multicolumn{2}{c|}{\textbf{TCGA-RCC}} & \multicolumn{2}{c|}{\textbf{DHMC-RCC}} & \multicolumn{2}{c|}{\textbf{Camelyon17}} & \multicolumn{2}{c|}{\textbf{UBC-Ocean}} & \textbf{Average} \\
 & F & BA & F & BA & F & BA & F & BA & F & BA & F & BA & F & BA & F & BA & F & BA &  BA \\
 \hline
CLIP & 0.156 & 0.11 & 0.167 & 0.156 & 0.146 & 0.109 & 0.02 & 0.01 & 0.234 & 0.203 & 0.203 & 0.233 & 0.304 & 0.255 & 0.143 & 0.123 & 0.123 & 0.109 & 0.140 \\
PLIP & 0.288 & 0.235 & 0.324 & 0.309 & 0.661 & 0.632 & 0.013 & 0.08 & 0.346 & 0.332 & 0.376 & 0.397 & 0.455 & 0.387 & 0.386 & 0.354 & 0.376 & 0.345 & 0.329 \\
PathCLIP & 0.461 & 0.455 & 0.506 & 0.487 & 0.706 & 0.677 & 0.223 & 0.187 & 0.487 & 0.455 & 0.623 & 0.607 & 0.721 & 0.687 & 0.486 & 0.441 & 0.657 & 0.612 & 0.494 \\
CPLIP & 0.445 & 0.42 & 0.507 & 0.498 & 0.809 & 0.766 & 0.253 & 0.233 & 0.523 & 0.487 & 0.607 & 0.634 & 0.687 & 0.706 & 0.473 & 0.459 & 0.706 & 0.653 & 0.526 \\
CONCH & 0.596 & 0.566 & 0.553 & 0.534 & 0.806 & 0.771 & 0.304 & 0.278 & 0.688 & 0.665 & 0.823 & 0.807 & 0.754 & 0.798 & 0.441 & 0.432 & 0.786 & 0.807 & 0.619 \\
QuiltNet & 0.532 & 0.509 & 0.611 & 0.587 & 0.773 & 0.752 & 0.229 & 0.201 & 0.656 & 0.623 & 0.807 & 0.786 & 0.786 & 0.754 & 0.446 & 0.406 & 0.753 & 0.776 & 0.590 \\
MR-PLIP & \textbf{0.701} & {\ul 0.681} & 0.655 & 0.633 & 0.842 & 0.806 & 0.332 & 0.314 & 0.655 & 0.643 & 0.845 & 0.823 & 0.812 & 0.832 & 0.548 & 0.507 & 0.855 & 0.833 & 0.661 \\
 PathGenCLIP & 0.511 & 0.488 & 0.577 & 0.558 & 0.774 & 0.734 & 0.255 & 0.221 & 0.557 & 0.523 & 0.706 & 0.723 & 0.823 & 0.856 & 0.405 & 0.386 & 0.786 & 0.756 & 0.576 \\
MI-Zero & 0.405 & 0.386 & 0.466 & 0.457 & 0.753 & 0.723 & 0.253 & 0.233 & 0.489 & 0.466 & 0.657 & 0.623 & 0.908 & 0.886 & 0.421 & 0.406 & 0.807 & 0.786 & 0.545 \\
KEP & 0.476 & 0.455 & 0.512 & 0.486 & 0.709 & 0.685 & 0.209 & 0.193 & 0.607 & 0.566 & 0.678 & 0.654 & 0.832 & 0.809 & 0.533 & 0.552 & 0.734 & 0.721 & 0.557 \\
TITAN & 0.621 & 0.608 & 0.674 & 0.657 & 0.916 & 0.919 & 0.365 & {\ul 0.543} & 0.706 & 0.667 & {\ul 0.917} & {\ul 0.94} & 0.943 & {\ul 0.966} & {\ul 0.723} & {\ul 0.733} & 0.908 & 0.865 & {\ul 0.751} \\
PRISM & 0.544 & 0.521 & 0.651 & 0.633 & {\ul 0.936} & {\ul 0.938} & 0.263 & 0.279 & 0.631 & 0.598 & 0.724 & 0.565 & 0.862 & 0.435 & 0.553 & 0.556 & 0.753 & 0.765 & 0.575\\
\hline
Proposed $\textrm{mViT}_{\textbf{\textrm{P}}}\textrm{-}\textrm{C}$-LLaVA  & 0.603 & 0.587 & {\ul 0.711} & {\ul 0.691} & 0.905 & 0.903 & {\ul 0.426} & 0.506 & {\ul 0.744} & {\ul 0.688} & 0.889 & 0.908 & 0.902 & 0.955 & 0.653 & 0.665 & {\ul 0.912} & {\ul 0.887} & 0.747 \\
Proposed $\textrm{mViT}_{\textbf{\textrm{R}}}\textrm{-}\textrm{P}$-LLaVA  & 0.566 & 0.552 & 0.706 & 0.675 & 0.891 & 0.886 & 0.405 & 0.488 & 0.708 & 0.671 & 0.912 & 0.887 & {\ul 0.955} & 0.922 & 0.687 & 0.706 & 0.886 & 0.853 & 0.730 \\
Proposed MRPT-LLaVA & {\ul 0.690} & \textbf{0.688} & \textbf{0.756} & \textbf{0.721} & \textbf{0.955} & \textbf{0.961} & \textbf{0.488} & \textbf{0.587} & \textbf{0.755} & \textbf{0.723} & \textbf{0.955} & \textbf{0.972} & \textbf{0.966} & \textbf{0.983} & \textbf{0.756} & \textbf{0.786} & \textbf{0.946} & \textbf{0.908} & \textbf{0.806} \\
\hline
\end{tabular}%
}
\label{tab:wsi_zeroshot}
\end{table*}

\begin{table*}[t!]
\caption{WSI-Level linear probe and weakly supervised MIL-based classification results and comparisons with SOTA CPath models.}
\label{tab:wsi_linear_probe}
\resizebox{\textwidth}{!}{%
\begin{tabular}{c|cc|cc|cc|cc|cc|cc|cc|cc|cc|c}
\hline
 & \multicolumn{2}{|c|}{\textbf{PANDA}} & \multicolumn{2}{c|}{\textbf{HunCRC}} & \multicolumn{2}{c|}{\textbf{CPTAC-NSCLC}} & \multicolumn{2}{c|}{\textbf{Brains30}} & \multicolumn{2}{c|}{\textbf{BRACS}} & \multicolumn{2}{c|}{\textbf{TCGA-RCC}} & \multicolumn{2}{c|}{\textbf{DHMC-RCC}} & \multicolumn{2}{c|}{\textbf{Camelyon17}} & \multicolumn{2}{c|}{\textbf{UBC-Ocean}} & \textbf{Average} \\
 & F & BA & F & BA & F & BA & F & BA & F & BA & F & BA & F & BA & F & BA & F & BA &BA  \\
 \hline
HIPT & 0.687 & 0.654 & 0.552 & 0.588 & 0.822 & 0.801 & 0.702 & 0.677 & 0.539 & 0.556 & 0.846 & 0.822 & 0.841 & 0.822 & 0.743 & 0.708 & 0.766 & 0.706 & 0.682 \\
CTransPath & 0.752 & 0.691 & 0.728 & 0.556 & 0.895 & 0.887 & 0.597 & 0.514 & 0.648 & 0.639 & 0.856 & 0.833 & 0.883 & 0.804 & 0.809 & 0.786 & 0.788 & 0.733 & 0.698 \\
REMEDIS & 0.766 & 0.711 & 0.787 & 0.604 & 0.866 & 0.841 & 0.471 & 0.382 & 0.696 & 0.676 & 0.823 & 0.807 & 0.877 & 0.865 & 0.776 & 0.752 & 0.733 & 0.706 & 0.679 \\
CHIEF & 0.745 & 0.724 & 0.643 & 0.651 & 0.934 & 0.922 & 0.706 & 0.688 & 0.666 & 0.656 & 0.928 & 0.914 & 0.901 & 0.897 & 0.795 & 0.786 & 0.823 & 0.789 & 0.755 \\
DinoPath & 0.682 & 0.706 & 0.657 & 0.687 & 0.863 & 0.855 & 0.771 & 0.755 & 0.702 & 0.687 & 0.901 & 0.881 & 0.918 & 0.901 & 0.835 & 0.823 & 0.844 & 0.821 & 0.767 \\
Phikon & 0.621 & 0.688 & 0.661 & 0.666 & 0.936 & 0.923 & 0.723 & 0.706 & 0.687 & 0.712 & 0.812 & 0.786 & 0.902 & 0.887 & 0.845 & 0.812 & 0.867 & 0.854 & 0.754 \\
Virchow & 0.741 & 0.728 & 0.621 & 0.767 & 0.855 & 0.82 & 0.723 & 0.701 & 0.644 & 0.621 & 0.886 & 0.819 & 0.931 & 0.922 & 0.88 & 0.865 & 0.902 & 0.889 & 0.769 \\
RudolfV & 0.653 & 0.677 & 0.653 & 0.706 & 0.811 & 0.788 & 0.706 & 0.688 & 0.622 & 0.608 & 0.856 & 0.831 & 0.882 & 0.856 & 0.86 & 0.855 & 0.881 & 0.865 & 0.747 \\
UNI & 0.809 & 0.757 & {\ul 0.824} & 0.643 & 0.935 & 0.904 & 0.746 & 0.675 & 0.691 & 0.687 & 0.959 & 0.947 & 0.926 & 0.919 & 0.923 & 0.901 & 0.94 & 0.922 & 0.798 \\
GigaPath & 0.789 & 0.794 & 0.667 & 0.641 & 0.915 & 0.9 & 0.704 & 0.687 & 0.661 & 0.568 & 0.892 & 0.823 & 0.936 & 0.921 & 0.902 & 0.881 & 0.901 & 0.889 & 0.764 \\
PRISM & 0.662 & 0.68 & 0.72 & 0.755 & 0.871 & 0.867 & 0.732 & 0.674 & 0.759 & 0.658 & 0.915 & 0.866 & 0.893 & 0.872 & 0.866 & 0.809 & 0.876 & 0.833 & 0.765 \\
CONCH & 0.733 & 0.702 & 0.721 & 0.681 & 0.902 & 0.881 & 0.715 & 0.644 & 0.748 & 0.723 & 0.91 & 0.871 & 0.866 & 0.856 & 0.902 & 0.871 & 0.867 & 0.844 & 0.769 \\
QuiltNet & 0.744 & 0.722 & 0.722 & 0.702 & 0.9 & 0.887 & 0.666 & 0.655 & 0.725 & 0.718 & 0.882 & 0.877 & 0.864 & 0.851 & 0.833 & 0.809 & 0.866 & 0.807 & 0.759 \\
MRPLIP & 0.816 & 0.786 & 0.701 & 0.688 & \textbf{0.995} & 0.93 & 0.763 & 0.745 & 0.767 & {\ul 0.741} & 0.902 & 0.891 & 0.952 & 0.941 & 0.922 & 0.901 & 0.884 & 0.866 & 0.812 \\
TITAN & {\ul 0.836} & {\ul 0.823} & 0.766 & 0.806 & 0.966 & {\ul 0.95} & {\ul 0.786} & 0.735 & {\ul 0.803} & 0.702 & 0.941 & 0.942 & 0.951 & 0.933 & 0.944 & {\ul 0.923} & {\ul 0.956} & 0.933 & 0.845\\
\hline
MRPT($\textrm{mViT}_{\textbf{\textrm{P}}}\textrm{-}\textrm{C}$) & 0.822 & 0.809 & 0.77 & 0.786 & 0.966 & 0.944 & 0.761 & {\ul 0.789} & 0.801 & 0.76 & {\ul 0.973} & {\ul 0.966} & 0.95 & 0.936 & {\ul 0.96} & 0.94 & 0.982 & {\ul 0.966} & {\ul 0.863} \\
MRPT($\textrm{mViT}_{\textbf{\textrm{R}}}\textrm{-}\textrm{P}$) & 0.833 & 0.811 & 0.781 & {\ul 0.807} & 0.93 & 0.926 & 0.75 & 0.774 & 0.744 & 0.722 & 0.967 & 0.955 & {\ul 0.966} & {\ul 0.954} & 0.945 & {\ul 0.923} & 0.95 & 0.944 & 0.851 \\
Proposed MRPT & \textbf{0.873} & \textbf{0.866} & \textbf{0.836} & \textbf{0.866} & {\ul 0.989} & \textbf{0.981} & \textbf{0.823} & \textbf{0.807} & \textbf{0.844} & \textbf{0.788} & \textbf{0.98} & \textbf{0.976} & \textbf{0.986} & \textbf{0.974} & \textbf{0.97} & \textbf{0.966} & \textbf{0.987} & \textbf{0.972} & \textbf{0.898} \\
\hline
\end{tabular}%
}
\end{table*}

\begin{table*}[t!]
\caption{\textbf{WSI classification using few-shot learning experiments} with $k \in \{1,2,4,8,16,32\}$ shots per class on 
EBRAINS and PANDA datasets. 
Results are given in balanced accuracy of linear probing evaluation or ABMIL training, respectively. 
If a class contains less than $k$ samples, then all samples are used. 
The best result is marked in bold and the second best is underlined.}
\centering
\resizebox{\linewidth}{!}{%
\begin{tabular}{c|l|cccccc}
\toprule
\textbf{Dataset} & \textbf{Models} & $k=1$ & $k=2$ & $k=4$ & $k=8$ & $k=16$ & $k=32$ \\
\midrule
\multirow{15}{*}{\textbf{EBRAINS}}
& Mean pool (CONCH) & $0.354{\pm}0.0267$ & $0.445{\pm}0.0310$ & $0.526{\pm}0.0212$ & $0.589{\pm}0.0147$ & $0.638{\pm}0.0176$ & $0.668{\pm}0.0127$ \\
& ABMIL (CONCH) & $0.388{\pm}0.0417$ & $0.544{\pm}0.0372$ & $0.629{\pm}0.0230$ & $0.671{\pm}0.0195$ & $0.702{\pm}0.0138$ & $0.729{\pm}0.0118$ \\
& GigaPath & $0.292{\pm}0.0284$ & $0.398{\pm}0.0250$ & $0.507{\pm}0.0206$ & $0.589{\pm}0.0190$ & $0.658{\pm}0.0175$ & $0.709{\pm}0.0122$ \\
& PRISM & $0.399{\pm}0.0328$ & $0.484{\pm}0.0192$ & $0.548{\pm}0.0225$ & $0.595{\pm}0.0193$ & $0.628{\pm}0.0173$ & $0.659{\pm}0.0156$ \\
& CHIEF & $0.226{\pm}0.0280$ & $0.319{\pm}0.0278$ & $0.418{\pm}0.0237$ & $0.509{\pm}0.0171$ & $0.578{\pm}0.0158$ & $0.637{\pm}0.0122$ \\
& FOCUS & $0.480{\pm}0.0264$ & $0.556{\pm}0.0230$ & $0.611{\pm}0.0204$ & $0.648{\pm}0.0183$ & $0.678{\pm}0.0161$ & $0.692{\pm}0.0134$ \\
& ViLa-MIL & $0.472{\pm}0.0281$ & $0.549{\pm}0.0221$ & $0.603{\pm}0.0198$ & $0.642{\pm}0.0175$ & $0.674{\pm}0.0149$ & $0.688{\pm}0.0130$ \\
& TOP-MIL & $0.476{\pm}0.0255$ & $0.553{\pm}0.0228$ & $0.606{\pm}0.0192$ & $0.645{\pm}0.0177$ & $0.676{\pm}0.0155$ & $0.691{\pm}0.0127$ \\
& WiKG-MIL & $0.487{\pm}0.0262$ & $0.563{\pm}0.0229$ & $0.616{\pm}0.0186$ & $0.653{\pm}0.0171$ & $0.681{\pm}0.0151$ & $0.698{\pm}0.0128$ \\
& TITAN$_\text{v}$ & $0.489{\pm}0.0261$ & $0.568{\pm}0.0208$ & $0.627{\pm}0.0195$ & $0.672{\pm}0.0184$&$0.708{\pm}0.0154$ &$0.730{\pm}0.0114$ \\
& \textbf{TITAN} & \textbf{$0.535{\pm}0.0303$} & \textbf{$0.609{\pm}0.0255$} & \textbf{$0.660{\pm}0.0206$} & \textbf{$0.695{\pm}0.0149$} & \textbf{$0.724{\pm}0.0146$} & \textbf{$0.733{\pm}0.0127$} \\
& MRPT($\textrm{mViT}_{\textbf{P}}\textrm{-C}$) & $\underline{0.549{\pm}0.0273}$ & $\underline{0.624{\pm}0.0222}$ & $\underline{0.675{\pm}0.0189}$ & $\underline{0.708{\pm}0.0168}$ & $\underline{0.740{\pm}0.0145}$ & $\underline{0.749{\pm}0.0126}$ \\
& MRPT($\textrm{mViT}_{\textbf{R}}\textrm{-P}$) & $0.547{\pm}0.0277$ & $0.620{\pm}0.0225$ & $0.671{\pm}0.0191$ & $0.705{\pm}0.0169$ & $0.737{\pm}0.0148$ & $0.746{\pm}0.0128$ \\
& \textbf{MRPT} & $\textbf{0.567}{\pm}\textbf{0.0269}$ & $\textbf{0.658}{\pm}\textbf{0.0210}$& $\textbf{0.708}{\pm}\textbf{0.0177}$ & $\textbf{0.737}{\pm}\textbf{0.0160}$ & $\textbf{0.764}{\pm}\textbf{0.0139}$ & $\textbf{0.773}{\pm}\textbf{0.0118}$ \\
\midrule
\multirow{15}{*}{\textbf{PANDA}}
& Mean pool (CONCH) & $0.362{\pm}0.0280$ & $0.435{\pm}0.0275$ & $0.512{\pm}0.0226$ & $0.567{\pm}0.0201$ & $0.605{\pm}0.0167$ & $0.628{\pm}0.0142$ \\
& ABMIL (CONCH) & $0.374{\pm}0.0305$ & $0.495{\pm}0.0264$ & $0.560{\pm}0.0213$ & $0.607{\pm}0.0184$ & $0.644{\pm}0.0160$ & $0.662{\pm}0.0137$ \\
& GigaPath & $0.245{\pm}0.0262$ & $0.322{\pm}0.0245$ & $0.401{\pm}0.0203$ & $0.457{\pm}0.0190$ & $0.512{\pm}0.0171$ & $0.548{\pm}0.0150$ \\
& PRISM & $0.412{\pm}0.0273$ & $0.487{\pm}0.0219$ & $0.556{\pm}0.0198$ & $0.602{\pm}0.0179$ & $0.632{\pm}0.0158$ & $0.653{\pm}0.0136$ \\
& CHIEF & $0.234{\pm}0.0237$ & $0.305{\pm}0.0255$ & $0.377{\pm}0.0216$ & $0.432{\pm}0.0188$ & $0.485{\pm}0.0162$ & $0.522{\pm}0.0138$ \\
& FOCUS & $0.472{\pm}0.0249$ & $0.543{\pm}0.0223$ & $0.598{\pm}0.0195$ & $0.634{\pm}0.0172$ & $0.668{\pm}0.0149$ & $0.686{\pm}0.0128$ \\
& ViLa-MIL & $0.463{\pm}0.0260$ & $0.537{\pm}0.0217$ & $0.590{\pm}0.0188$ & $0.627{\pm}0.0171$ & $0.662{\pm}0.0153$ & $0.680{\pm}0.0132$ \\
& TOP-MIL & $0.468{\pm}0.0251$ & $0.540{\pm}0.0220$ & $0.593{\pm}0.0185$ & $0.631{\pm}0.0169$ & $0.664{\pm}0.0146$ & $0.683{\pm}0.0125$ \\
& WiKG-MIL & $0.476{\pm}0.0265$ & $0.549{\pm}0.0225$ & $0.602{\pm}0.0192$ & $0.639{\pm}0.0170$ & $0.671{\pm}0.0150$ & $0.689{\pm}0.0129$ \\
& TITAN$_\text{v}$ & \underline{$0.502{\pm}0.0254$} & \underline{$0.573{\pm}0.0211$} & \underline{$0.624{\pm}0.0186$} & \underline{$0.662{\pm}0.0167$} & \underline{$0.695{\pm}0.0143$} & \underline{$0.712{\pm}0.0119$} \\
& \textbf{TITAN} & \textbf{$0.549{\pm}0.0287$} & \textbf{$0.615{\pm}0.0235$} & \textbf{$0.667{\pm}0.0199$} & \textbf{$0.695{\pm}0.0178$} & \textbf{$0.723{\pm}0.0151$} & \textbf{$0.732{\pm}0.0125$} \\
& MRPT($\textrm{mViT}_{\textbf{P}}\textrm{-C}$) & $\underline{0.563{\pm}0.0271}$ & $\underline{0.630{\pm}0.0227}$ & $\underline{0.681{\pm}0.0188}$ & $\underline{0.711{\pm}0.0169}$ & $\underline{0.740{\pm}0.0145}$ & $\underline{0.749{\pm}0.0122}$ \\
& MRPT($\textrm{mViT}_{\textbf{R}}\textrm{-P}$) & $0.561{\pm}0.0275$ & $0.625{\pm}0.0230$ & $0.677{\pm}0.0192$ & $0.707{\pm}0.0170$ & $0.737{\pm}0.0146$ & $0.746{\pm}0.0123$ \\
& \textbf{MRPT} & $\textbf{0.582}{\pm}\textbf{0.0259}$ & $\textbf{0.659}{\pm}\textbf{0.0213}$ & $\textbf{0.713}{\pm}\textbf{0.0178}$ & $\textbf{0.735}{\pm}\textbf{0.0159}$ &$\textbf{0.761}{\pm}\textbf{0.0137}$ & $\textbf{0.770}{\pm}\textbf{0.0114}$ \\
\bottomrule
\end{tabular}
}

\label{tab_fsl}
\end{table*}

\section{MRPT Inference and Downstream Analysis Tasks}
\subsection{WSI-Level Inference}
\begin{enumerate}
    \item \textbf{WSI Zero-shot Transferability.} We evaluated the WSI zero-shot classification performance of the pre-trained MRPT-LLaVA model (vision and text encoders from Stage~I) following established protocols~\cite{ding2024multimodal, lu2024visual, ikezogwo2024quilt}. For each WSI, we extracted its multi-resolution embedding using the MRPT encoder and computed its cosine similarity with class-specific textual prompts encoded by the text encoder.
All embeddings were $\ell_{2}$-normalized before similarity computation, and the label was assigned based on the highest similarity score.
We adopted dataset-specific testing prompts consistent with TITAN~\cite{ding2024multimodal}, CONCH~\cite{lu2024visual}, and QuiltNet~\cite{ikezogwo2024quilt}.
Please see our zero-shot WSI classification results and comparisons in Fig. 4 (d) of the manuscript and Table \ref{tab:wsi_zeroshot} for more detailed comparisons.        

\item \textbf{WSI Linear Probe Evaluations.}  Linear probing is a widely used technique for evaluating the quality of feature representations learned by a neural network \cite{ding2024multimodal, chen2024towards}. 
It involves training a simple linear classifier—such as logistic regression—on top of the features extracted from a pre-trained model, while keeping the network’s parameters frozen. Only the weights of the linear classifier are updated during training. This approach measures how effectively the pre-trained model encodes meaningful and discriminative information: if a linear classifier achieves strong performance using the extracted features, it indicates that the network has learned rich and transferable representations.
To evaluate the transfer capabilities and representation quality of the MRPT encoder, we adopted recent work in representation learning using self-supervised frameworks and performed linear probing (logistic regression) \cite{ding2024multimodal, chen2024towards}. 
For linear probing, we minimized cross-entropy loss using the scikit-learn L-BFGS solver with $\ell_{2}$-regularization.
The maximum number of L-BFGS iterations was set to 500. 
For datasets without a validation set, e.g., in small datasets or in few-shot experiments, we choosed default values of $\ell_{2}$ = 1 with 1,000 iterations.
Please see the WSI linear probe evaluation results and comparisons in Fig. 4 (e)  of the manuscript and Table \ref{tab:wsi_linear_probe} for more detailed comparisons.

\item \textbf{WSI Few-shot Learning.} We also evaluated the few-shot classification performance of our MRPT model by varying the number of shots, e.g., $k \in \{1, 2, 4, 8, 16, 32\}$. 
For each $k$, we randomly selected $k$ samples per class, or all available samples if a class contains fewer than $k$. 
Following prior works~\cite{wang2019simpleshot}, we assessed the few-shot learning capability of our MRPT representation. 
To enable fair comparison with supervised baselines such as ABMIL \cite{ilse2018attention}, we conduct few-shot classification using linear probing and weakly supervised MIL. 
Since no validation set is available in the few-shot setting, we follow the default \texttt{scikit-learn} configuration, employing an $\ell_2$ regularization strength of 1 and up to 1{,}000 iterations of the L-BFGS optimizer.

Table \ref{tab_fsl} shows WSI classification results and comparisons with SOTA CPath models including CONCH \cite{lu2024visual}, GigaPath \cite{xu2024whole}, PRISM \cite{shaikovski2024prism}, CHIEF\cite{wang2024pathology}, FOCUS \cite{guo2025focus}, Vila-MIL\cite{shi2024vila}, TOP-MIL\cite{qu2023rise}, and WiKG-MIL \cite{li2024dynamic} using few-shot learning experiments on three datasets.
 MRPT achieves better few-shot performance compared to the SOTA methods because its multi-resolution hierarchical design captures both global tissue context and fine cellular details, producing richer and more transferable representations than single-scale models. 
 The Consecutive Cross-Resolution Attention (CCRA) mechanism ensures semantic consistency across magnifications, leading to well-structured and discriminative embeddings even with limited samples. By mirroring the diagnostic reasoning of pathologists, MRPT introduces a strong inductive bias that enhances generalization under data scarcity. Consequently, it learns robust, context-aware features that enable efficient adaptation to new pathology tasks with minimal supervision.

\item \textbf{Weakly Supervised MIL (WSMIL) for WSI Classification.} We also conducted a weakly supervised WSI classification to evaluate the effectiveness of the multi-resolution patch-level representation extraction from $\textrm{mViT}_{\textbf{\textrm{P}}}\textrm{-}\textrm{C}$ encoder, and region-level representation extracted from $\textrm{mViT}_{\textrm{\textbf{R}}}\textrm{-}\textrm{P}$ encoder.
For this purpose, the multi-resolution patch-level features using $\textrm{mViT}_{\textbf{\textrm{P}}}\textrm{-}\textrm{C}$ encoder are subsequently aggregated using the ABMIL method~\cite{ilse2018attention} for MIL-based WSI classification \cite{wang2022transformer, vorontsov2024foundation, chen2024towards}.
We dub this variant as MRPT ($\textrm{mViT}_{\textbf{\textrm{P}}}\textrm{-}\textrm{C}$).
MRPT ($\textrm{mViT}_{\textbf{\textrm{P}}}\textrm{-}\textrm{C}$) model training is performed using the AdamW optimizer with a cosine learning rate scheduler, an initial learning rate of $1 \times 10^{-4}$, cross-entropy loss, and a maximum of 20 epochs. 
Similar to MRPT ($\textrm{mViT}_{\textbf{\textrm{P}}}\textrm{-}\textrm{C}$) we also pre-trained MRPT ($\textrm{mViT}_{\textrm{\textbf{R}}}\textrm{-}\textrm{P}$) and dub this variant as MRPT ($\textrm{mViT}_{\textrm{\textbf{R}}}\textrm{-}\textrm{P}$).
To ensure fair comparison, we adhered to the experimental protocols established by existing SOTA approaches for WSI classification \cite{chen2024towards}. 
In cases where official data folds are not provided, we adopted case-stratified and label-stratified train–validation–test splits as recommended by UNI~\cite{chen2024towards}.
Please see WSMIL results and comparisons in Fig. 4 (e) of the manuscript and Table \ref{tab:wsi_linear_probe} for more detailed comparisons.

\item \textbf{Dense Spatial Reasoning.} MRPT is not limited to global WSI tasks: patch-level classification and VQA results are already presented. 
We further include region-level retrieval results (Table~\ref{table_dense_reasoning}), where MRPT embeddings show strong intra-class consistency and outperform SOTA models.


\end{enumerate}

 \begin{table}[t!]
\captionof{table}{Retrieval results. ACC:Accuracy. MV:Majority Voting.}
\centering
\makebox[\linewidth]{
\scalebox{0.80}{
\begin{tabular}{lcc|cc|cc|cc}
\hline
\multirow{2}{*}{Metric} 
& \multicolumn{2}{c|}{HIPT}
& \multicolumn{2}{c|}{UNI}
& \multicolumn{2}{c|}{MRPT ($\textrm{mViT}_{\textbf{P}}\textrm{-}\textrm{C}$)}
& \multicolumn{2}{c}{MRPT ($\textrm{mViT}_{\textbf{R}}\textrm{-}\textrm{P}$)} \\
\cmidrule(lr){2-3} \cmidrule(lr){4-5} \cmidrule(lr){6-7} \cmidrule(lr){8-9}
& TissueNet  & UniToPatho 
& TissueNet & UniToPatho 
& TissueNet & UniToPatho 
& TissueNet & UniToPatho \\
\hline
ACC\text{@}1 & 0.688 & 0.633 & 0.766 & 0.731 & \textbf{0.822} & \textbf{0.816} & \underline{0.820} & \underline{0.798} \\
ACC\text{@}3 & 0.765 & 0.728 & 0.809 & 0.786 & \textbf{0.881} & \textbf{0.853} & \underline{0.876} & \underline{0.832} \\
ACC\text{@}5 & 0.855 & 0.812 & 0.883 & 0.859 & \textbf{0.964} & \textbf{0.936} & \underline{0.953} & \underline{0.928} \\
mMV\text{@}5 & 0.703 & 0.646 & 0.782 & 0.756 & \textbf{0.842} & \textbf{0.821} & \underline{0.838} & \underline{0.804} \\
\hline
\end{tabular}
}
}
\label{table_dense_reasoning}
\end{table}

\begin{table}[t!]
\caption{WSI-Level VQA results and comparisons with SOTA CPath MLLMs across four datasets.}
\label{tab:wsi_vqa}
\resizebox{\columnwidth}{!}{%
\begin{tabular}{l|cccc}
\hline
 & \textbf{SlideBench-VQA} & \textbf{SlideBench-VQA} & \textbf{WSI-VQA} & \textbf{WSI-Bench} \\
 & \textbf{(TCGA)} & \textbf{(BCNB)} & & \textbf{(Close-ended)} \\
 \hline
SlideChat & 0.811 & 0.541 & \underline{0.601} & 0.920 \\
Quilt-LLaVA & 0.393 & 0.415 & 0.354 & 0.932 \\
LLaVAMed & 0.373 & 0.112 & 0.187 & 0.321 \\
MedDr & 0.642 & 0.354 & 0.509 & 0.511 \\
GPT-4O & 0.340 & 0.08 & 0.140 & 0.622 \\
WSI-LLaVA & \underline{0.866} &\underline{ 0.553} & 0.546 & \underline{0.938} \\
WSI-VQA & 0.821 & 0.113 & 0.469 & 0.751 \\
MRPT-LLaVA & \textbf{0.944}&\textbf{ 0.622} & \textbf{0.688}&\textbf{0.977} \\
\hline
\end{tabular}%
}
\vspace{-2em}
\end{table}

\subsection{Patch-Level Inference}

\begin{enumerate}
    \item \textbf{Patch-level Zero-shot Transferability.} Similar to the WSI-level zero-shot classification, we conduct patch-level zero-shot classification.
    Instead of extracting multi-resolution WSI-level representation using the MRPT model, we extract multi-resolution patch-level and region-level representations using $\textrm{mViT}_{\textbf{\textrm{P}}}\textrm{-}\textrm{C}$ and $\textrm{mViT}_{\textrm{\textbf{R}}}\textrm{-}\textrm{P}$ encoders and then compute their cosine similarities with class-specific textual prompts encoded by the text encoder.
    We adopt dataset-specific testing patch-level prompts consistent with CONCH~\cite{lu2024visual}, and QuiltNet~\cite{ikezogwo2024quilt}.
    We dub these variants as $\textrm{mViT}_{\textbf{\textrm{P}}}\textrm{-}\textrm{C}$-LLaVA and $\textrm{mViT}_{\textrm{\textbf{R}}}\textrm{-}\textrm{P}$-LLaVA.
    Please see the results and comparisons in Fig. 4(a) of the manuscript and Table \ref{tab:patch_level_zero_shot} for more detailed comparisons.

    \item \textbf{Patch-level Linear Probe Evaluations.} Similar to the WSI-level linear probe evaluations, we also evaluated the patch-level and region-level transfer capabilities and representation quality of the $\textrm{mViT}_{\textbf{\textrm{P}}}\textrm{-}\textrm{C}$ and $\textrm{mViT}_{\textrm{\textbf{R}}}\textrm{-}\textrm{P}$ encoders. 
    We performed linear (logistic regression) probing by minimizing cross-entropy loss using the scikit-learn L-BFGS solver with $\ell_{2}$-regularization.
The maximum number of L-BFGS iterations is set to 500. 
Please see our patch-level linear probe evaluation results and comparisons in Fig. 4(b) of the manuscript and Table \ref{tab:patch_level_linear_probe} for more detailed comparisons.

\end{enumerate}

\section{Patch-level VQA Task}
Patch-level VQA results and comparisons with CPath-specific and general-purpose models across four patch-level datasets are shown in Table \ref{tab:patch_level_vqa_tasks} and Table \ref{tab:omni_pathmmu}.
Across all datasets, our proposed patch- and region-level models outperform SOTA models by a significant margin, underscoring the significance of multi-resolution representation learned using SSL.

\section{WSI-level VQA Task}
WSI-level VQA results and comparisons with CPath-specific MLLMs across four datasets are shown in Table \ref{tab:wsi_vqa}.
Across all datasets, MRPT-LLaVA outperforms SOTA CPath models by a significant margin, underscoring the significance of multi-resolution representation learned using SSL.

\begin{table*}[t!]
\caption{Patch-level linear probe evaluation results and comparisons with SOTA CPath models.}
\resizebox{\textwidth}{!}{%
\begin{tabular}{c | lll | lll | lll | lll | lll | lll | lll | lll | lll | lll | lll | lll | lll}
\hline
 & \multicolumn{3}{c|}{WSSS4LUAD} & \multicolumn{3}{c|}{SICAP} & \multicolumn{3}{c|}{DigestPath} & \multicolumn{3}{c|}{CRC100K} & \multicolumn{3}{c|}{RenalCall} & \multicolumn{3}{c|}{MHIST} & \multicolumn{3}{c|}{Pcam} & \multicolumn{3}{c|}{Osteo} & \multicolumn{3}{c|}{BACH} & \multicolumn{3}{c|}{SkinCancer} & \multicolumn{3}{c|}{LC-Lung} & \multicolumn{3}{c|}{LC-Colon} & \multicolumn{3}{c}{Databiox}  \\
 & F & A & BA & F & A & BA & F & A & BA & F & A & BA & F & A & BA & F & A & BA & F & A & BA & F & A & BA & F & A & BA & F & A & BA & F & A & BA & F & A & BA & F & A & BA \\
  \hline
HIPT (ViT\_\{256\}-16) & 0.811 & 0.851 & 0.842 & 0.255 & 0.271 & 0.266 & 0.772 & 0.802 & 0.783 & 0.922 & 0.933 & 0.91 & 0.466 & 0.503 & 0.481 & 0.512 & 0.535 & 0.541 & 0.654 & 0.702 & 0.688 & 0.423 & 0.481 & 0.455 & 0.423 & 0.471 & 0.451 & 0.405 & 0.433 & 0.421 & 0.725 & 0.806 & 0.756 & 0.771 & 0.802 & 0.786 & 0.351 & 0.396 & 0.381 \\
CTransPath & 0.857 & 0.871 & 0.844 & 0.747 & 0.681 & 0.678 & 0.688 & 0.702 & 0.661 & 0.867 & 0.871 & 0.845 & 0.521 & 0.544 & 0.529 & 0.826 & 0.841 & 0.811 & {\ul 0.935} & \textbf{0.966} & 0.911 & 0.48 & 0.522 & 0.502 & 0.872 & 0.9 & 0.875 & 0.485 & 0.502 & 0.501 & 0.722 & 0.75 & 0.747 & 0.822 & 0.844 & 0.833 & 0.511 & 0.533 & 0.52 \\
REMEDIS & 0.782 & 0.79 & 0.769 & 0.811 & 0.821 & 0.806 & 0.722 & 0.731 & 0.726 & 0.802 & 0.831 & 0.787 & 0.57 & 0.611 & \textbf{0.97} & 0.807 & 0.813 & 0.781 & 0.822 & 0.831 & 0.805 & 0.511 & 0.531 & 0.513 & 0.864 & 0.861 & 0.863 & 0.5 & 0.522 & 0.513 & 0.751 & 0.787 & 0.766 & 0.788 & 0.803 & 0.792 & 0.487 & 0.541 & 0.524 \\
CHIEF & 0.828 & 0.812 & 0.844 & 0.783 & 0.778 & 0.771 & 0.8 & 0.812 & 0.788 & 0.856 & 0.87 & 0.844 & 0.605 & 0.644 & 0.61 & 0.813 & 0.82 & 0.791 & 0.851 & 0.86 & 0.833 & 0.533 & 0.56 & 0.544 & 0.847 & 0.857 & 0.863 & 0.655 & 0.671 & 0.661 & 0.702 & 0.712 & 0.71 & 0.756 & 0.772 & 0.767 & 0.523 & 0.556 & 0.547 \\
DinoPath & 0.833 & 0.855 & 0.843 & 0.802 & 0.811 & 0.781 & 0.822 & 0.855 & 0.831 & 0.831 & 0.853 & 0.84 & 0.633 & 0.671 & 0.663 & 0.788 & 0.806 & 0.792 & 0.83 & 0.845 & 0.81 & 0.506 & 0.526 & 0.511 & 0.855 & 0.86 & 0.851 & 0.623 & 0.643 & 0.633 & 0.8 & 0.822 & 0.814 & 0.822 & 0.843 & 0.833 & 0.611 & 0.631 & 0.624 \\
Phikon & 0.802 & 0.833 & 0.812 & 0.788 & 0.803 & 0.772 & 0.755 & 0.791 & 0.77 & 0.803 & 0.823 & 0.822 & 0.601 & 0.631 & 0.62 & 0.824 & 0.834 & 0.83 & 0.815 & 0.837 & 0.836 & 0.454 & 0.482 & 0.479 & 0.802 & 0.823 & 0.82 & 0.605 & 0.626 & 0.616 & 0.811 & 0.844 & 0.833 & 0.803 & 0.82 & 0.812 & 0.65 & 0.675 & 0.662 \\
Virchow & 0.873 & 0.881 & 0.866 & 0.873 & 0.881 & 0.855 & 0.844 & 0.851 & 0.828 & {\ul 0.968} & 0.97 & 0.96 & 0.661 & 0.701 & 0.68 & 0.836 & 0.851 & 0.831 & 0.933 & 0.942 & \textbf{0.933} & 0.603 & 0.63 & 0.613 & 0.92 & 0.916 & 0.915 & 0.688 & 0.702 & 0.694 & 0.788 & 0.8 & 0.792 & 0.855 & 0.876 & 0.872 & 0.751 & 0.779 & 0.768 \\
RudolfV & 0.79 & 0.811 & 0.773 & 0.791 & 0.816 & 0.781 & 0.75 & 0.778 & 0.76 & 0.811 & 0.82 & 0.831 & 0.643 & 0.65 & 0.659 & 0.803 & 0.826 & 0.816 & 0.796 & 0.816 & 0.815 & 0.58 & 0.601 & 0.606 & 0.847 & 0.866 & 0.85 & 0.582 & 0.622 & 0.603 & 0.755 & 0.777 & 0.76 & 0.897 & 0.922 & 0.914 & 0.73 & 0.764 & 0.746 \\
UNI & 0.835 & 0.843 & 0.831 & 0.841 & 0.851 & 0.826 & 0.851 & 0.877 & 0.861 & 0.875 & 0.901 & 0.874 & 0.701 & 0.723 & 0.73 & 0.881 & 0.902 & 0.856 & 0.93 & 0.946 & 0.901 & 0.622 & 0.66 & 0.646 & 0.926 & 0.944 & 0.933 & 0.702 & 0.736 & 0.722 & 0.881 & 0.896 & 0.885 & 0.931 & 0.951 & 0.946 & 0.788 & 0.802 & 0.793 \\
GigaPath & 0.872 & 0.892 & 0.86 & 0.861 & 0.873 & 0.845 & 0.841 & 0.867 & 0.861 & 0.942 & 0.953 & 0.929 & 0.721 & 0.75 & 0.743 & 0.879 & 0.881 & 0.851 & 0.931 & 0.951 & {\ul 0.925} & 0.61 & 0.645 & 0.634 & {\ul 0.947} & 0.961 & 0.942 & 0.687 & 0.71 & 0.708 & 0.902 & 0.916 & 0.92 & 0.955 & {\ul 0.976} & \textbf{0.974} & 0.802 & 0.821 & 0.811 \\
\hline
Proposed $\textrm{mViT}_{\textrm{\textbf{P}}}\textrm{-}\textrm{C}$ & \textbf{0.933} & \textbf{0.922} & \textbf{0.891} & {\ul 0.906} & \textbf{0.933} & \textbf{0.881} & \textbf{0.923} & \textbf{0.955} & {\ul 0.915} & \textbf{0.976} & \textbf{0.98} & \textbf{0.973} & {\ul 0.803} & {\ul 0.833} & {\ul 0.842} & \textbf{0.922} & \textbf{0.955} & \textbf{0.941} & \textbf{0.941} & {\ul 0.964} & 0.92 & \textbf{0.802} & \textbf{0.833} & \textbf{0.814} & 0.94 & {\ul 0.966} & {\ul 0.961} & {\ul 0.877} & {\ul 0.902} & {\ul 0.881} & \textbf{0.955} & \textbf{0.966} & \textbf{0.953} & \textbf{0.97} & \textbf{0.986} & {\ul 0.97} & \textbf{0.844} & \textbf{0.863} & \textbf{0.851} \\
Proposed $\textrm{mViT}_{\textrm{\textbf{R}}}\textrm{-}\textrm{P}$ & {\ul 0.931} & {\ul 0.916} & {\ul 0.886} & \textbf{0.921} & {\ul 0.932} & {\ul 0.873} & {\ul 0.916} & {\ul 0.933} & \textbf{0.922} & 0.966 & {\ul 0.975} & {\ul 0.972} & \textbf{0.821} & \textbf{0.866} & 0.833 & {\ul 0.915} & {\ul 0.94} & {\ul 0.937} & 0.922 & 0.931 & 0.923 & {\ul 0.788} & {\ul 0.801} & {\ul 0.796} & \textbf{0.955} & \textbf{0.98} & \textbf{0.977} & \textbf{0.902} & \textbf{0.933} & \textbf{0.92} & {\ul 0.946} & {\ul 0.951} & {\ul 0.949} & {\ul 0.953} & 0.974 & 0.963 & {\ul 0.824} & {\ul 0.857} & {\ul 0.845}\\
\hline
\end{tabular}%
}
\label{tab:patch_level_linear_probe}
\end{table*}

\begin{table*}[t!]
\caption{Patch-Level zero-shot classification results and comparisons with SOTA CPath models.}
\resizebox{\linewidth}{!}{%
\begin{tabular}{c | ll | ll | ll | ll | ll | ll | ll | ll | ll | ll | ll | ll | ll}
\hline
 & \multicolumn{2}{c|}{WSSS4LUAD} & \multicolumn{2}{c|}{SICAP} & \multicolumn{2}{c|}{DigestPath} & \multicolumn{2}{c|}{CRC100K} & \multicolumn{2}{c|}{RenalCell} & \multicolumn{2}{c|}{MHIST} & \multicolumn{2}{c|}{Pcam} & \multicolumn{2}{c|}{Osteo} & \multicolumn{2}{c|}{BACH} & \multicolumn{2}{c|}{SkinCancer} & \multicolumn{2}{c|}{LC-Lung} & \multicolumn{2}{c|}{LC-Colon} & \multicolumn{2}{c}{Databiox} \\ 
 & F & A & F & A & F & A & F & A & F & A & F & A & F & A & F & A & F & A & F & A & F & A & F & A & F & A \\
 \hline
CLIP & 0.196 & 0.322 & 0.14 & 0.254 & 0.123 & 0.203 & 0.185 & 0.403 & 0.176 & 0.256 & 0.288 & 0.401 & 0.223 & 0.55 & 0.244 & 0.539 & 0.201 & 0.343 & 0.231 & 0.194 & 0.321 & 0.704 & 0.351 & 0.811 & 0.121 & 0.302 \\
PLIP & 0.408 & 0.451 & 0.226 & 0.426 & 0.782 & 0.811 & 0.687 & 0.528 & 0.314 & 0.355 & 0.431 & 0.477 & 0.362 & 0.518 & 0.409 & 0.731 & 0.336 & 0.529 & 0.355 & 0.425 & 0.502 & 0.879 & 0.524 & 0.902 & 0.331 & 0.501 \\
PathCLIP & 0.823 & 0.851 & 0.461 & 0.483 & 0.823 & 0.855 & 0.5 & 0.553 & 0.681 & 0.711 & 0.588 & 0.622 & 0.688 & 0.725 & 0.622 & 0.692 & 0.401 & 0.468 & 0.301 & 0.351 & 0.851 & 0.889 & 0.855 & 0.943 & 0.481 & 0.544 \\
CPLIP & 0.791 & 0.803 & 0.388 & 0.351 & 0.856 & 0.841 & 0.681 & 0.723 & 0.491 & 0.551 & 0.541 & 0.561 & 0.531 & 0.588 & 0.544 & 0.573 & 0.528 & 0.588 & 0.476 & 0.502 & 0.752 & 0.788 & 0.791 & 0.822 & 0.462 & 0.502 \\
CONCH & 0.59 & 0.622 & 0.241 & 0.321 & 0.866 & 0.881 & 0.542 & 0.591 & 0.467 & 0.502 & 0.502 & 0.531 & 0.543 & 0.602 & 0.535 & 0.603 & 0.483 & 0.533 & 0.433 & 0.467 & 0.735 & 0.753 & 0.822 & 0.851 & 0.416 & 0.451 \\
QuiltNet & 0.673 & 0.705 & 0.363 & 0.373 & 0.803 & 0.811 & 0.553 & 0.495 & 0.487 & 0.533 & 0.521 & 0.559 & 0.551 & 0.587 & 0.521 & 0.538 & 0.485 & 0.438 & 0.367 & 0.464 & 0.73 & 0.8 & 0.806 & 0.91 & 0.397 & 0.435 \\
MR-PLIP & 0.831 & 0.851 & 0.487 & 0.513 & 0.902 & 0.924 & 0.832 & 0.851 & 0.531 & 0.577 & 0.596 & 0.622 & 0.604 & 0.644 & 0.606 & 0.655 & 0.587 & 0.621 & 0.479 & 0.516 & 0.824 & 0.855 & 0.851 & 0.871 & 0.498 & 0.544 \\
OmniPath &- & 0.871 & - & 0.631 & - & - & - & 0.78 & - & - & - & - & - & \textbf{0.959} & - & \textbf{0.807} & - & 0.723 & 0.691 & 0.742 & - & \textbf{0.971} & - & \textbf{1.00} & - & - \\
PathGenCLIP & 0.788 & 0.822 & 0.597 & 0.635 & 0.792 & 0.823 & 0.755 & 0.78 & 0.451 & 0.499 & 0.521 & 0.564 & 0.822 & 0.882 & 0.701 & 0.746 & 0.655 & 0.715 & 0.679 & 0.706 & 0.851 & 0.898 & \textbf{0.955} & 0.993 & 0.451 & 0.506 \\
MI-Zero & 0.582 & 0.633 & 0.235 & 0.291 & 0.861 & 0.875 & 0.536 & 0.571 & 0.455 & 0.491 & 0.497 & 0.523 & 0.51 & 0.566 & 0.498 & 0.533 & 0.403 & 0.451 & 0.392 & 0.441 & 0.716 & 0.755 & 0.802 & 0.844 & 0.401 & 0.497 \\
KEP & 0.743 & 0.799 & 0.302 & 0.339 & 0.826 & 0.833 & 0.622 & 0.644 & 0.501 & 0.522 & 0.481 & 0.502 & 0.655 & 0.684 & 0.451 & 0.478 & 0.526 & 0.55 & 0.427 & 0.463 & 0.877 & 0.916 & 0.902 & 0.989 & 0.624 & 0.702 \\
\hline
Proposed $\textrm{mViT}_{\textrm{\textbf{P}}}\textrm{-}\textrm{C}$-LLaVA & \textbf{0.877} & \textbf{0.902} & \textbf{0.633} & \textbf{0.661} & \textbf{0.955} & \textbf{0.971} & \textbf{0.966} & \textbf{0.971} & \textbf{0.733} & \textbf{0.766} & \textbf{0.651} & \textbf{0.702} & \textbf{0.881} & {\ul 0.933} & \textbf{0.766} & {\ul 0.789} & \textbf{0.802} & \textbf{0.833} & \textbf{0.756} & \textbf{0.833} & \textbf{0.944} & {\ul 0.97} & {\ul 0.911} & {\ul 0.955} & \textbf{0.771} & \textbf{0.822} \\
Proposed $\textrm{mViT}_{\textrm{\textbf{R}}}\textrm{-}\textrm{P}$-LLaVA & {\ul 0.871} & {\ul 0.895} & {\ul 0.621} & {\ul 0.642} & {\ul 0.932} & {\ul 0.961} & {\ul 0.943} & {\ul 0.973} & {\ul 0.721} & {\ul 0.75} & {\ul 0.648} & {\ul 0.697} & {\ul 0.874} & 0.918 & {\ul 0.743} & 0.766 & {\ul 0.781} & {\ul 0.816} & {\ul 0.731} & {\ul 0.8} & {\ul 0.92} & 0.966 & 0.903 & 0.931 & {\ul 0.759} & {\ul 0.788} \\ \hline
\end{tabular}%
}
\label{tab:patch_level_zero_shot}
\end{table*}

\begin{table*}[t!]
\caption{Patch-level Visual Questioning Answering (VQA) results and comparisons with CPath-specific and general-purpose models.}
\resizebox{\textwidth}{!}{%
\begin{tabular}{lccccccc}
\hline
 & PathMMU (Tiny) & PathhMMU (ALL) & Quilt-VQA (Open) & Quilt-VQA (Closed) & PathVQA (Open) & PathVQA (Closed) & PMC-VQA (Closed) \\
\hline
GPT-4V & 0.539 & 0.498 & 0.334 & 0.509 & 0.112 & 0.523 & 0.201 \\
LLaVA & 0.388 & 0.376 & 0.558 & 0.577 & 0.116 & 0.54 & 0.188 \\
Qwen-VL-MAX & 0.492 & 0.459 & 0.486 & 0.433 & 0.195 & 0.446 & 0.224 \\
Gemini Pro-V & 0.428 & 0.427 & 0.472 & 0.498 & 0.109 & 0.498 & 0.152 \\
Quilt-LLaVA & 0.456 & 0.415 & 0.553 & 0.688 & 0.15 & 0.586 & 0.285 \\
MedDr & 0.322 & 0.264 & 0.312 & 0.407 & 0.132 & 0.422 & 0.196 \\
LLaVA-Med & 0.253 & 0.262 & 0.548 & 0.612 & 0.119 & 0.523 & 0.013 \\
PathGen-LLaVA & 0.601 & 0.584 & 0.496 & 0.557 & 0.22 & 0.556 & 0.176 \\
CPathOmni & 0.724 & 0.722 & - & - & - & - & - \\
\hline
Proposed $\textrm{mViT}_{\textbf{\textrm{P}}}\textrm{-}\textrm{C}$-LLaVA & \textbf{0.788} & \textbf{0.809} & \textbf{0.678} & {\ul 0.755} & \textbf{0.443} & \textbf{0.798} & \textbf{0.506} \\
Proposed $\textrm{mViT}_{\textrm{\textbf{R}}}\textrm{-}\textrm{P}$-LLaVA & {\ul 0.756} & {\ul 0.785} & {\ul 0.634} & \textbf{0.776} & {\ul 0.405} & {\ul 0.772} & {\ul 0.487}\\
\hline
\end{tabular}%
}
\label{tab:patch_level_vqa_tasks}
\end{table*}

\begin{table*}[t!]
\caption{Patch-level VQA results and comparisons with CPath-specific and general-purpose models on PathMMU test set.}
\resizebox{\textwidth}{!}{%
\begin{tabular}{lcccccccccccc}
\hline 
\multirow{2}{*}{} & \multicolumn{2}{c}{Test Overall} & \multicolumn{2}{c}{PubMed} & \multicolumn{2}{c}{SocialPath} & \multicolumn{2}{c}{EduContent} & \multicolumn{2}{c}{Atlas} & \multicolumn{2}{c}{PathCLS} \\
 & Tiny & ALL & Tiny & ALL & Tiny & All & Tiny & All & Tiny & ALL & Tiny & ALL\\
 & (1156) & (9677) & (281) & (3068) & (235) & (1855) & (255) & (1938) & (208) & (1007) & (177) & (1809) \\
\hline 
Expert performance & 71.8 & - & 72.9 & - & 71.5 & - & 69.0 & - & 68.3 & - & 78.9 & - \\
\hline 
\multicolumn{13}{c}{\textbf{General Large Multimodal Models}} \\
\hline 
InstructBLIP-FLAN-T5-XXL & 34.3 & 33.9 & 39.1 & 37.2 & 33.6 & 34.3 & 34.5 & 36.0 & 38.5 & 39.3 & 22.6 & 22.7 \\
 LLaVA-1.5-13B & 38.8 & 37.6 & 44.5 & 41.0 & 40.4 & 40.4 & 34.1 & 39.4 & 47.1 & 44.3 & 24.9 & 23.5 \\
 Qwen-VL-MAX & 49.2 & 45.9 & 53.0 & 50.9 & 53.6 & 49.3 & 52.2 & 47.9 & 51.4 & 49.8 & 30.5 & 29.6 \\
 Gemini Pro Vision & 42.8 & 42.7 & 43.8 & 44.9 & 42.4 & 42.0 & 43.5 & 43.7 & 49.5 & 49.4 & 32.8 & 34.7 \\
 GPT-4V-1106 & 53.9 & 49.8 & 59.4 & 53.5 & 58.7 & 53.9 & 60.4 & 53.6 & 48.1 & 52.8 & 36.2 & 33.8 \\
\hline 
\multicolumn{13}{c}{\textbf{Pathology-specific Large Multimodal Models}} \\
\hline
LLaVA-Med & 25.3 & 26.2 & 28.5 & 27.7 & 28.9 & 27.3 & 22.7 & 27.2 & 22.6 & 30.7 & 22.6 & 20.3 \\
Quilt-LLaVA & 45.6 & 41.5 & 47.3 & 42.6 & 46.4 & 46.6 & 51.8 & 45.3 & 46.2 & 42.7 & 32.2 & 29.2 \\
PathGen-LLaVA & 60.1 & 58.4 & 60.1 & 60.1 & 60.9 & 58.8 & 60.8 & 60.7 & 63.5 & 64.9 & 54.2 & 48.9 \\
CPath-Omni & 72.4 & 72.2 & 74.0 & 69.9 & 76.6 & 71.8 & 69.8 & 70.6 & 65.9 & 70.6 & 75.7 & 79.0 \\
\hline
Proposed $\textrm{mViT}_{\textbf{\textrm{P}}}\textrm{-}\textrm{C}$-LLaVA &\textbf{78.8}&\textbf{80.9}&\textbf{80.4}&\textbf{82.6}&\textbf{83.4}&\textbf{85.6}&\textbf{76.0}&\textbf{78.0}&\textbf{71.7}&\textbf{73.7}&\textbf{82.4}&\textbf{84.6}\\
Proposed $\textrm{mViT}_{\textrm{\textbf{R}}}\textrm{-}\textrm{P}$-LLaVA &\underline{75.6}&\underline{78.5}&\underline{77.2}&\underline{80.2}&\underline{80.0}&\underline{83.0}&\underline{72.8}&\underline{75.6}&\underline{68.8}&\underline{71.4}&\underline{79.0}&\underline{82.1}\\
\hline
\end{tabular}}
\label{tab:omni_pathmmu}
\end{table*}

\section{Pre-Training Dataset}

We pre-train \textbf{MRPT} on 36{,}000 WSIs in total, comprising 30{,}000 WSIs from TCGA diverse primary sites  \cite{hutter2018cancer} and 6{,}000 WSIs from CPTAC \cite{Edwards2015CPTAC}.
The TCGA primary cancer sites include Adrenocortical carcinoma (ACC), Bladder urothelial carcinoma (BLCA), Breast invasive carcinoma (BRCA), Cervical squamous cell carcinoma and endocervical adenocarcinoma (CESC), Cholangiocarcinoma (CHOL), Colorectal adenocarcinoma (COADREAD), Esophageal carcinoma (ESCA), Glioblastoma Multiforme and Brain Lower Grade Glioma (GBMLGG), Head and Neck squamous cell carcinoma (HNSC), Kidney renal papillary cell carcinoma (KIR), Liver hepatocellular carcinoma (LIHC), Lung adenocarcinoma and Lung squamous cell carcinoma (LUADLUSC), Lymphoid neoplasm (LYM), Mesothelioma (MESO), Ovarian serous cystadenocarcinoma (OV), Pancreatic adenocarcinoma (PAAD), Prostate adenocarcinoma (PRAD), Sarcoma (SARC), Skin cutaneous melanoma (SKCM), Stomach adenocarcinoma (STAD), Testicular germ cell tumors (TGCT), Thyroid carcinoma (THCA), Uterine corpus endometrial carcinoma (UCEC), and Uveal melanoma (UVM).
The CPTAC primary cancer types are breast, endometrial, renal, lung adenocarcinoma, lung squamous cell carcinoma, colorectal, ovarian, brain, head and neck, and pancreatic cancer.

\section{Patch-Level Tasks and Datasets}
\subsection{Classification Datasets}
\textbf{1. WSSS4LUAD (2 Classes):} is a lung adenocarcinoma dataset containing tiles of $200 \times 500$ pixels \cite{han2022wsss4luad}.
It encompasses three distinct classes: tumor, tumor-associated stroma, and normal. 
We conducted a binary classification of tumor vs. normal. 
The training dataset comprises 7,063 images, while the testing set comprises 3,028 images (2,015 tumors, 1,013 normal). 
\\
\textbf{2. SICAP (4 Classes):} is a prostate cancer dataset tailored for Gleason pattern classification \cite{silva2021self}.
It encompasses $512 \times 512$ pixels tiles extracted from 155 WSIs. 
The official training split comprises 9,959 images sourced from 124 WSIs, while the testing split includes 2,122 images from 31 WSIs. 
The dataset encompasses four labels, indicating the primary Gleason pattern (3, 4, or 5) or noncancerous (NC).
\\
\textbf{3. DigestPath (2 Classes):} is a colonoscopy dataset containing H\&E 660 tissue images \cite{da2022digestpath}.
Similar to PLIP, we performed tile-based zero-shot classification for Tumor Vs. Normal on the testing split containing 18814 images.
\\
\textbf{4. RC100K (9 Classes):} is a colorectal cancer dataset comprising H\&E stained images encompassing nine distinct classes including Adipose, background, debris, lymphocytes, mucus, smooth muscle, normal colon mucosa, cancer-associated stroma, and colorectal adenocarcinoma epithelium \cite{Kather2018_colorectal}. 
Tissue patches, corresponding to $224 \times 224$ pixels, are extracted at 20$\times$ magnification level.
The training dataset comprises 100K patches extracted from 86 patients, while the testing set consists of 7,180 images extracted from 50 patients. 
\\
\textbf{5. RenalCell (5 Classes):} dataset contains histology patterns of clear-cell renal cell carcinoma \cite{brummer2022integrative}. 
The dataset consists of 52,713 H\&E-stained images with $300 \times 300$ pixels captured at 40$\times$.
The dataset is annotated into five distinct classes, including red blood cells, renal cancer, normal tissue, torn adipose necrotic tissue, and muscle fibrous stroma blood vessels. 
\\
\textbf{6. MHIST (2 Classes):} is a colorectal polyps dataset that contains 3,152 tissue patches of size $224 \times 224$ pixels extracted at 40$\times$ magnification level from 328 WSIs \cite{wei2021petri}.
The tiles are annotated into two classes, including hyperplastic polyps and sessile serrated adenomas. 
For training, 2,175 tiles are utilized, while 977 tiles are reserved for testing. 
\\
\textbf{7. PatchCamelyon (2 Classes):} is a breast cancer dataset containing normal and metastatic tumor tissues \cite{veeling2018rotation}.
This dataset is collected from 400 WSIs, containing 327,680 H\&E stained histology images with 96$\times$96 pixel tiles.
The samples are extracted from lymph node sections at 10$\times$ magnification level to provide an increased field of view.
The official training, validation, and testing splits contain 262,144, 32,768, and 32,768 histology images.
\\
\textbf{8. Osteo (3 Classes):} dataset focuses on osteosarcoma and contains 1,144 patches of size $1024 \times 1024$ pixels \cite{arunachalam2019viable}.
The dataset is collected from 40 heterogeneous WSIs at 10$\times$ magnification level.
The dataset contains three distinct classes, including tumor, non-tumor, and necrotic tumor. 
The official train and test splits are provided with a ratio of 80:20.
\\
\textbf{9. BACH (4 Classes):} is a breast cancer dataset containing 500 large tiles, each with $2048 \times 1536$ pixels captured at 40$\times$ magnification and sampled from 500 WSIs \cite{iciar2018grand}. 
The dataset is classified into four different tissue types, including normal, benign, in-situ carcinoma, and invasive carcinoma. 
The official training (320 tiles) and testing (80 tiles) splits are provided. 
\\
\textbf{10. SkinCancer (16 Classes):} comprises 36,890 skin tissue patches extracted from 386 patients, each with $395 \times 395$ pixels \cite{kriegsmann2022deep}.
The patches are captured at 10$\times$ magnification from patients with basal cell carcinoma, squamous cell carcinoma, naevi, and melanoma. 
The tiles are categorized into 16 distinct categories, including chondral tissue, dermis, elastosis, epidermis, hair follicle, skeletal muscle, necrosis, nerves, sebaceous glands, subcutis, eccrine glands, vessels, BCC, SqCC, naevi, melanoma. 
The official train (88971), validation (72348), and testing (28039) splits are provided. 
\\
\textbf{11. LC25000Colon (2 Classes):} is a colon cancer dataset containing H\&E stained images each of size $768 \times 768$ pixels extracted from 500 patients \cite{Borkowski2019_LC25000}. 
This dataset contains two classes: benign colon tissue and colon adenocarcinomas. 
The official training and testing splits are provided with a ratio of 80:20.
\\
\textbf{12. LC25000Lung (3 Classes):} is a lung cancer dataset comprising 25K H\&E stained images corresponding to $768 \times 768$ pixels extracted from 750 patients \cite{Borkowski2019_LC25000}.
The dataset encompasses three classes, including lung adenocarcinoma, benign lung, and lung squamous cell carcinoma. 
The official training and testing splits are provided with a ratio of 80:20.
\\
\textbf{13. Databiox (3 Classes):} is an invasive ductal carcinoma dataset collected from pathological biopsy samples of 124 patients \cite{bolhasani2020histopathological}. 
This dataset comprises 922 samples, corresponding to $2100 \times 1574$ and $1276 \times 956$ pixels. 
Each sample is captured at four different levels of magnification, including 4$\times$, 10$\times$, 20$\times$, and 40$\times$. 
The samples are annotated into Grade I (well-differentiated), Grade II (moderately differentiated), and Grade III (poorly differentiated).

\subsection{VQA Datasets}
\textbf{1. Quilt-VQA:} comprises 1,961 VQA pairs curated from naturally occurring question–answer interactions narrated within the QUILT videos \cite{seyfioglu2024quilt}. It includes a total of 985 images and 1,283 QA pairs, with the questions categorized into 940 open-ended and 343 closed-ended types.
\\
\textbf{2. PathVQA:} contains 32,799 question–answer pairs derived from 4,998 pathology image–caption pairs collected from textbooks and digital libraries \cite{He2020_PathVQA}. The questions are divided into two categories: open-ended and closed-ended. Open-ended questions include types such as what, where, when, whose, how, and how much/how many, while closed-ended questions consist of yes/no formats. For our work, we utilize 6,761 samples from the evaluation set.
\\
\textbf{3. PMC-VQA v2}: consists of a test set containing 34,823 VQA pairs derived from non-compound images spanning various modalities and diseases \cite{zhang2023pmcvqa}. The dataset was constructed from image–caption pairs in PMC-OA articles and presented in a multiple-choice format. From this resource, we obtained the PMC-VQA-Subset, which includes 2,318 histopathology-specific VQA pairs.
\\
\textbf{4. PathMMU-VQA:} dataset contains 33,428 multiple-choice VQA pairs derived from 24,067 pathology images spanning diverse organs and diseases \cite{Sun2024_PathMMU}. Questions were generated using GPT-4V and include expert-validated explanations, with seven pathologists reviewing the validation and test sets. The dataset is organized into a training set, a validation set comprising 710 QA pairs from 510 images, and a test set containing 9,677 QA pairs from 7,213 images.

\section{WSI-Level Tasks and Datasets}
\subsection{Classification Datasets}
\textbf{1. PANDA (6 classes)} is the International Society of Urological Pathology (ISUP) grading task derived from the PANDA challenge \cite{Bulten2022_PANDA}.
It consists of 10,616 prostate cancer core needle biopsies of the prostate. 
Each slide is assigned an ISUP score that defines prostate cancer grade (6-class grading task). 
We removed the noisy labels and considered 9,555 slides only in which Grade 0 (G0) contains 2,603 WSIs, G1 contains 2,399 WSIs, G2 contains 1,209WSIs, G3 contains 1,118 WSIs, G4 contains 1,124 WSIs, and G5 contains 1,102). 
For training and evaluation, we employed 80:10:10 train–validation–test folds (7,647:954:954 WSIs). 
\\
\textbf{2. HunCRC (4 Classes)} is colorectal cancer screening dataset containing of 200 H\&E diagnostic histopathology WSIs \cite{pataki2022huncrc}.
Similar to UNI \cite{chen2024towards}, we employed a  4-way coarse-grained subtyping task using the categories of normal (10 slides), non-neoplastic lesion (38 slides), CRC (46 slides), and adenoma (106 slides), in which the ground-truth label was set by the study’s pathologist. 
For training and evaluation, we split the dataset into 50:25:25 train–validation–test folds (158:21:21 slides).
\\
\textbf{3. NSCLC-CPTAC (2 Classes)} is a Non-Small Cell Lung Carcinoma (NSCLC) subtyping based on CPTAC contains two classes including LUAD and LUSC \cite{Edwards2015CPTAC}.
We excluded slides that were frozen tissue, nontumor tissue, or were not labeled as having acceptable tumor segments, which resulted in 1,091 slides (578 LUAD and 513 LUSC). 
For training and evaluation, we divided datasets into train-validation-test folds with 80:10:10 ratio and 872:109:109 slides.
For zero-shot classification, we used 109 testing WSIs.
\\
\textbf{4. BRAINS \cite{roetzer2022digital} (30 classes)} is a dataset consists of H\&E histopathology WSIs of brain tissue selected from The Digital Brain Tumour Atlas an open histopathology resource. 
Similar to the CONCH \cite{lu2023visual}, we used a subset of 2,319 WSIs out of 3,114 WSIs and defined a 30-way fine-grained brain tumor subtyping task limited to diagnostic labels that have at least 30 slides. 
For the supervised dataset, we performed a 50–25–25 split for training (1,151 slides), validation (595 slides) and testing (573 slides).
For the zero-shot test set, we used the testing split of 573 slides. 
The WSI counts for each class in the dataset were also set according to the CONCH \cite{lu2023visual}.
\\
\textbf{5. BRCA-BRACS (3 classes)} is a BReast CAncer (BRCA) dataset containing 547 breast carcinoma H\&E WSIs from 187 patients sourced from the breast carcinoma subtyping task \cite{brancati2022bracs}.
The dataset contains 3 classes, including benign, atypical, and malignant tumor labels.
For training and evaluation, we used the official train–validation–test folds with a 72:12:16 ratio (395:65:87 WSIs).
\\
\textbf{6. TCGA-RCC (3 classes)} This dataset contains 477 normal WSIs and 726 WSIs with three cancer subtypes, including Kidney Renal Papillary Cell Carcinoma (KIRP) (218 WSIs) \cite{hutter2018cancer}, Kidney Renal
Clear Cell Carcinoma (KIRC) (390 WSIs), and Kidney Chromophobe Renal Cell Carcinoma (KICH) (118 WSIs) \cite{hutter2018cancer}.
\textit{Kindly note that we did not utilize any testing slide during the pre-training process of MRPT.}
\textit{All TCGA dataset results are reported under the test held-out setting.}
\\
\textbf{7. DHMC-RCC (5 classes)} is a Renal Cell Carcinoma (RCC) subtyping dataset consisting of 563 RCC H\&E diagnostic histopathology WSIs \cite{zhu2021development}.
The dataset contains six cancer subtypes, including primary Clear Cell Renal Cell Carcinoma (CCRCC) (344 slides), Papillary Renal Cell Carcinoma (PRCC) (101 slides), CHromophobe RCC (CHRCC) (23 slides), Renal Oncocytomas (ROCY)(66 slides), and benign cases (29 slides).
Similar to UNI \cite{chen2024towards}, for training and evaluation, we used a modified configuration of the train–validation–test folds with a 70:4:26 ratio (393:23:147 slides), with eight CHRCC cases moved from the test to the training fold due to CHRCC being absent in the training fold.
\\
\textbf{8. CAMELYON17 (CAM17) (2 Classes)} is a breast cancer dataset generated from a sentinel lymph node section of the Breast from 200 patients \cite{bandi2018detection}. 
The Camelyon17 dataset contains 1000 WSIs with 5 slides per patient. 
The WSI is labeled as metastasis vs. normal. The proposed model is evaluated on the official train (500 WSIs) and test set (500 WSIs) splits.
\\
\textbf{9. UBC-OCEAN (5 Classes)} comprises 538 whole-slide images, preprocessed to 527 that meet foreground tissue criteria \cite{UBC-OCEAN}. It covers five subtypes: Clear Cell (CC), Endometrioid (EC), High-Grade Serous Carcinoma (HGSC), Low-Grade Serous Carcinoma (LGSC), and Mucinous Carcinoma (MC). The dataset is split in a stratified fashion into train:validation:test with approximately 369:52:106 WSIs.
\\
\\
\subsection{VQA and Caption Datasets}

\textbf{1. SlideBench-Caption}: consists of 734 WSIs, each paired with human-written captions describing the slides \cite{Chen_2025_CVPR}. It is organized into a training set of 4,181 WSI-caption pairs and a test set of 734 WSIs, enabling evaluation of vision-language models on generating accurate and coherent slide descriptions.
\textit{All TCGA dataset results are reported under the test held-out setting.}
\\
\textbf{3. SlideBench-VQA (BNCB)}: contains 7,247 VQA pairs collected from 1,058 patients, created to assess SlideChat’s ability to perform zero-shot generalization across seven different classification tasks \cite{Chen_2025_CVPR}.
\\
\textbf{4. WSI-VQA:} dataset consists of 977 whole-slide images (WSIs) paired with 8,672 QA pairs, averaging about 8.9 QA pairs per WSI \cite{Chen2025_WSI-VQA}. Of these, 4,535 are close-ended and 4,137 are open-ended questions.
\\
\textbf{5. WSI-Bench (Pathological Capabilities)}: consists of 179,569 VQA pairs in total, with the training set containing 175,450 pairs (122,133 open-ended and 53,317 closed-ended) across 9,642 WSIs, and the test set comprising 4,119 pairs (2,838 open-ended and 1,281 closed-ended) from 208 WSIs \cite{Liang2024_WSI-LLaVA}.
All TCGA dataset results are reported under the test held-out setting.
\subsection{Report Generation Datasets}
\textbf{1. HistGen-Report}: comprises 7,753 WSI–report pairs collected from the TCGA platform and refined using large language models to ensure coherent, diagnostically relevant report texts \cite{guo2024histgen}.
\\
\textbf{2. WSI-Bench(Report)}: In addition to its pathological capabilities, WSI-Bench includes 208 report generation VQA pairs \cite{Liang2024_WSI-LLaVA}.
All TCGA dataset results are reported under the test held-out setting.

\end{document}